\documentclass[sigconf,authorversion]{acmart}

\acmSubmissionID{291}

\makeatletter
\DeclareRobustCommand\onedot{\futurelet\@let@token\@onedot}
\def\@onedot{\ifx\@let@token.\else.\null\fi\xspace}

\makeatother

\newcommand{\figref}[1]{Figure~\ref{fig:#1}}%

\newcommand{\secref}[1]{Section~\ref{sec:#1}}

\newcommand{\ignore}[1]{}   %

\newcommand{\mpage}[2]
{
\begin{minipage}{#1\linewidth}\centering
#2
\end{minipage}
}

\usepackage{enumitem}

\AtBeginDocument{%
  \providecommand\BibTeX{{%
    \normalfont B\kern-0.5em{\scshape i\kern-0.25em b}\kern-0.8em\TeX}}}

\copyrightyear{2023} 
\acmYear{2023} 
\setcopyright{acmlicensed}\acmConference[SA Conference Papers '23]{SIGGRAPH Asia 2023 Conference Papers}{December 12--15, 2023}{Sydney, NSW, Australia}
\acmBooktitle{SIGGRAPH Asia 2023 Conference Papers (SA Conference Papers '23), December 12--15, 2023, Sydney, NSW, Australia}
\acmPrice{15.00}
\acmDOI{10.1145/3610548.3618153}
\acmISBN{979-8-4007-0315-7/23/12}

\begin{document}

\title{Single-Image 3D Human Digitization with Shape-Guided Diffusion}

\author{Badour AlBahar}
\affiliation{%
  \institution{Kuwait University}
  \city{Kuwait City}
  \country{Kuwait}
  }
\email{badour.albahar@ku.edu.kw}

\author{Shunsuke Saito}
\affiliation{%
  \institution{Meta}
  \city{Pittsburgh}
  \state{Pennsylvania}
  \country{USA}
  }
\email{shunsukesaito@meta.com}

\author{Hung-Yu Tseng}
\affiliation{%
  \institution{Meta}
  \city{Seattle}
  \state{Washington}
  \country{USA}
  }
\email{hungyutseng@meta.com}

\author{Changil Kim}
\affiliation{%
  \institution{Meta}
  \city{Seattle}
  \state{Washington}
  \country{USA}
  }
\email{changil@meta.com}

\author{Johannes Kopf}
\affiliation{%
  \institution{Meta}
  \city{Seattle}
  \state{Washington}
  \country{USA}
  }
\email{jkopf@meta.com}

\author{Jia-Bin Huang}
\affiliation{%
  \institution{University of Maryland}
  \city{College Park}
  \state{Maryland}
  \country{USA}
  }
\email{jbhuang@umd.edu}

\begin{abstract}
We present an approach to generate a 360-degree view of a person with a consistent, high-resolution appearance from a \emph{single} input image.
NeRF and its variants typically require videos or images from different viewpoints. 
Most existing approaches taking monocular input either rely on ground-truth 3D scans for supervision or lack 3D consistency. 
While recent 3D generative models show promise of 3D consistent human digitization, these approaches do not generalize well to diverse clothing appearances, and the results lack photorealism. 
Unlike existing work, we utilize high-capacity 2D diffusion models pretrained for general image synthesis tasks as an appearance prior of clothed humans. 
To achieve better 3D consistency while retaining the input identity, we progressively synthesize multiple views of the human in the input image by inpainting missing regions with shape-guided diffusion conditioned on silhouette and surface normal.
We then fuse these synthesized multi-view images via inverse rendering to obtain a fully textured high-resolution 3D mesh of the given person.
Experiments show that our approach outperforms prior methods and achieves photorealistic 360-degree synthesis of a wide range of clothed humans with complex textures from a single image. 
\end{abstract}



\begin{teaserfigure}
\begin{center}
\mpage{0.13}{\includegraphics[width=\linewidth, trim=260 0 260 0, clip]{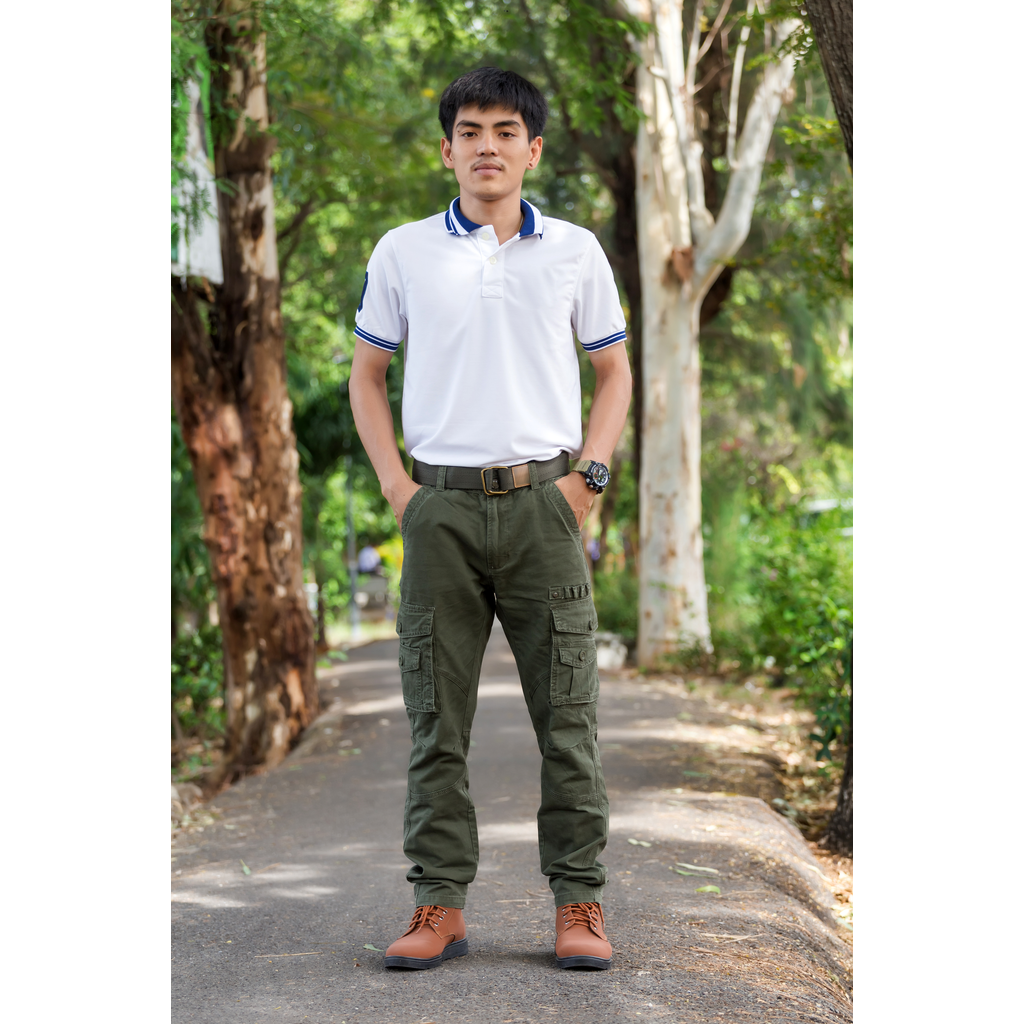}}\hfill
\mpage{0.07}{\includegraphics[width=\linewidth, trim=360 0 360 0, clip]{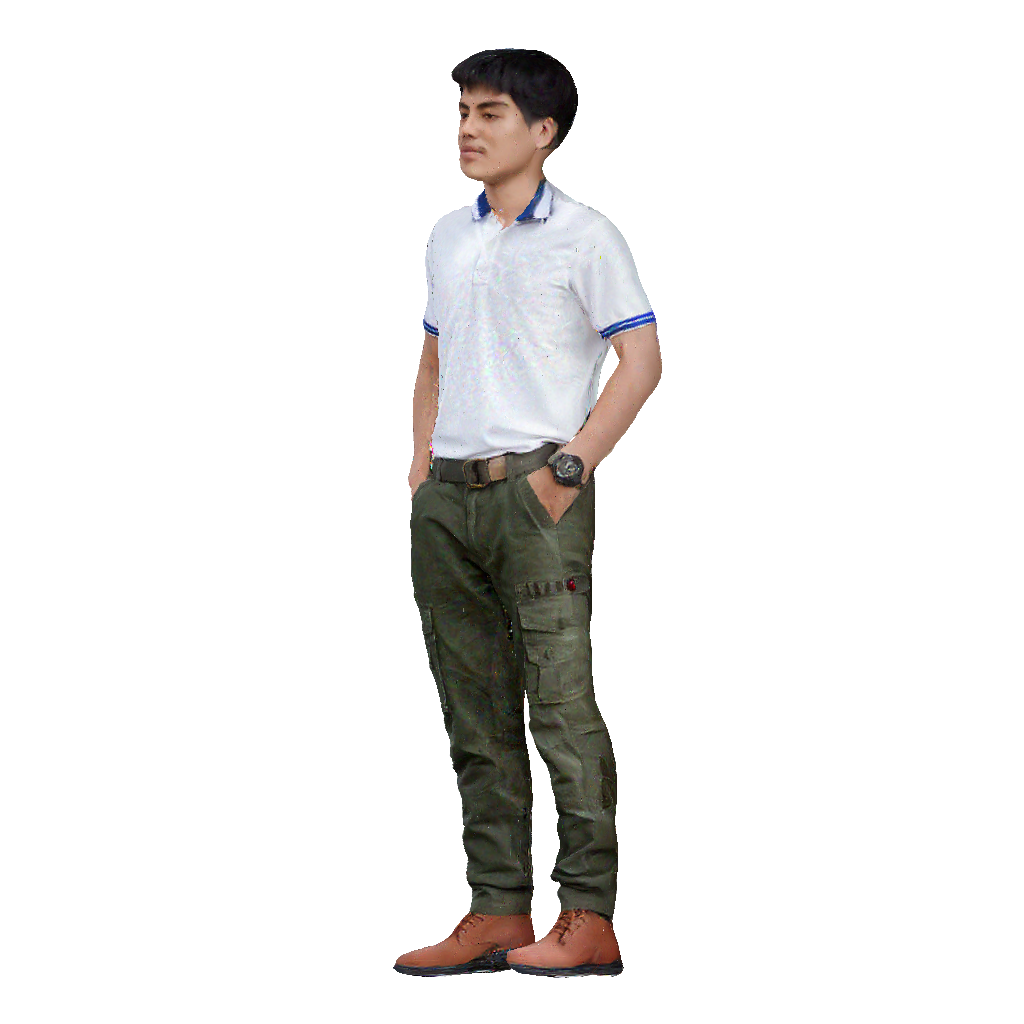}}\hfill
\mpage{0.07}{\includegraphics[width=\linewidth, trim=360 0 360 0, clip]{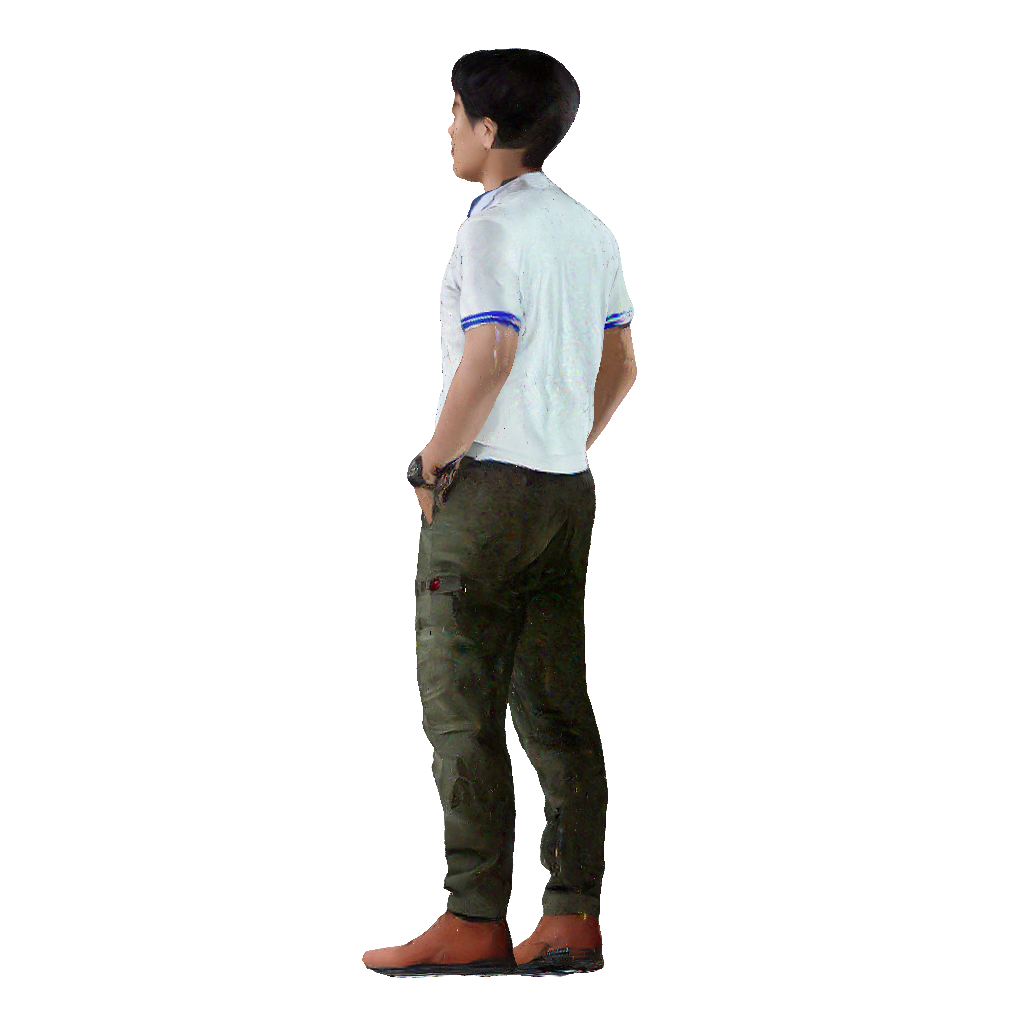}}\hfill
\mpage{0.07}{\includegraphics[width=\linewidth, trim=360 0 360 0, clip]{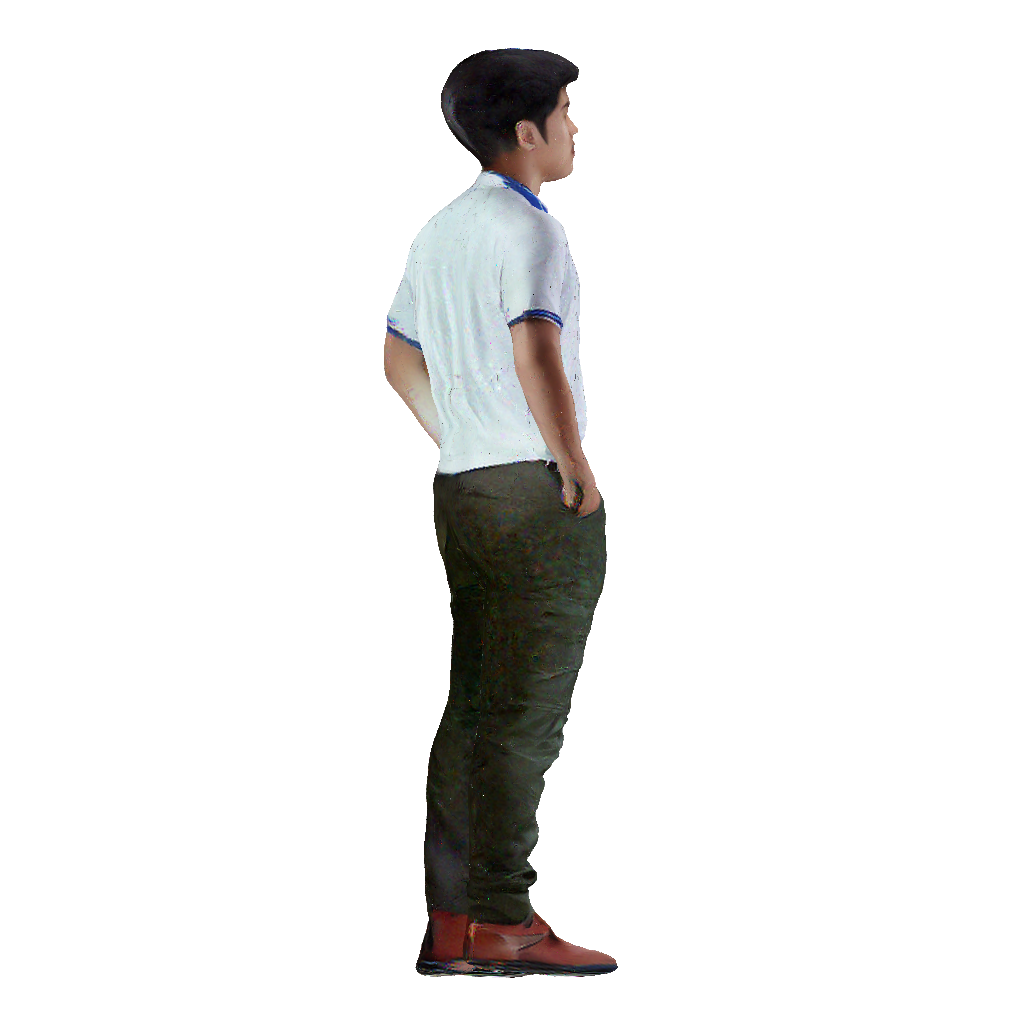}}\hfill
\mpage{0.07}{\includegraphics[width=\linewidth, trim=360 0 360 0, clip]{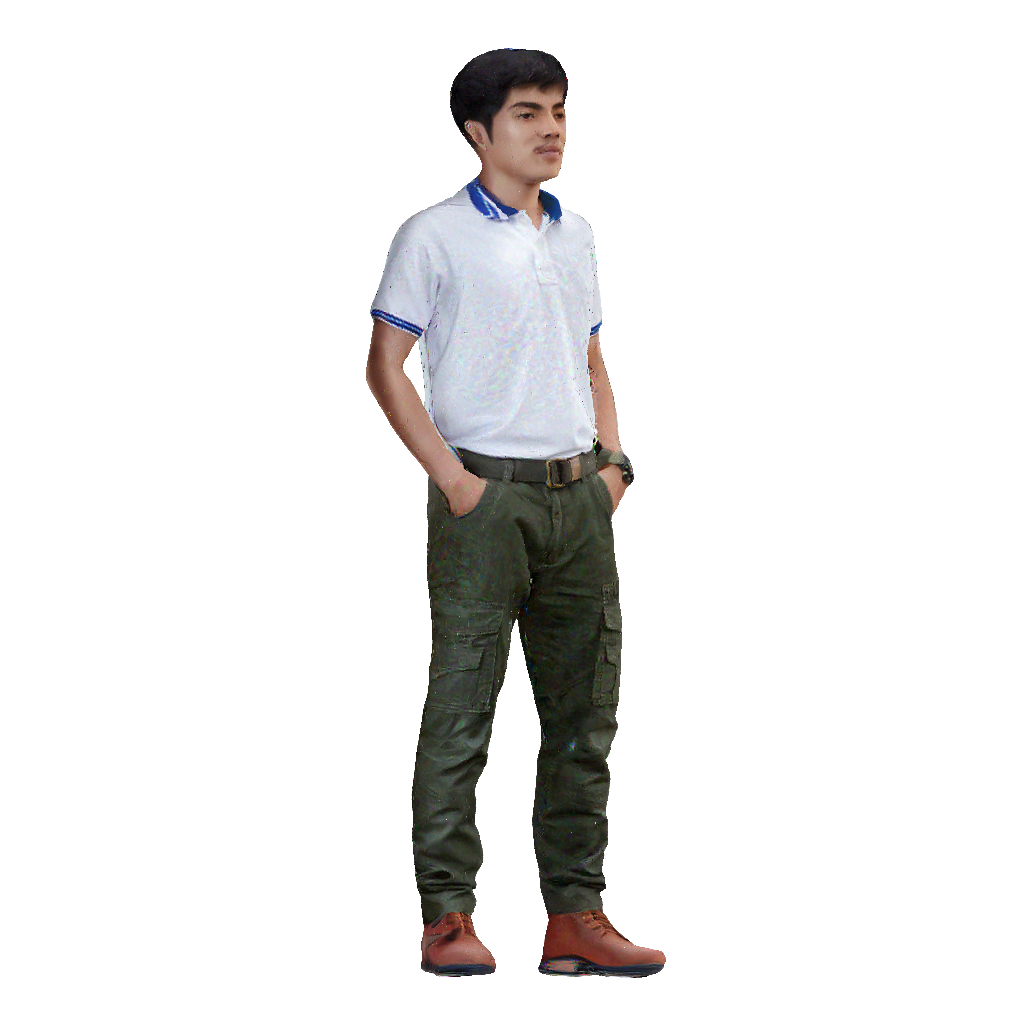}}\hfill
\hspace{2mm}
\mpage{0.13}{\includegraphics[width=\linewidth, trim=260 0 260 0, clip]{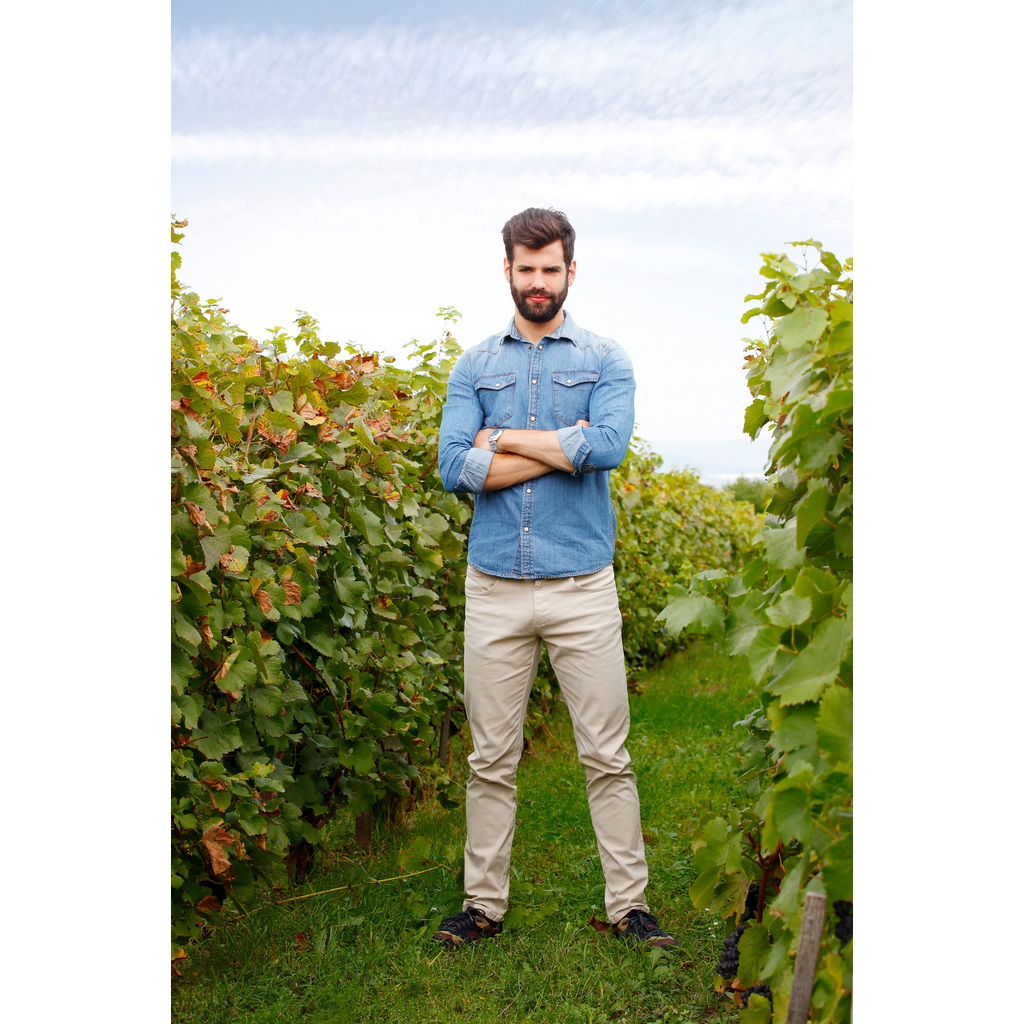}}\hfill
\mpage{0.07}{\includegraphics[width=\linewidth, trim=360 0 360 0, clip]{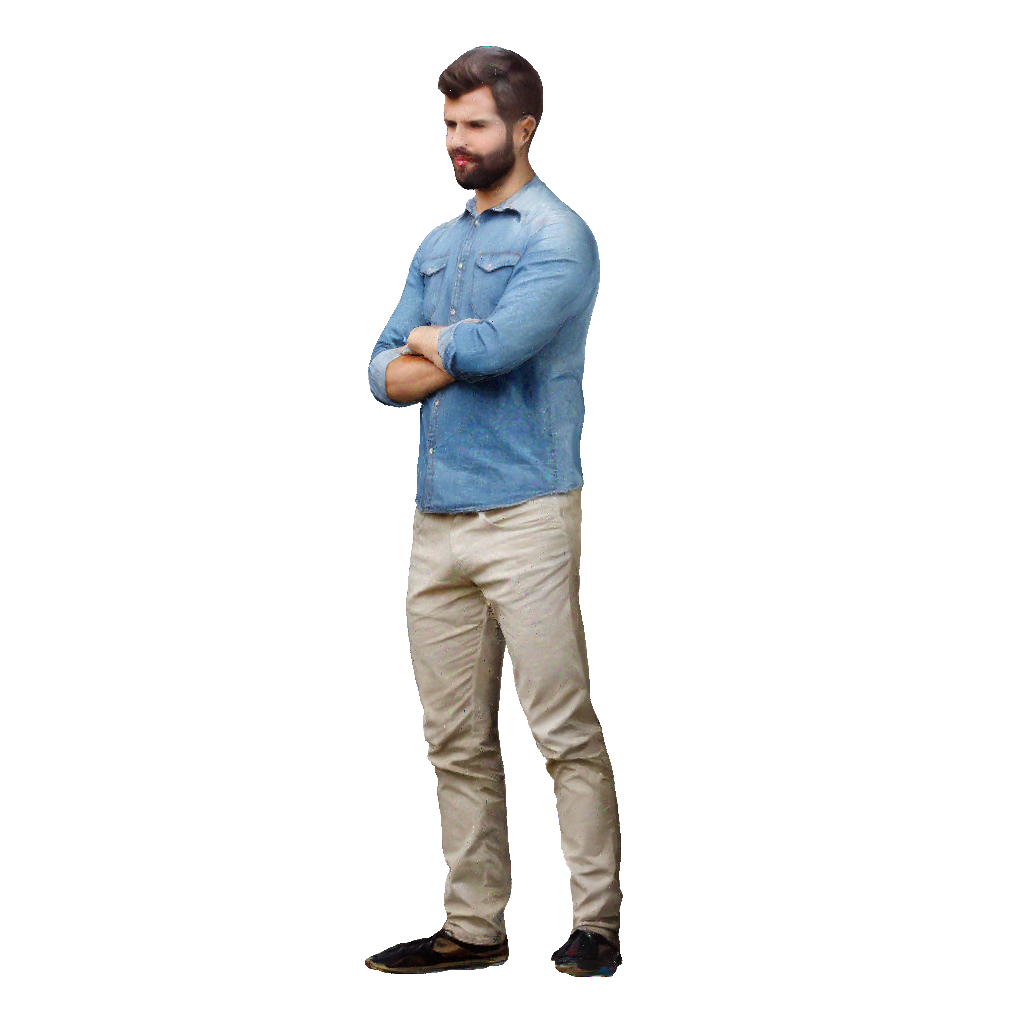}}\hfill
\mpage{0.07}{\includegraphics[width=\linewidth, trim=360 0 360 0, clip]{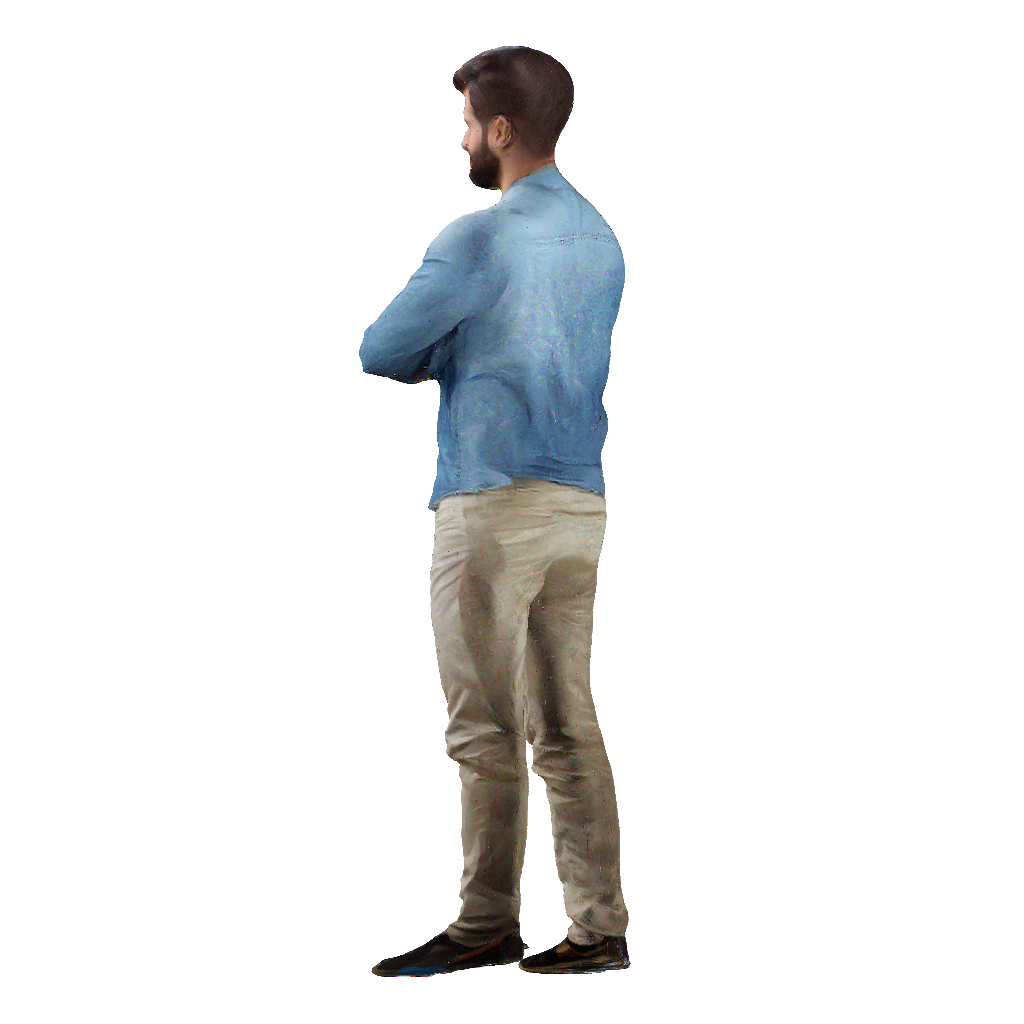}}\hfill
\mpage{0.07}{\includegraphics[width=\linewidth, trim=360 0 360 0, clip]{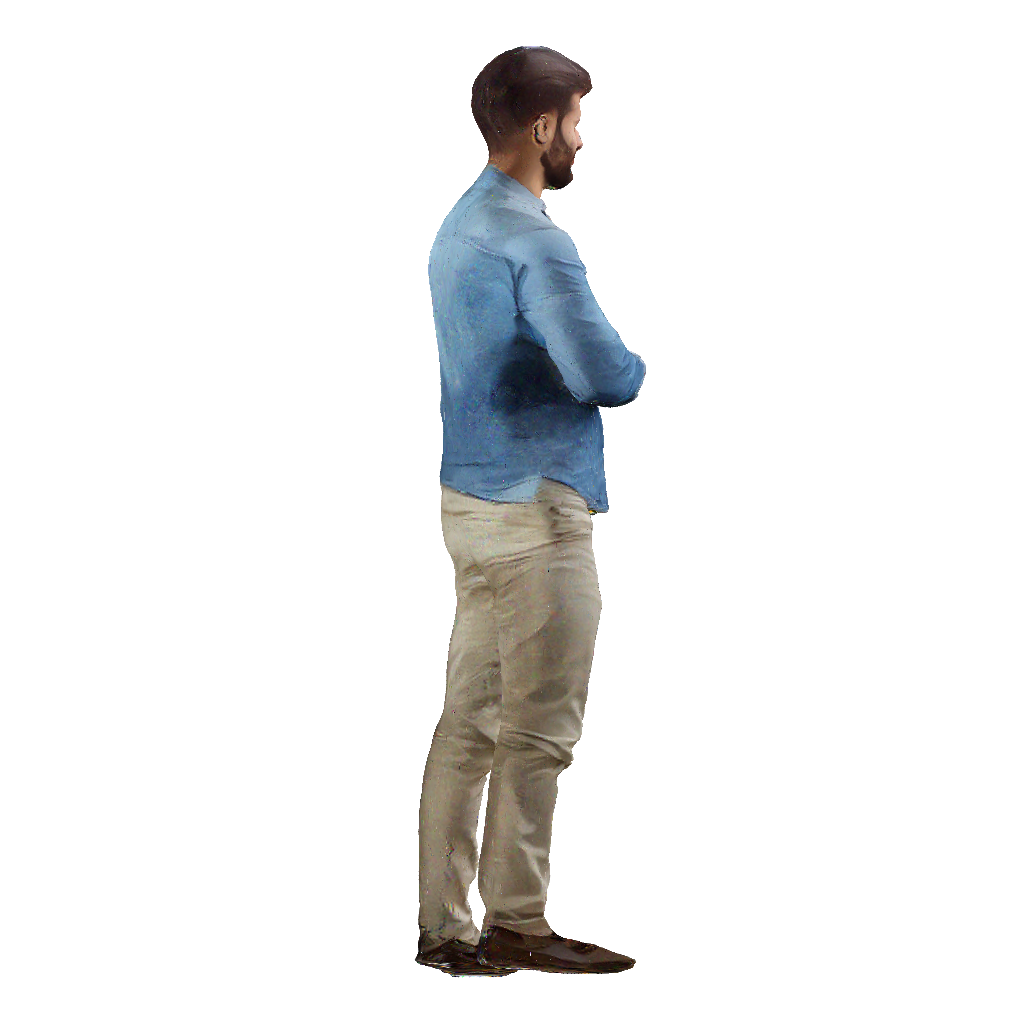}}\hfill
\mpage{0.07}{\includegraphics[width=\linewidth, trim=360 0 360 0, clip]{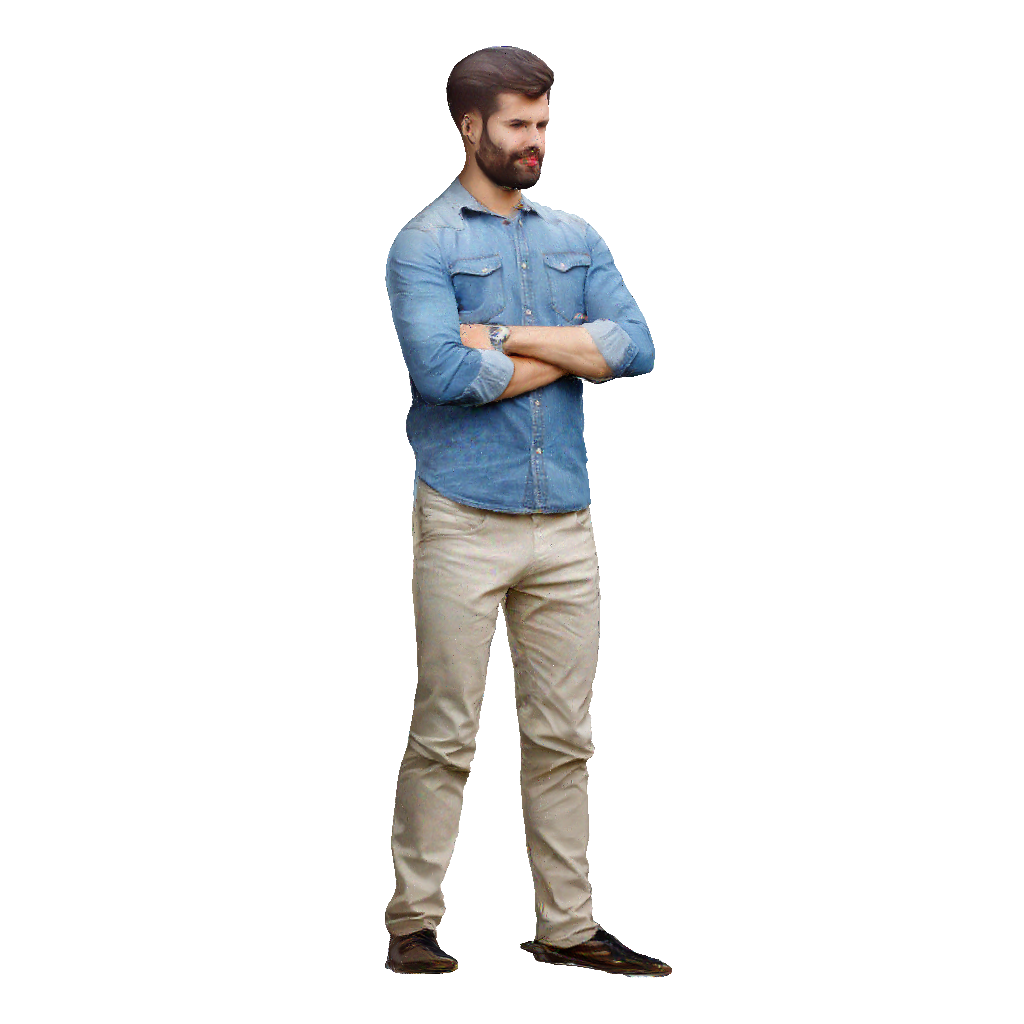}}\\
\vspace{-2mm}
\mpage{0.13}{\includegraphics[width=\linewidth, trim=260 0 260 0, clip]{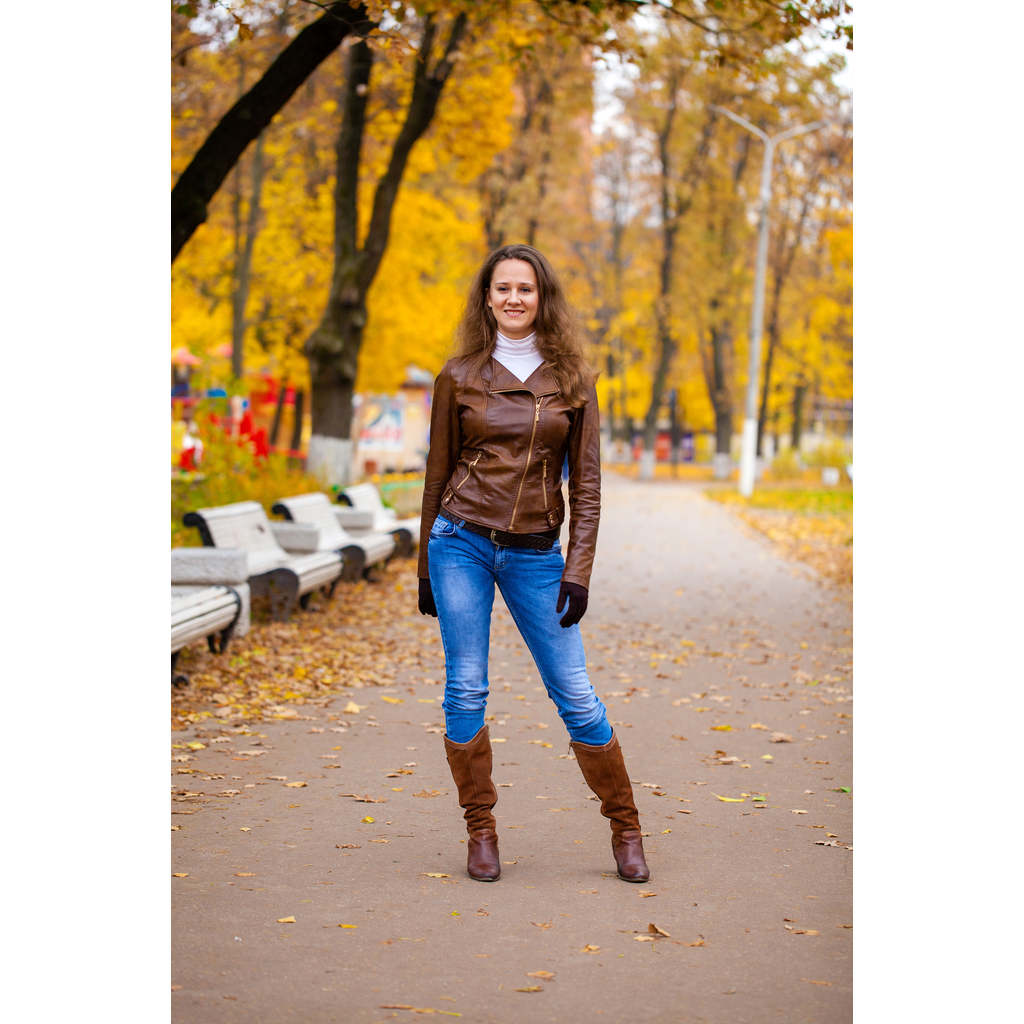}}\hfill
\mpage{0.07}{\includegraphics[width=\linewidth, trim=390 0 330 0, clip]{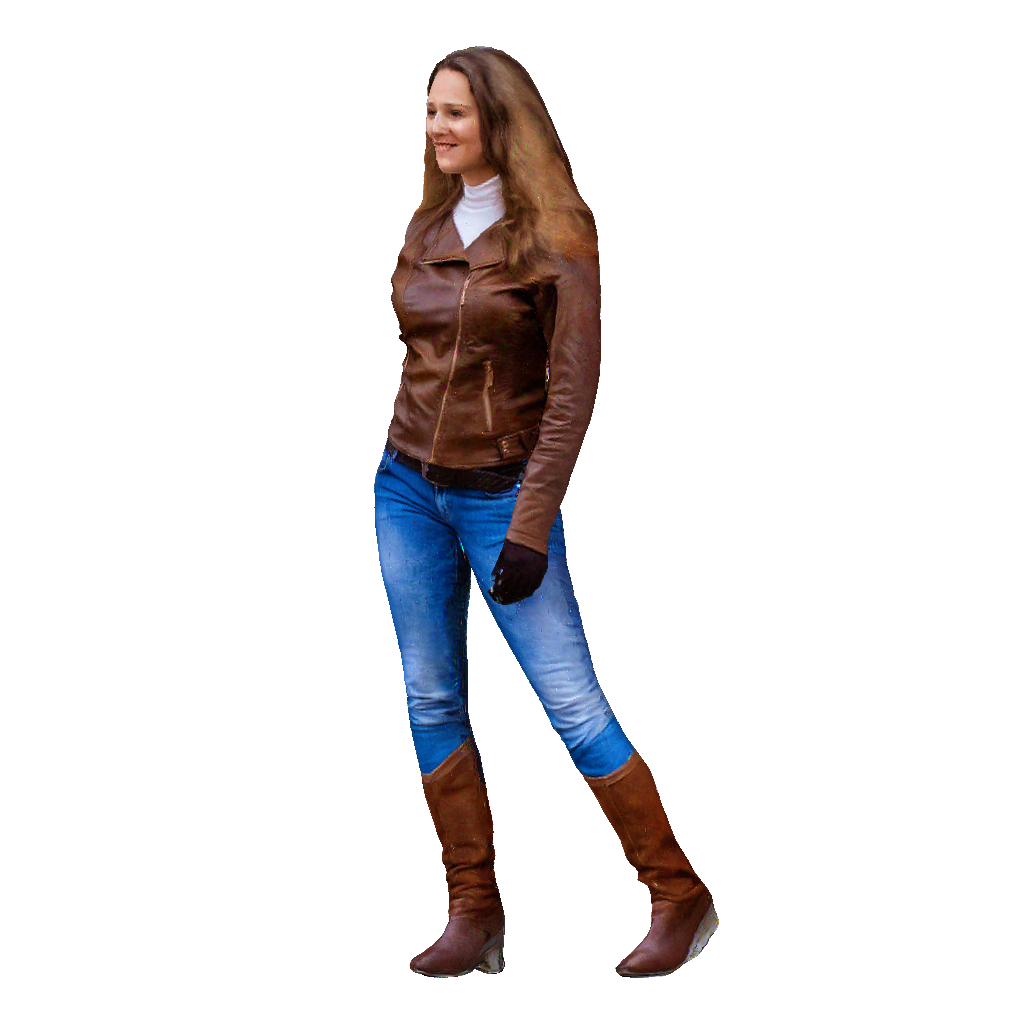}}\hfill
\mpage{0.07}{\includegraphics[width=\linewidth, trim=360 0 360 0, clip]{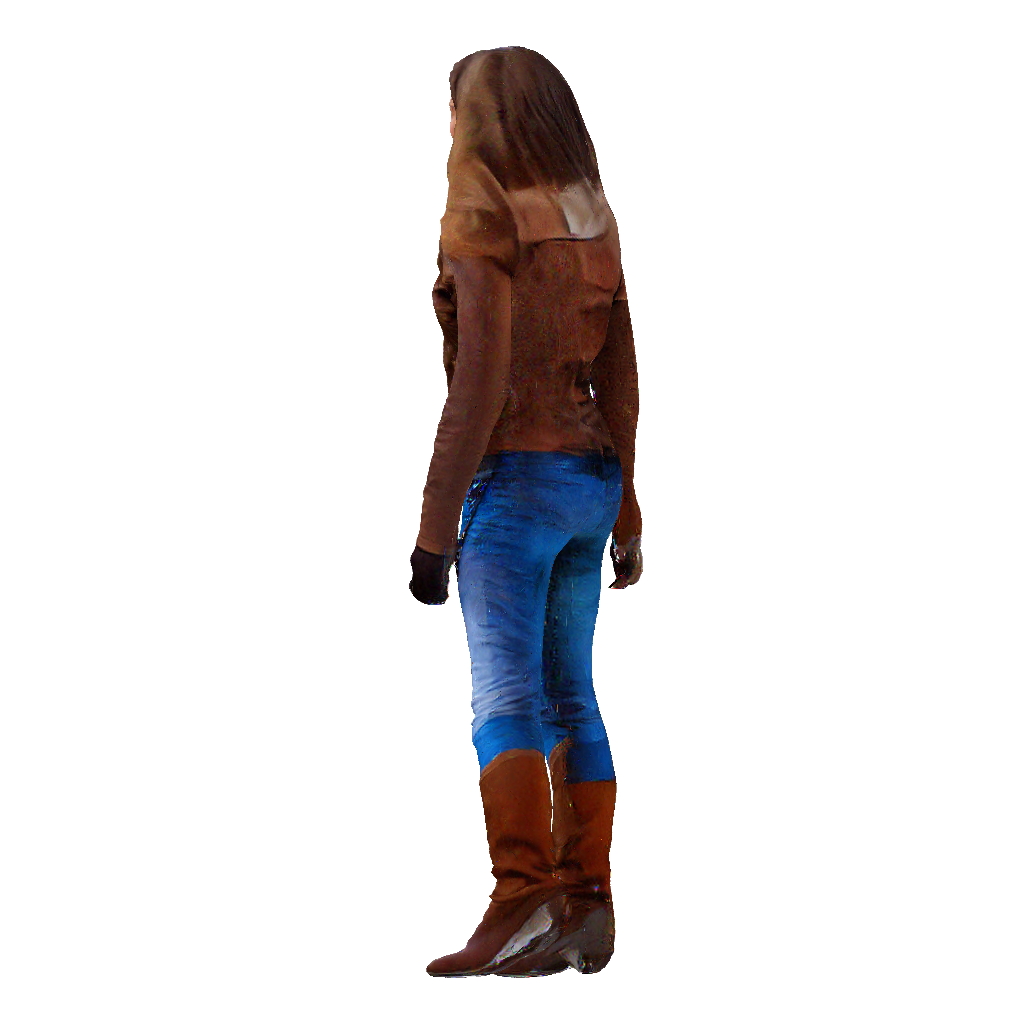}}\hfill
\mpage{0.07}{\includegraphics[width=\linewidth, trim=320 0 400 0, clip]{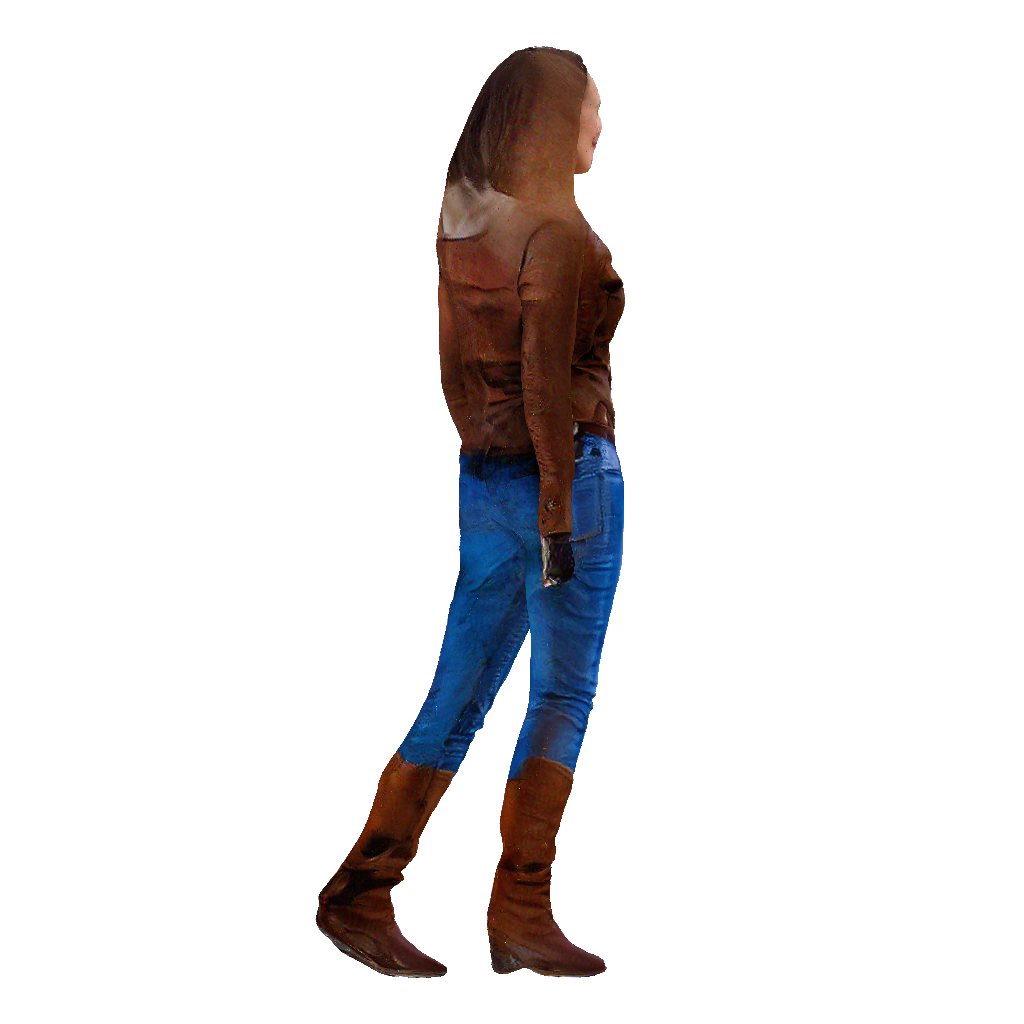}}\hfill
\mpage{0.07}{\includegraphics[width=\linewidth, trim=330 0 390 0, clip]{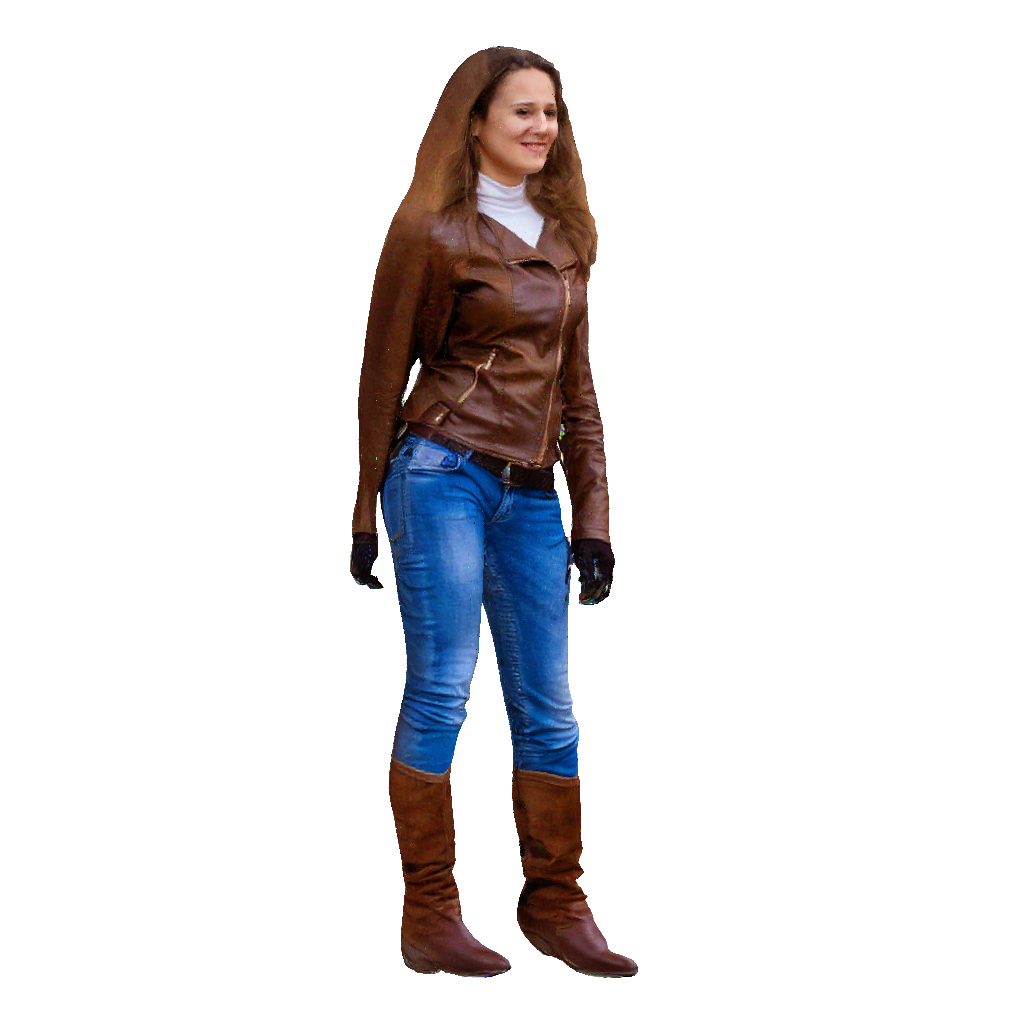}}\hfill
\hspace{2mm}
\mpage{0.13}{\includegraphics[width=\linewidth, trim=260 0 260 0, clip]{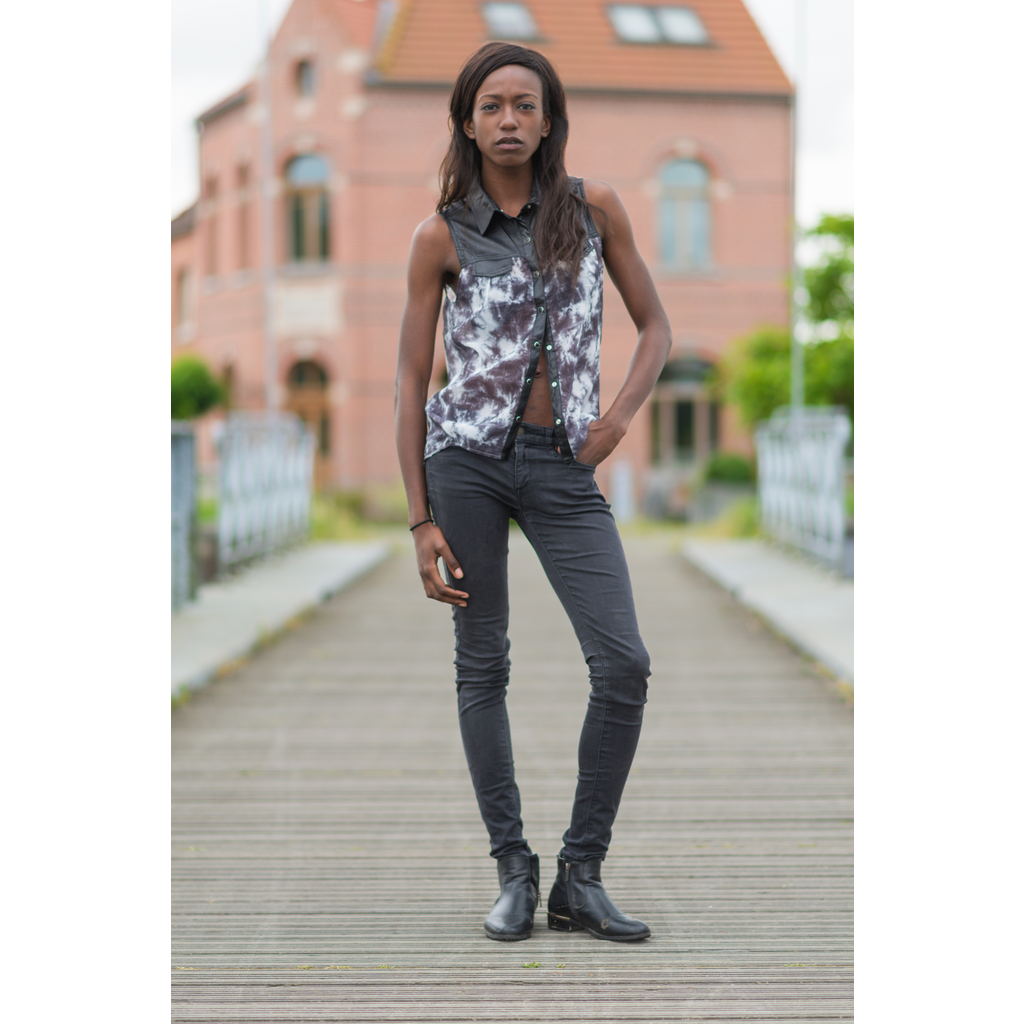}}\hfill
\mpage{0.07}{\includegraphics[width=\linewidth, trim=360 0 360 0, clip]{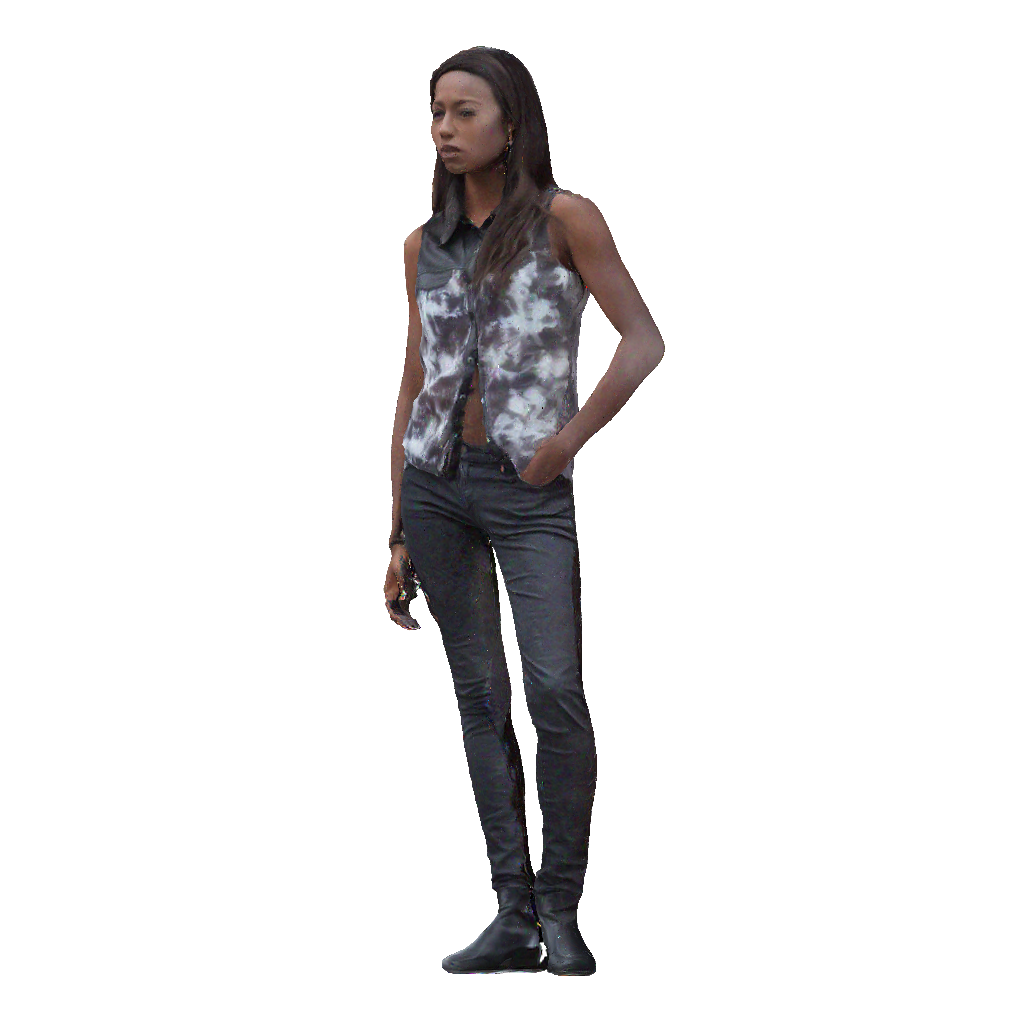}}\hfill
\mpage{0.07}{\includegraphics[width=\linewidth, trim=360 0 360 0, clip]{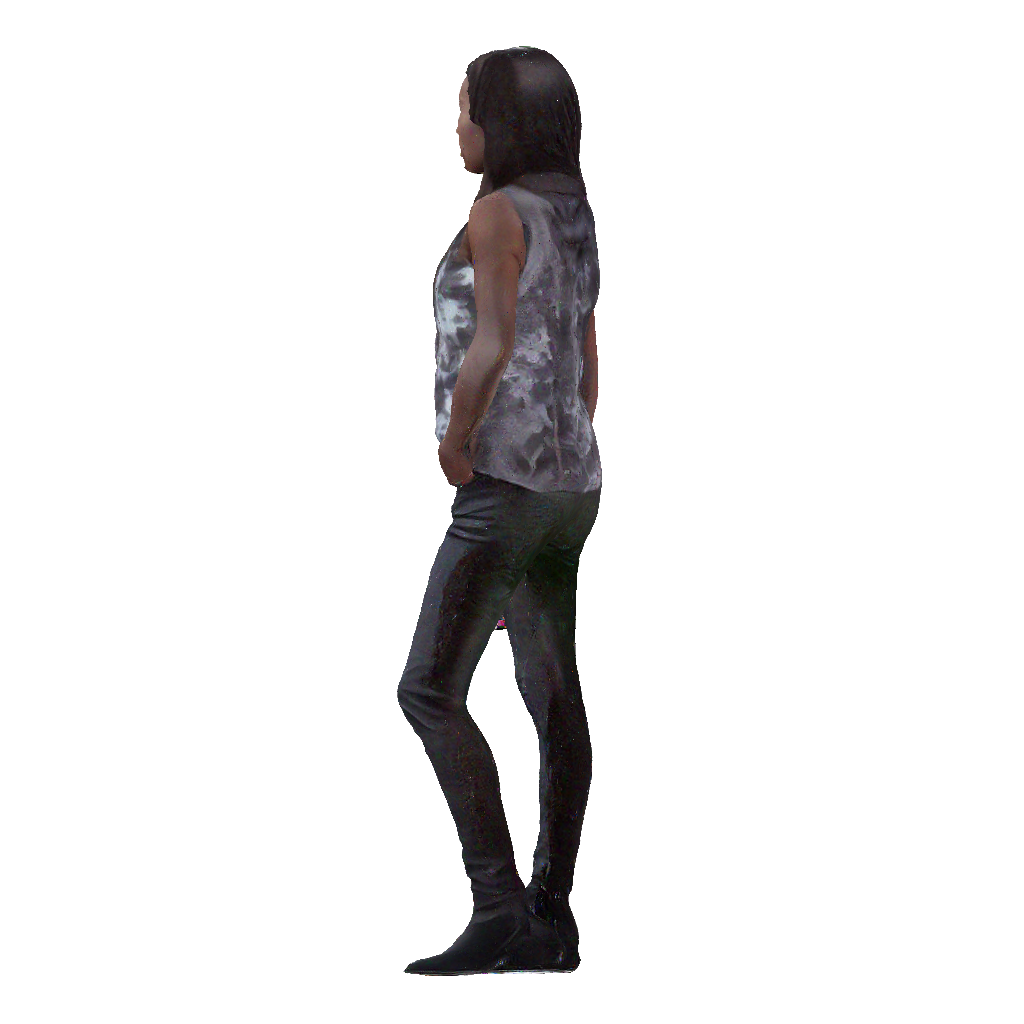}}\hfill
\mpage{0.07}{\includegraphics[width=\linewidth, trim=360 0 360 0, clip]{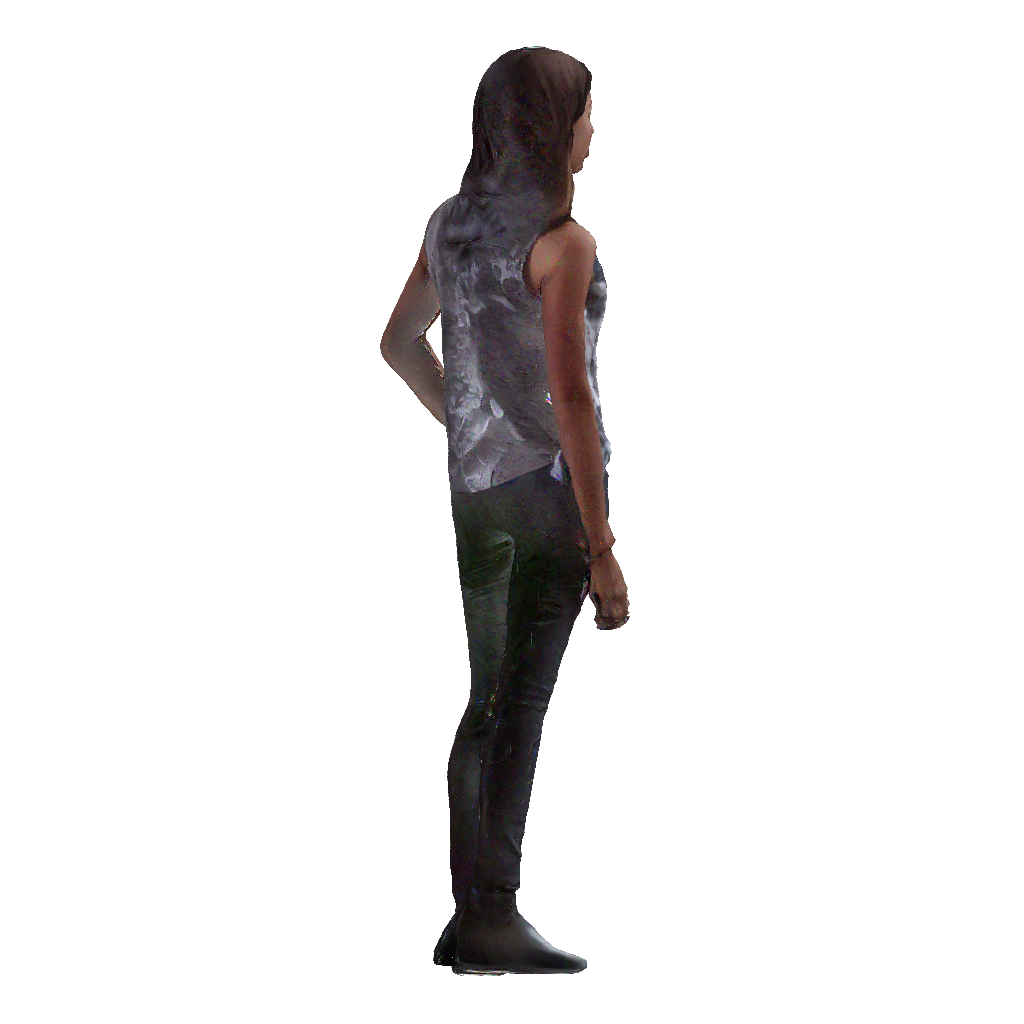}}\hfill
\mpage{0.07}{\includegraphics[width=\linewidth, trim=360 0 360 0, clip]{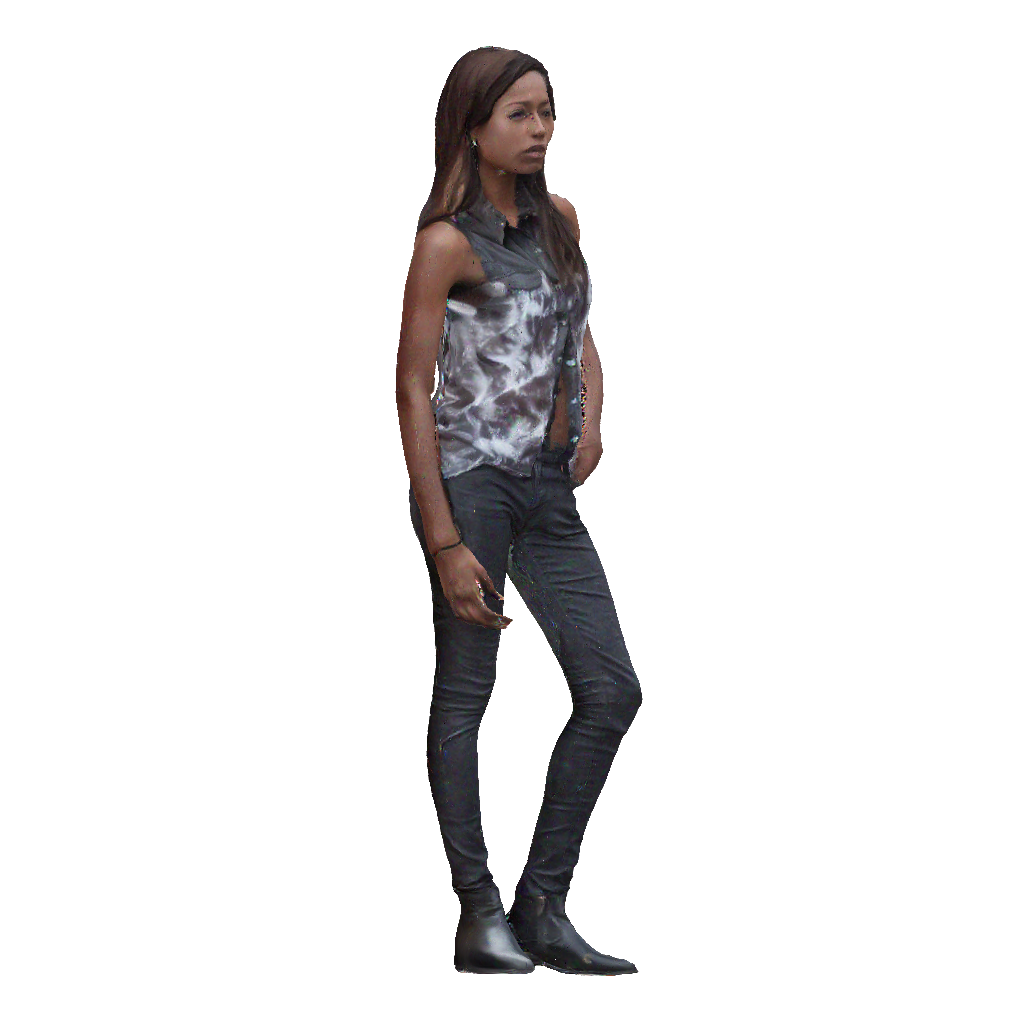}}\\
\mpage{0.13}{\small{Input image}}\hfill
\mpage{0.35}{$\underbrace{\hspace{\textwidth}}_{\substack{\vspace{-4.0mm}\\\colorbox{white}{~~360$^\circ$ generation~~}}}$}\hfill
\hspace{2mm}
\mpage{0.13}{\small{Input image}}\hfill
\mpage{0.35}{$\underbrace{\hspace{\textwidth}}_{\substack{\vspace{-5.0mm}\\\colorbox{white}{~~360$^\circ$ generation~~}}}$}\\
\vspace{-3mm}
\captionof{figure}{\textbf{3D Human Digitization from a Single Image.} For a single image as input, our approach synthesizes the 3D consistent texture of a person without relying on any 3D scans for supervised training. Our key idea is to leverage high-capacity 2D diffusion models pretrained for general image synthesis tasks as a human appearance prior.
Images from Adobe Stock.}
\label{fig:teaser}
\end{center}
\end{teaserfigure}

\maketitle

\section{Introduction}
\label{sec:intro}
\begin{figure}[t]
\centering

\hfill
\mpage{0.16}{\includegraphics[width=\linewidth, trim=185 0 165 0, clip]{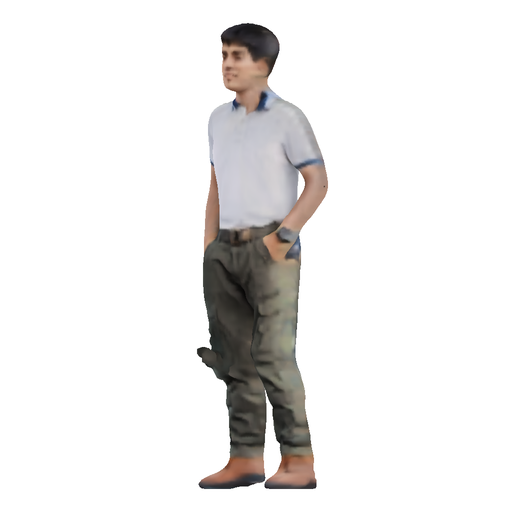}}\hfill
\mpage{0.16}{\includegraphics[width=\linewidth, trim=175 0 175 0, clip]{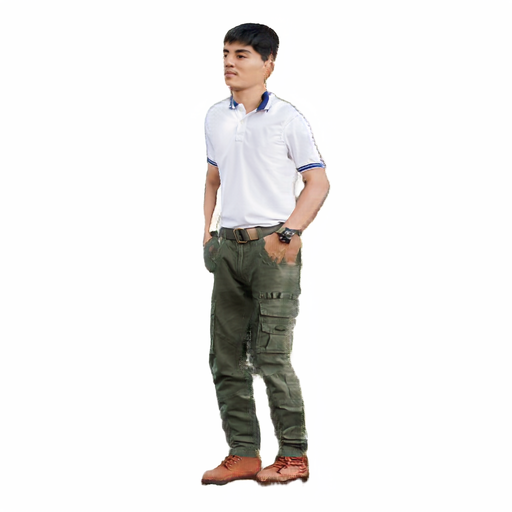}}\hfill
\mpage{0.16}{\includegraphics[width=\linewidth, trim=175 0 175 0, clip]{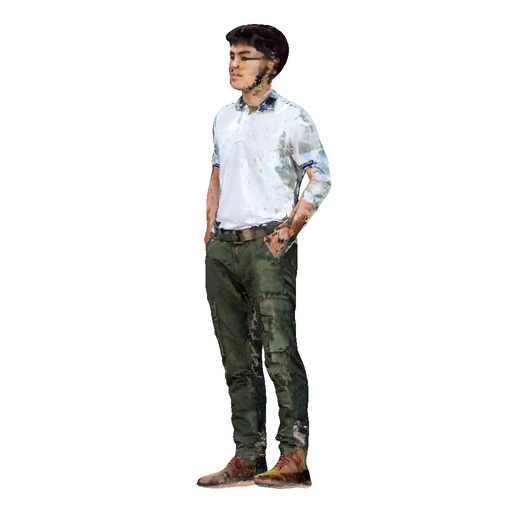}}\hfill
\mpage{0.16}{\includegraphics[width=\linewidth, trim=155 0 195 0, clip]{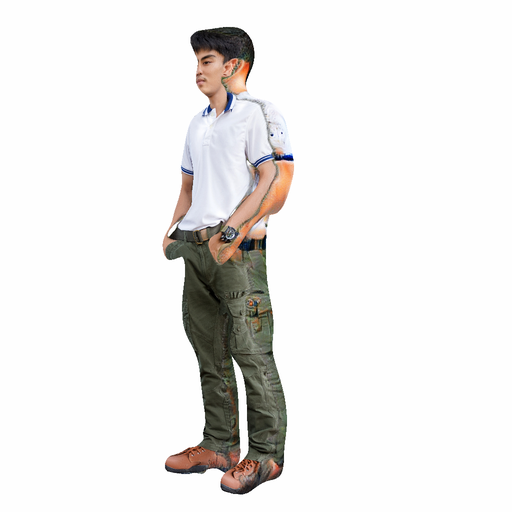}}\hfill
\mpage{0.16}{\includegraphics[width=\linewidth, trim=350 0 350 0, clip]{images/AdobeStock/ours/197597752/view_010.png}}\\

\hfill
\mpage{0.16}{{PIFu}}\hfill
\mpage{0.16}{{Imp++}}\hfill
\mpage{0.16}{{TEXTure}}\hfill
\mpage{0.16}{{Magic123}}\hfill
\mpage{0.16}{{Ours}}\hfill\\

\vspace{-2mm}
\captionof{figure}{
\textbf{Limitations of existing methods.}
Existing 3D human generation approaches from a single image lack photorealism.
Existing methods such as PIFu~\cite{saito2019pifu} suffer from blurriness; Impersonator++~\cite{liu2021liquid} tends to duplicate content from the front view, suffering from projection artifacts; TEXTure~\cite{richardson2023texture} fails to preserve the appearance of the input view and results in saturated colors; Magic123~\cite{qian2023magic123} fails to synthesize realistic shape and appearance. Images from Adobe Stock.
}
\label{fig:motivation}
\end{figure}

A photorealistic 3D human synthesis is indispensable for a myriad of applications in various fields, including fashion, entertainment, sports, and AR/VR. 
However, creating a photorealistic 3D human model typically requires multi-view images~\cite{liu2021neural, peng2021neural, peng2021animatable, kwon2021neural} or 3D scanning systems~\cite{bagautdinov2021driving,saito2021scanimate} as input, which hinders everyone from effortlessly experiencing personalized 3D human digitization. 
In this work, we aim to create a photorealistic 3D human that can be rendered from arbitrary viewpoints from a \emph{single} input image. 
Despite its attractive utility, reducing the input to monocular data is highly challenging because the person's backside is not observable, and 3D reconstruction from a single image inherently suffers from depth ambiguity.

To address these challenges, data-driven methods have made significant progress in recent years by incorporating prior information into various 3D representations such as meshes~\cite{alldieck2019learning}, voxels~\cite{varol2018bodynet}, and neural fields~\cite{saito2019pifu}. 
While the geometric fidelity of 3D reconstruction drastically improved over the last several years~\cite{saito2020pifuhd,alldieck2022photorealistic,zheng2020pamir,xiu2022icon,huang2020arch,he2021arch++}, its \emph{appearance}, especially for the occluded regions, is still far from photorealistic (\figref{motivation}). 
This is primarily because these approaches require 3D ground-truth data for supervision, and the available 3D scans of clothed humans are insufficient to learn the entire span of clothing appearance. 
The appearance of clothing is significantly more diverse than the geometry, and creating a large set of high-quality textured 3D scans of people remains infeasible.

An image collection in the wild is another source of human appearance prior. 
Images are easily accessible at scale and provide a high variation of clothing appearances. 
By leveraging large-scale image datasets and high-capacity generative models~\cite{karras2019style,karras2020analyzing}, 2D human synthesis approaches show impressive reposing of clothed humans from a single image~\cite{lewis2021tryongan,albahar2021pose}. 
However, they often produce an incoherent appearance with the input image for large rotations because their underlying representation is not in 3D.
While 3D generative models have recently demonstrated 3D-consistent view synthesis of clothed humans~\cite{EVA3D,bergman2022gnarf,zhang2022avatargen}, we observe that these approaches do not generalize well to various clothing appearances and the results are not sufficiently photorealistic. 

In this paper, we argue that the suboptimal performance of existing approaches stems from the limited diversity of training data. 
However, expanding existing 2D-clothed human datasets also requires nontrivial curation and annotation efforts. 
To address this limitation, we propose a simple yet effective algorithm to create a 3D consistent textured human from a single image \emph{without} relying on a curated 2D clothed human dataset for appearance synthesis. 
Our key idea is to utilize powerful 2D generative models trained on an extremely large corpus of images as a human appearance prior. 
In particular, we use latent diffusion models~\cite{rombach2021highresolution}, which allows us to synthesize diverse and photorealistic images. 
Unlike recent works that leverage 2D diffusion models for 3D object generation from text inputs~\cite{poole2022dreamfusion,lin2022magic3d,richardson2023texture}, we employ diffusion models to reconstruct a 360-degree view of a real person in the input image in a 3D-consistent manner.

We first reconstruct the 3D geometry of the person using an off-the-shelf tool~\cite{saito2020pifuhd} and then generate the back-view of the input image using a 2D single image human reposing approach~\cite{albahar2021pose} to ensure that the completed views are consistent with the input view.
Next, we synthesize multi-view images of the person by progressively inpainting novel views utilizing a pretrained inpainting diffusion model guided by both normal and silhouette maps to constrain the synthesis to the underlying 3D structure.
To generate a (partial) novel view, we aggregate all other views by blending their RGB color based on importance. 
Similar to previous work~\cite{xiang20233d, buehler2001unstructured, rong2022bvs}, we use the angular differences between the visible pixels of those views and the current view of interest as well as their distance to the nearest missing pixel to determine the appropriate weight for each view in the blending process. 
This ensures that the resulting multi-view images are consistent with each other.
Finally, we perform multi-view fusion by accounting for slight misalignment in the synthesized multi-view images to obtain a fully textured high-resolution 3D human mesh.

Our experiments show that the proposed approach achieves a more detailed and faithful synthesis of clothed humans than prior methods without requiring high-quality 3D scans or curated large-scale clothed human datasets. 

\emph{Our contributions} include:
\begin{itemize}[noitemsep,topsep=0pt]
\item We demonstrate, for the first time, that a 2D diffusion model trained for general image synthesis can be utilized for 3D textured human digitization from a \emph{single} image.
\item Our approach preserves the shape and the structural details of the underlying 3D structure by using both normal maps and silhouette to guide the diffusion model.
\item We enable 3D consistent texture reconstruction by fusing the synthesized multi-view images into the shared UV texture map.
\end{itemize}

\section{Related Work}
\label{sec:related}

\subsection{2D human synthesis.}
Generative adversarial networks (GANs) enable the photorealistic synthesis of human faces~\cite{karras2019style,karras2020analyzing} and bodies~\cite{fu2022stylegan}. While these models are unconditional, several works extend them to conditional generative models such that we can control poses while retaining the identity of an input subject.
By incorporating additional conditions these works can achieve human reposing~\cite{albahar2019guided,albahar2021pose,sarkar2021style,ma2017pose,ma2018disentangled,ADGAN_2020,siarohin2018deformable,PATN_2019,GFLA_2020,liu2021liquid}, virtual try-on~\cite{albahar2021pose,lewis2021tryongan}, motion transfer~\cite{chan2019everybody,aberman2019deep,yoon2020pose,liu2021liquid}.
Pose-with-style~\cite{albahar2021pose} utilizes dense pose~\cite{guler2018densepose} to warp input images to the target view as an initialization of the synthesis. Impersonator++~\cite{liu2021liquid} further improves the robustness to a large pose change by leveraging a parametric human body model~\cite{loper2015smpl} and warping blocks to better preserve the information from the input. While these methods enable the control of viewpoints by changing the input pose, the results suffer from view inconsistency. In contrast, our approach achieves 3D consistent generation of textured clothed humans.

\subsection{Unconditional 3D human synthesis.}
More recently, neural fields and inverse rendering techniques allow us to train 3D GANs with only 2D images~\cite{chan2021pi,niemeyer2021giraffe,chan2022efficient}. These 3D GANs are extended to articulated full-body humans using warping based on linear blend skinning ~\cite{EVA3D,zhang2022avatargen,bergman2022gnarf}. 
By applying inversion~\cite{roich2022pivotal}, these methods can generate a 360-degree rendering of a clothed human from a single image. 
While these results are 3D consistent, we observe that they are plausible only for relatively simple clothing and degrade for more complex texture patterns. 
Achieving photorealistic and generalizable 3D human digitization with 3D GANs remains an open problem. 
Our work achieves better generalization and photorealism by incorporating more general yet highly expressive image priors from diffusion models. 

\subsection{3D human reconstruction from a single image.}
3D reconstruction of clothed humans from a single image is a long-standing problem. 
A parametric body model~\cite{loper2015smpl} provides strong prior about the underlying shape of a person, but only for minimally clothed bodies~\cite{kanazawa2018end,lassner2017unite,pavlakos2018learning,kolotouros2019learning}. 
To enable clothed human reconstruction, regression-based 3D reconstruction has been extended to various shape representations such as voxels~\cite{varol2018bodynet}, mesh displacements~\cite{alldieck2019learning,alldieck2019tex2shape,bhatnagar2019multi}, silhouettes~\cite{natsume2019siclope}, depth maps~\cite{gabeur2019moulding,wang2020normalgan}, and neural fields~\cite{xie2022neural,saito2019pifu,saito2020pifuhd,huang2020arch,he2021arch++,xiu2022icon,xiu2023econ,corona2021smplicit,smith2019facsimile}. 
Among them, several works also support texture synthesis for the occluded regions. 
SiCloPe~\cite{natsume2019siclope} shows that an image-to-image translation network in screen space can infer occluded textures. 
PIFu~\cite{saito2019pifu} infers continuous texture fields~\cite{oechsle2019texture} in 3D, which is later improved by explicitly modeling reflectances~\cite{alldieck2022phorhum}. 
These approaches, however, often fail to produce photorealistic textures for the back side due to the limited 3D scan data for supervised training. 
Differentiable rendering based on NeRFs~\cite{mildenhall2020nerf} has also been applied to learn 3D human representations from images.
Both person-specific models~\cite{peng2021neural,liu2021neural,weng_humannerf_2022_cvpr} and generalizable models across identities~\cite{kwon2021neural,choi2022mononhr,mihajlovic2022keypointnerf,huang2022one,SHERF,MPS-NeRF} have been proposed, but the training requires multi-view images or videos. 
They are difficult to collect at scale such that the collected data covers a sufficient span of clothing types and textures. 
Our approach, on the other hand, does not require multi-view images or person-specific video capture.

\subsection{Diffusion models for 3D synthesis.} 
Denoising diffusion models have shown impressive image synthesis results. 
These powerful 2D generative models are recently adopted to learn 3D scene representations. 
Recent methods~\cite{poole2022dreamfusion, lin2022magic3d, metzer2022latent, wang2023prolificdreamer, Chen_2023_ICCV, sjc} have shown that text-to-image models can be repurposed for 3D object generation from text input with remarkable results. 
Unlike these methods, our method is conditioned on a human input image to create a 3D consistent texture of the person, where the results are photorealistic. 
Diffusion models can be customized for a specific subject, but this customization typically requires multiple images and a considerable amount of time~\cite{ruiz2022dreambooth, gal2022textual}. Moreover, such methods may not consistently maintain the subject's appearance details (i.e. clothing, hairstyle, facial expression, etc.)~\cite{e4t}.
These customization methods can be utilized to generate 3D objects conditioned on a single image ~\cite{Xu_2022_neuralLift, qian2023magic123}. 
Unlike these customization methods, our method can generate 3D textured human models without test-time finetuning.
Moreover, current image-to-3D techniques~\cite{Xu_2022_neuralLift, qian2023magic123,tang2023make} lack human-specific prior and hence struggle to synthesize realistic and detailed textured human models.
The closest to our work is TEXTure~\cite{richardson2023texture}, which utilizes 2D diffusion models to synthesize texture of an input mesh. 
We observe that their shape guidance based on depth maps is insufficient for photorealistic clothed human synthesis. 
Instead of progressively refining the texture based on viewing angles, we improve the consistency by blending the RGB color of existing views, weighted by visibility, viewing angles, and distance to missing regions. 
We also improve the per-view synthesis by incorporating normal and silhouette maps as guidance signals.

\begin{figure*}[t]    
\centering
\includegraphics[width=0.95\linewidth]{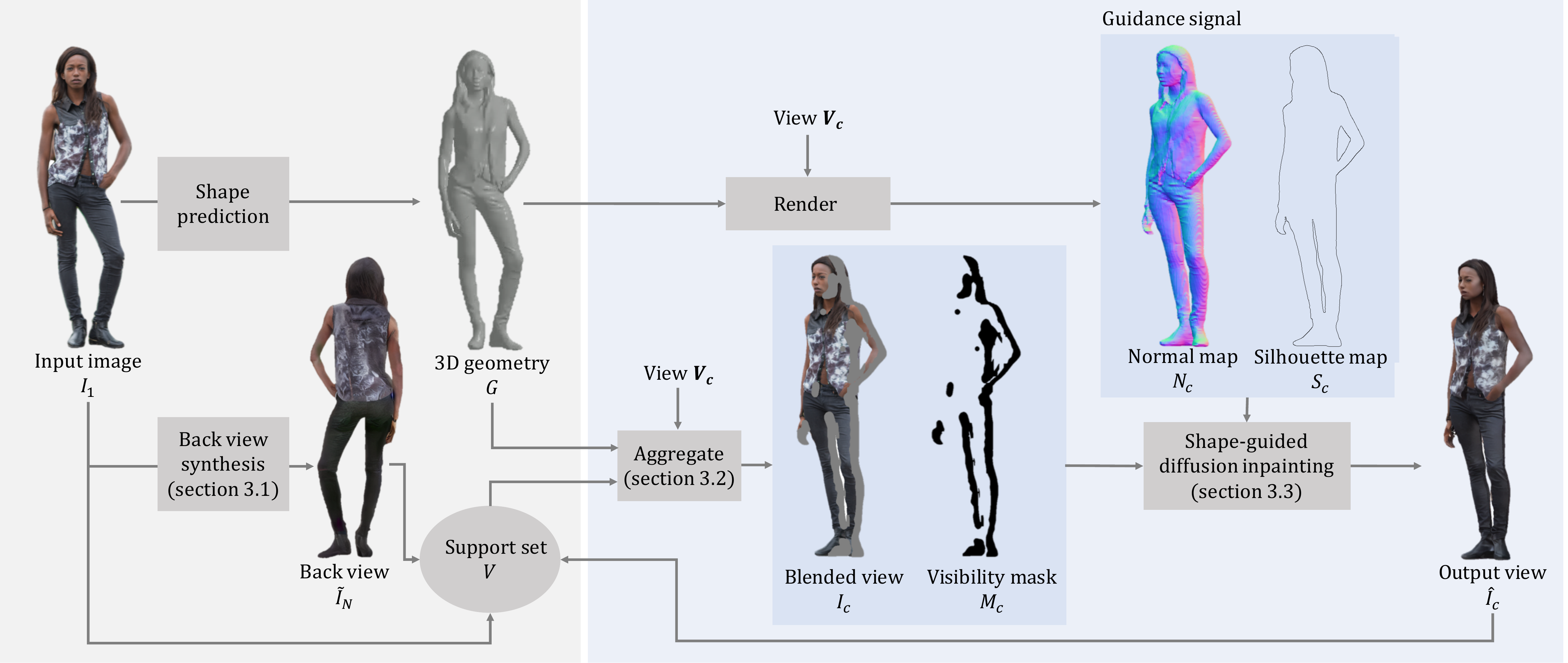}
\mpage{0.35}{{Initialization}}\hfill
\mpage{0.55}{{Shape-guided diffusion}}\\
\vspace{-1mm}
\caption{\textbf{Person image generation with shape-guided diffusion.}
To generate a 360-degree view of a person from a \emph{single} image $I_1$, we first synthesize multi-view images of the person. 
We use an off-the-shelf method to infer the 3D geometry~\shortcite{saito2020pifuhd} and synthesize an initial back-view $\tilde{I}_N$ of the person~\shortcite{albahar2021pose} as a guidance.
We add our input view $I_1$ and the synthesized initial back-view $\tilde{I}_N$ to our support set $V$.
To generate a new view $V_c$, we aggregate all the visible pixels from our support set $V$ 
by blending their RGB color, weighted by visibility, viewing angles, and the distance to missing regions. 
To hallucinate the unseen appearance and synthesize view $\hat{I}_c$, we use a pretrained inpainting diffusion model guided by shape cues (normal $N_c$ and silhouette $S_c$ maps). 
We include the generated view $\hat{I}_c$ in our support set and repeat this process for all the remaining views. Images from Adobe Stock.
}
\label{fig:multiviewsyn}
\end{figure*}

\section{Method}
\label{sec:method}

Our goal is to generate a 360-degree view of a person with a consistent, high-resolution appearance from a \emph{single} input image.
To this end, we first synthesize a set of multi-view images of the person $\{\hat{I}_2, ..., \hat{I}_N\}$ that are consistent among each other and coherent with the input image $I_{1}$ (\figref{multiviewsyn}).
In particular, we use the reconstructed 3D geometry of the person to guide the inpainting with diffusion models (\figref{shapeguidance}). 
For 3D shape reconstruction, we employ an off-the-shelf method~\cite{saito2020pifuhd} to obtain a triangular mesh $G$ of the input person using Marching cubes~\cite{lorensen1987marching}.

We synthesize the multi-view images in an \emph{auto-regressive} manner. 
More specifically, we start with synthesizing the back-view of the person with~\cite{albahar2021pose} (\secref{backview}). 
The input and the synthesized back-view images form an initial \emph{support set} $V$ (i.e., currently available views).
Using the images from the support set and the mesh $G$, we can render a new view of the person (\secref{aggregate}).
Here, this blended view is consistent with the previously generated images but may have missing regions (that are not covered by any of the images in the support set).
We use a shape-guided diffusion model to inpaint the appearance details while respecting the estimated shape (\secref{inpaint}).
We expand the support set by adding this inpainted view and proceed to a new view until all the views are generated. 
We sample views at intervals of $45^\circ$, specifically in the order of [$45^\circ$, $-45^\circ$, $90^\circ$, $-90^\circ$, $135^\circ$, $-135^\circ$, $180^\circ$]. Thus, our support set will have a total of 8 views $(N=8)$.
When we use more viewpoints, the missing regions become very small. In such cases, we found that the inpainting performance deteriorates. On the other hand, when we use less viewpoints, the missing regions become very large. We found that the inpainting fails to preserve the input appearance.

We then fuse these multi-view images $\{I_{1}, \hat{I}_2, ..., \hat{I}_N\}$ via inverse rendering robust to slight misalignment and optimize a UV texture map $T$ (\figref{multiviewfusion}). 
We finally use this UV texture map $T$ to render the 360-degree view of the person.
Note that our approach assumes weak perspective projection for simplicity, following ~\cite{saito2019pifu,saito2020pifuhd,xiu2022icon}, but extending it to a perspective camera is also possible.

\subsection{Back-view Synthesis}
\label{sec:backview}
The input frontal and back views have strong semantics correlations (e.g., the back side of a T-shirt is likely a T-shirt with similar textures), and its silhouette contour provides structural guidance.
Thus, we first synthesize the back-view of the person for guidance \emph{prior} to synthesizing other views. 
While prior works~\cite{natsume2019siclope,he2021arch++} show that front-to-back synthesis is highly effective with supervised training, our approach achieves the front-to-back synthesis without relying on ground-truth paired data. 
More specifically, we apply the SoTA 2D human synthesis method~\cite{albahar2021pose} with the inferred dense pose prediction for the back-view. 
To generate a dense pose prediction that aligns precisely with the input image, we render the surface normal and depth map of the shape $G$ from the view opposite to the input view and create a photorealistic back-view using ControlNet~\cite{zhang2023adding} with the text prompt of 
\textit{``back view of a person wearing nice clothes in front of a solid gray background, best quality.''} 
We then run dense pose~\cite{guler2018densepose}, which is finally fed into Pose-with-Style~\cite{albahar2021pose}. 
We empirically find that using Pose-with-Style~\cite{albahar2021pose} with the aforementioned procedure leads to a more semantically consistent back-view than just using ControlNet~\cite{zhang2023adding}.
See~\figref{nobackview} for the impact of the back-view initialization.

\subsection{Multi-view visible texture aggregation}
\label{sec:aggregate}

Prior to inpainting, we aggregate all the views in the support set $V$ to the target view $V_c$. However, naively averaging all views leads to a blurry image due to slight misalignment in each view. 
To ensure that high-resolution details are all retained, we use weighted averaging using confidence based on visibility, viewing angles, and distance.

For each view $V_v$ in the set of synthesized views $V_v$, we render the normal map $N_{v}^{c}$ as well as its color $C_{v}^{c}$ from $V_c$.
In addition, we set the visibility mask $M_{v}$ of each view $V_v$ by comparing its visible faces to the visible faces from $V_c$.
We use this visibility mask $M_{v}$ to compute distance transform $d_{v}$ from the boundary of the visible pixels and the invisible pixels in each view $V_v$.
We also compute the angular difference $\phi_{v}$ of each visible pixel between view $V_v$ and the current view of interest $V_c$ as follows:
\begin{equation}\label{eqn:angle}
    \phi_{v} = M_{v} \arccos \left( \frac{N_{v}^{c} \cdot N_{c}}{\mathrm{max} \left( ||N_{v}^{c}||_2 \cdot ||N_{c}||_2, \epsilon \right)} \right),
\end{equation}
where $\epsilon=10^{-8}$ is a small value to avoid dividing by zero. 

Finally, we compute the blending weight $w_{v}$ of view $V_v$ as follows:
\begin{equation}\label{eqn:weight}
    w_{v} = \frac{M_{v} B_{v} e^{-\alpha \phi_{v}} d_{v}^{\beta}} {\sum_{i \in V} M_{i} B_{i} e^{-\alpha \phi_{i}} d_{i}^{\beta} + \epsilon}.
\end{equation}
In our experiments, we set both $\alpha$, which determines the strength of the angular difference, and $\beta$, which determines the strength of the Euclidean distance, to 3. 
Using the angular difference $\phi_{v}$ ensures a higher weight to closer views, while using the Euclidean distance $d_{v}$ ensures a lower weight for pixels close to the missing region.
Moreover, if only one existing view contains a specific pixel, we mark its boundary $B_{v}$ as invisible.
This ensures that the target view does not suffer from boundary artifacts.

We use the computed weights $w_{v}$ to blend the color $C_{v}$ of the previously synthesized views $V_v$ together, where the blended image of the current view $I_{c}$ and its visibility mask $M_{c}$ are as follows:
\begin{equation}%
    \begin{split}
        M_{c} = \bigcup_{i \in V} M_{i}, \quad\text{and}\quad
        I_{c} = \sum_{i \in V} w_{i} C_{i}. \\ 
    \end{split}
\end{equation}
The final blended image $I_{c}$ and its visibility mask $M_{c}$ are then used to synthesize a complete view $\hat{I}_{c}$ using our shape-guided diffusion.

\subsection{Shape-guided diffusion inpainting}
\label{sec:inpaint}

\begin{figure}[t]
\centering

\mpage{0.15}{\includegraphics[width=\linewidth, trim=45 0 25 0, clip]{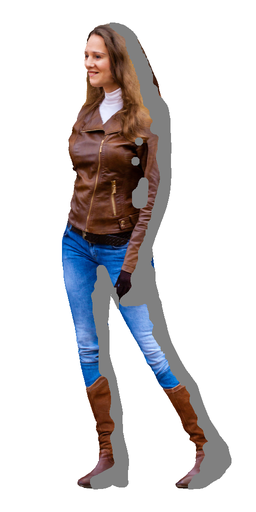}}\hfill
\mpage{0.15}{\includegraphics[width=\linewidth, trim=45 0 25 0, clip]{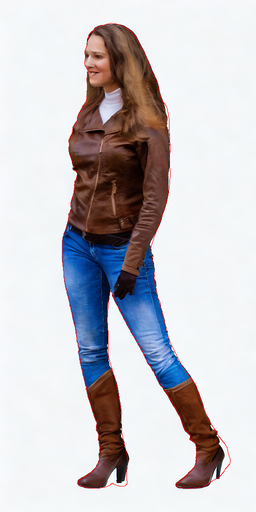}}\hfill
\mpage{0.15}{\includegraphics[width=\linewidth, trim=45 0 25 0, clip]{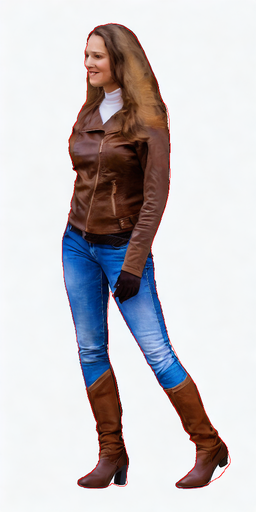}}\hfill
\mpage{0.15}{\includegraphics[width=\linewidth, trim=45 0 25 0, clip]{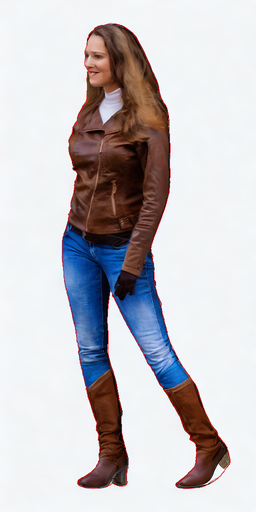}}\hfill
\mpage{0.15}{\includegraphics[width=\linewidth, trim=45 0 25 0, clip]{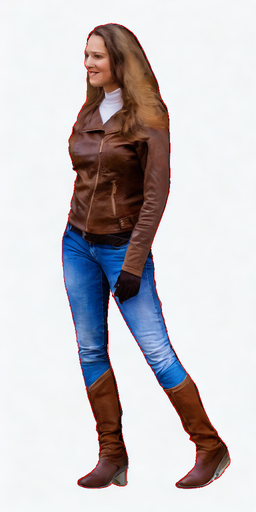}}\\

\mpage{0.16}{\footnotesize{Input}}\hfill
\mpage{0.16}{\footnotesize{(a) No guidance}}\hfill
\mpage{0.16}{\footnotesize{(b) Normal}}\hfill
\mpage{0.16}{\footnotesize{(c) Silhouette}}\hfill
\mpage{0.16}{\footnotesize{(d) Normal and silhouette}}\\
\vspace{-2mm}

\captionof{figure}{
\textbf{Shape-guided diffusion inpainting.}
To synthesize the unseen appearance in a new view, we use a pretrained inpainting diffusion model.
With no guidance, the inpainted regions often do not preserve the shape (red silhouette) nor the structural details of the 3D geometry (a).
If we use normal maps as a control signal for ControlNet~\shortcite{zhang2023adding} (b), the inpainted region preserves the structural details of the mesh (e.g., fingers), but not the shape of the human body.
Using the silhouette map preserves the shape of the human body, but not the structural details of the mesh (c).
We propose to use both normal and silhouette maps to guide the inpainting model to respect the underlying 3D geometry (d). Images from Adobe Stock.
}
\label{fig:shapeguidance}
\end{figure}

To synthesize the unseen appearance indicated by the visibility mask $M_{c}$ in the blended image $I_{c}$, we use a 2D inpainting diffusion model~\cite{rombach2021highresolution}. 
However, we observe that without any guidance, the inpainted regions often do not respect the underlying geometry $G$ (see \figref{shapeguidance}(a)).
To address this, we use the method of ControlNet~\cite{zhang2023adding} by incorporating additional structural information into the diffusion model.
When we use normal maps as a control signal, we can preserve the structural details of the mesh but not the shape of the human body (\figref{shapeguidance}(b)).
On the other hand, using the silhouette map alone preserves the shape of the human body, but not the structural details of the mesh (\figref{shapeguidance}(c)).
To best guide the inpainting model to respect the underlying 3D geometry, we propose to use both normal map and silhouette maps, 
as shown in~ \figref{shapeguidance}(d).
We add this generated view to our support set $V$ and proceed to the next view until all $N$ views are synthesized.

\subsection{Multi-view fusion}
\begin{figure}[t]    
\centering
\includegraphics[width=\linewidth]{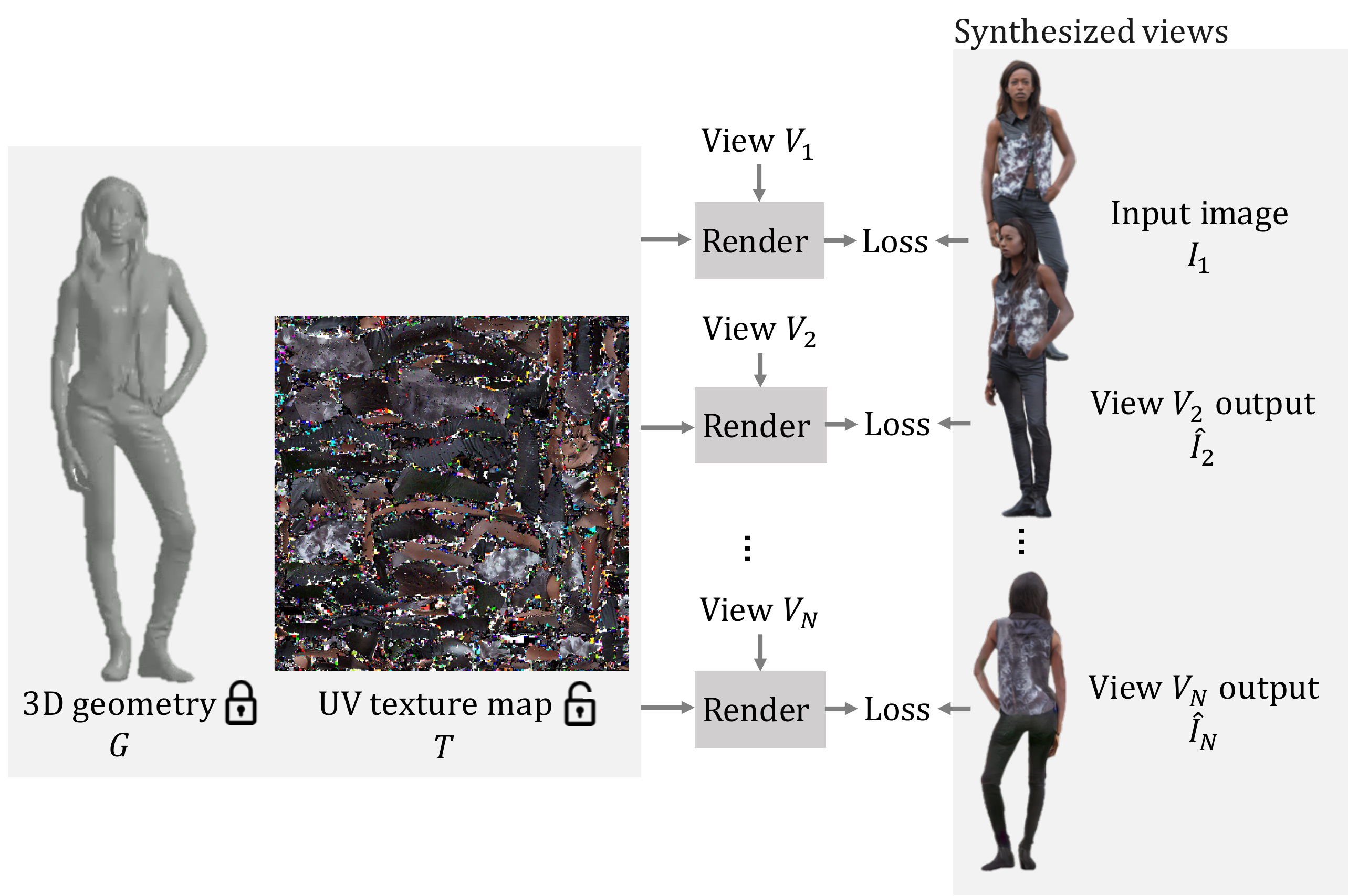}
\vspace{-5mm}
\caption{\textbf{Multi-view fusion.}
We fuse the synthesized multi-view images $\{I_1, \hat{I}_2, ..., \hat{I}_N\}$ (see~\figref{multiviewsyn}) to obtain a textured 3D human mesh.
We use the computed UV parameterization~\shortcite{xatlas} to optimize a UV texture map $T$ with the geometry $G$ fixed. 
In each iteration, we differentiably render the UV texture map $T$ in every synthesized view from our set of views $\{V = V_1, V_1, ..., V_N\}$. 
We minimize the reconstruction loss between the rendered view and our synthesized view using both LPIPS loss~\shortcite{zhang2018perceptual} and L1 loss.
The fusion results in a textured mesh that can be rendered from any view. Images from Adobe Stock.
}
\label{fig:multiviewfusion}
\end{figure}

Since the latent diffusion model operates inpainting in the low-resolution latent space, the final synthesized images do not form geometrically consistent multi-view images. 
Therefore, we consolidate these slightly misaligned multi-view images $I_{1}, \hat{I}_{2}, ..., \hat{I}_{N}\}$ into a single consistent 3D texture map $T$. 
We show the overview of our multi-view fusion in ~\figref{multiviewfusion}.

We first compute the UV parameterization of the reconstructed 3D geometry using xatlas~\cite{xatlas}. 
Then, we optimize a UV texture map $T$ via inverse rendering with loss functions that are robust to small misalignment.
In every iteration, we render the UV texture map $T$ in each view $i$ from our set of synthesized views $\{V = V_1, V_1, ..., V_N\}$ and minimize the reconstruction loss of this rendered view and the synthesized view using both LPIPS loss~\cite{zhang2018perceptual} and L1 loss such that:
\begin{equation}%
    L(T) = \sum_{i \in V} L_\mathrm{\textsc{lpips}}\left(Render(T; G, i), \hat{I}_{i}\right) + \lambda L_1\left(Render(T; G, i), \hat{I}_{i}\right),
\end{equation}
where $\hat{I}_{1}=I_{1}$ and $\lambda$ is set to 10.

Once the texture map $T$ is optimized, one can render the textured mesh from arbitrary viewpoints.

\section{Experimental Results}
\begin{figure*}[t]
\centering

\mpage{0.08}{\includegraphics[width=\linewidth, trim=130 0 130 0, clip]{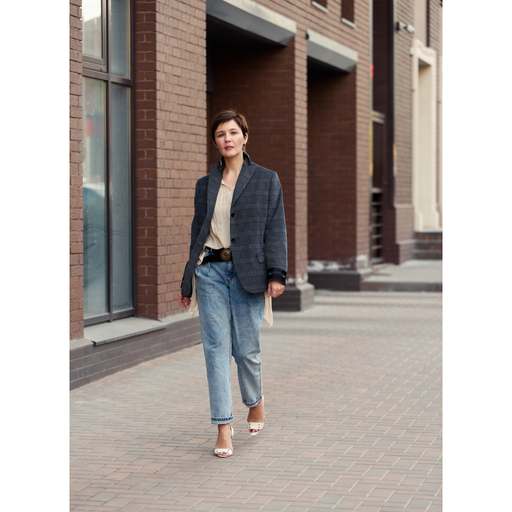}}\hfill
\mpage{0.05}{\includegraphics[width=\linewidth, trim=195 0 155 0, clip]{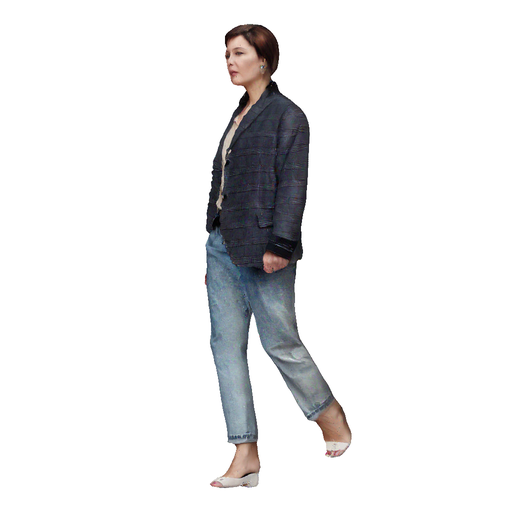}}\hfill
\mpage{0.05}{\includegraphics[width=\linewidth, trim=175 0 175 0, clip]{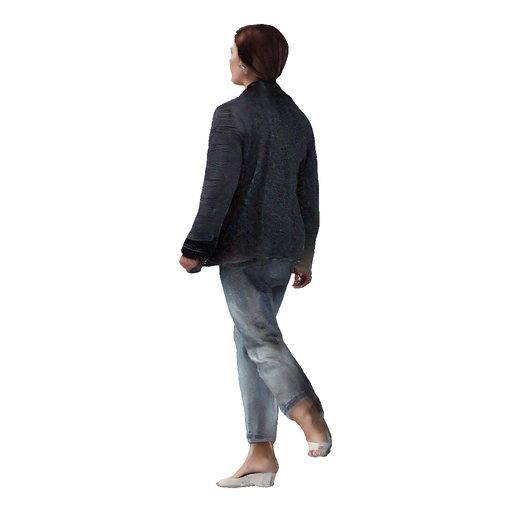}}\hfill
\mpage{0.05}{\includegraphics[width=\linewidth, trim=175 0 175 0, clip]{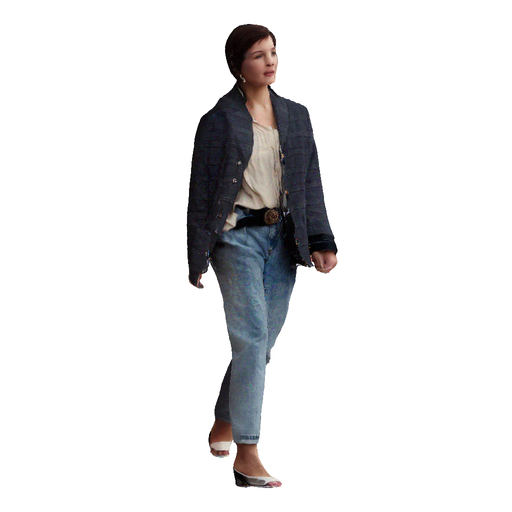}}\hfill
\mpage{0.08}{\includegraphics[width=\linewidth, trim=130 0 130 0, clip]{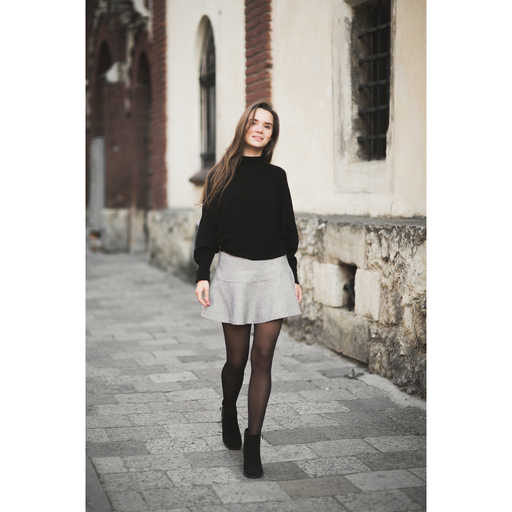}}\hfill
\mpage{0.05}{\includegraphics[width=\linewidth, trim=195 0 155 0, clip]{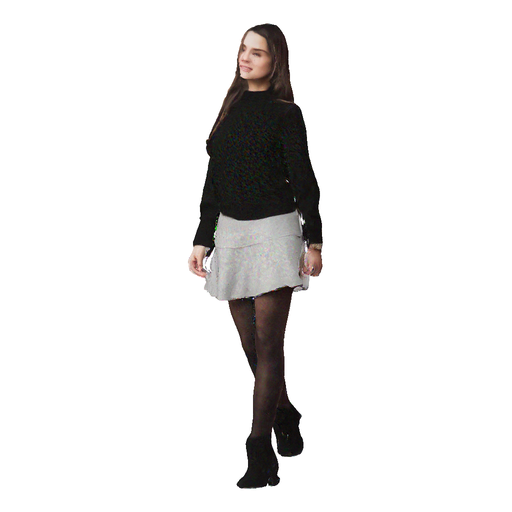}}\hfill
\mpage{0.05}{\includegraphics[width=\linewidth, trim=175 0 175 0, clip]{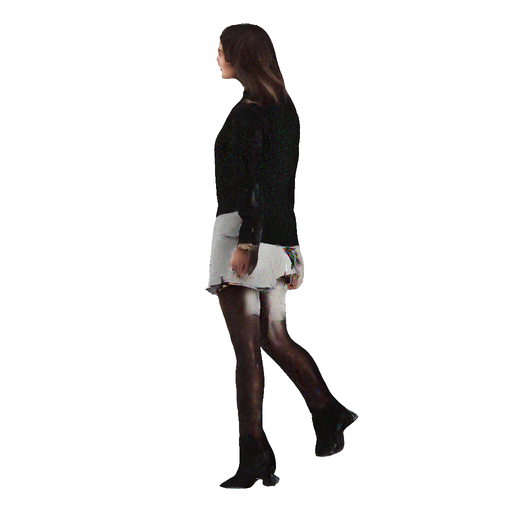}}\hfill
\mpage{0.05}{\includegraphics[width=\linewidth, trim=175 0 175 0, clip]{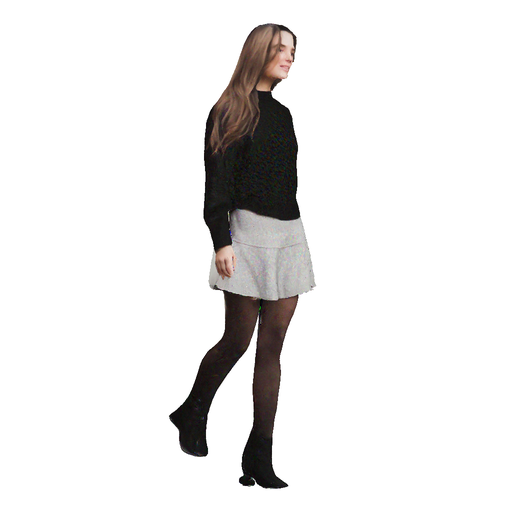}}\hfill
\mpage{0.08}{\includegraphics[width=\linewidth, trim=130 0 130 0, clip]{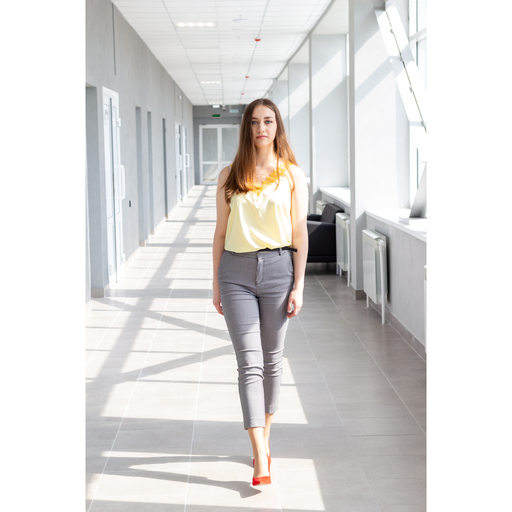}}\hfill
\mpage{0.05}{\includegraphics[width=\linewidth, trim=175 0 175 0, clip]{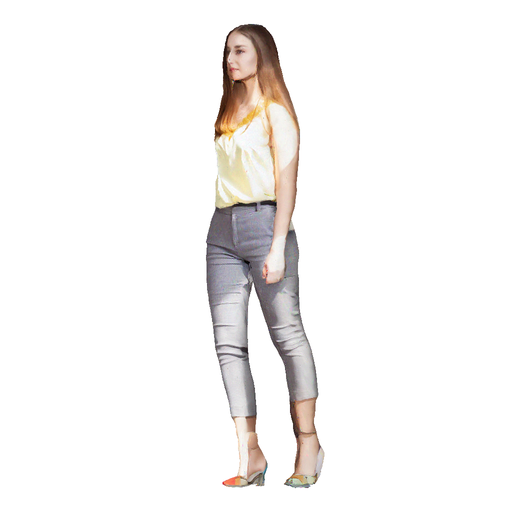}}\hfill
\mpage{0.05}{\includegraphics[width=\linewidth, trim=175 0 175 0, clip]{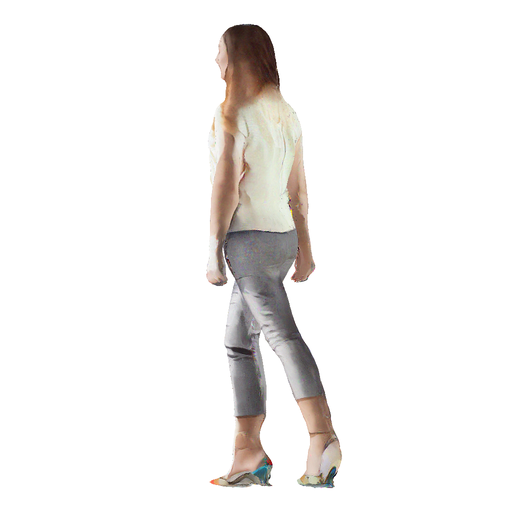}}\hfill
\mpage{0.05}{\includegraphics[width=\linewidth, trim=175 0 175 0, clip]{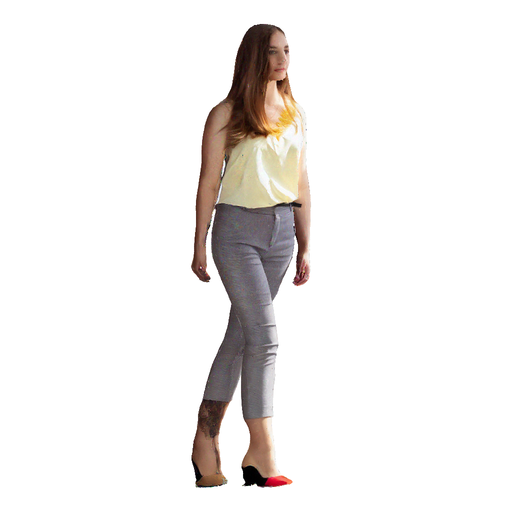}}\\

\mpage{0.08}{\small{Input}}\hfill
\mpage{0.22}{$\underbrace{\hspace{\textwidth}}_{\substack{\vspace{-5.0mm}\\\colorbox{white}{~~360$^\circ$ generation~~}}}$}\hfill
\mpage{0.08}{\small{Input}}\hfill
\mpage{0.22}{$\underbrace{\hspace{\textwidth}}_{\substack{\vspace{-5.0mm}\\\colorbox{white}{~~360$^\circ$ generation~~}}}$}\hfill
\mpage{0.08}{\small{Input}}\hfill
\mpage{0.22}{$\underbrace{\hspace{\textwidth}}_{\substack{\vspace{-5.0mm}\\\colorbox{white}{~~360$^\circ$ generation~~}}}$}\\
\vspace{-3mm}
\captionof{figure}{
\textbf{Limitations.}
Our approach inherits limitations from existing methods for shape reconstruction (unusual foot shape (left)) and back-view synthesis (misaligned skirt length due to lack of geometry awareness (middle)).
We also show the baked specularity on the face and garment texture, which is ideally view-dependent (right). Images from Adobe Stock.
}
\label{fig:limitation}
\end{figure*}

\begin{figure}[t]
\centering

\mpage{0.12}{\includegraphics[width=\linewidth, trim=360 0 340 0, clip]{images/AdobeStock/input/629843553.png}}\hfill
\hspace{1mm}
\mpage{0.12}{\includegraphics[width=\linewidth, trim=180 0 180 0, clip]{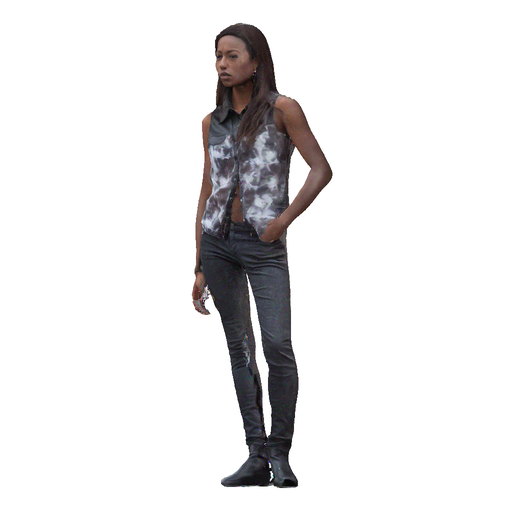}}\hfill
\mpage{0.12}{\includegraphics[width=\linewidth, trim=180 0 180 0, clip]{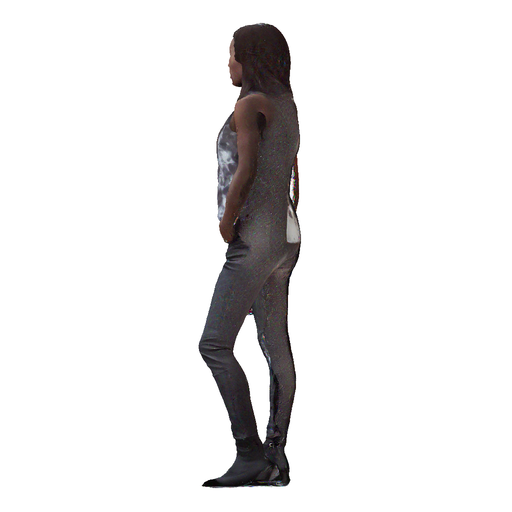}}\hfill
\mpage{0.12}{\includegraphics[width=\linewidth, trim=180 0 180 0, clip]{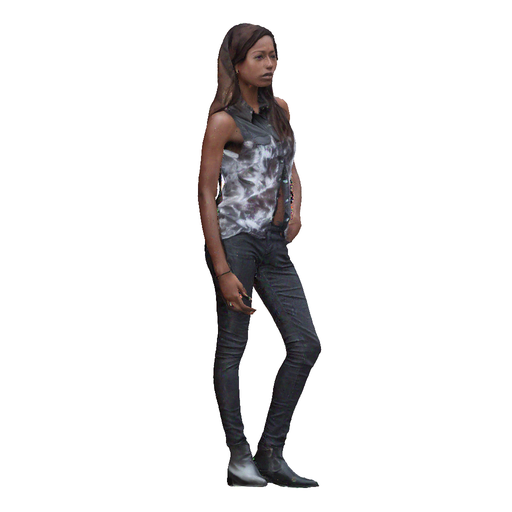}}\hfill
\hspace{1mm}
\mpage{0.12}{\includegraphics[width=\linewidth, trim=360 0 360 0, clip]{images/AdobeStock/ours/629843553/view_010.png}}\hfill
\mpage{0.12}{\includegraphics[width=\linewidth, trim=360 0 360 0, clip]{images/AdobeStock/ours/629843553/view_030.png}}\hfill
\mpage{0.12}{\includegraphics[width=\linewidth, trim=360 0 360 0, clip]{images/AdobeStock/ours/629843553/view_080.png}}\hfill
\\

\vspace{-2mm}
\mpage{0.12}{\small{Input}}\hfill
\hspace{1mm}
\mpage{0.4}{$\underbrace{\hspace{\textwidth}}_{\substack{\vspace{-5.0mm}\\\colorbox{white}{~~Without back-view~~}}}$}\hfill
\hspace{1mm}
\mpage{0.4}{$\underbrace{\hspace{\textwidth}}_{\substack{\vspace{-5.0mm}\\\colorbox{white}{~~Ours~~}}}$}\\
\vspace{-2mm}
\caption{\textbf{The need of back-view synthesis.}
Having an initial back-view encourages all other views to preserve the appearance of the person in the input image especially when a target view is far from the input view. Images from Adobe Stock.
}
\label{fig:nobackview}
\end{figure}

\subsection{Experimental Setup}

\subsubsection{Implementation details.}
We implement our approach with PyTorch on a single RTX A6000 GPU. 
We set the guidance scale of the pretrained inpainting diffusion model to 15 and the number of inference steps per view to 25.
In all our experiments, we use a generic text prompt for all subjects: \textit{``a person wearing nice clothes in front of a solid white background, <VIEW> view, best quality, extremely detailed",} where <VIEW> is set to ``front'' for frontal views; ``left'' and ``right'' for $45^\circ$ and $-45^\circ$ views, respectively; ``side'' for $\pm 90^\circ$ views; and ``back'' for the rest of viewing angles ($\pm 135^\circ$ and $180^\circ$).
We use the ADAM optimizer with a learning rate of $0.1$ and with $\beta_1=0.9$ and $\beta_2=0.999$ to learn the UV texture map $T$.
The entire process of generating a 3D textured model from a single image takes approximately 7 minutes on an RTX A6000 GPU.

\subsubsection{Datasets.} 
To evaluate our approach, we utilize the THuman2.0 dataset~\cite{tao2021function4d}, using 30 subjects, evenly split between 15 males and 15 females. We use front-facing images as input. We also evaluate our approach on the DeepFashion dataset~\cite{liuLQWTcvpr16DeepFashion} to compare with ELICIT~\cite{huang2022one}. We additionally use in-the-wild images from Adobe Stock\footnote{https://stock.adobe.com/} to showcase results from images with diverse subjects, clothing, and poses.\footnote{All datasets used in this research were exclusively downloaded, accessed, and utilized on UMD clusters.}

\subsubsection{Baselines.}
We compare our 360-degree view synthesis approach with Pose with Style (PwS) baseline. We use Pose with Style~\cite{albahar2021pose} to generate multi-view images and then fuse them using our multi-view fusion. We also compare with PIFu~\cite{saito2019pifu}, Impersonator++~\cite{liu2021liquid}, TEXTure~\cite{richardson2023texture}, Magic123~\cite{qian2023magic123}, and S3F~\cite{corona2022structured}. To make TEXTure~\cite{richardson2023texture} conditional on an input image, we use the input image directly instead of generating an initial view from the depth-to-image diffusion model.
We also compare our work with ELICIT~\cite{huang2022one} on a subset of the DeepFashion dataset~\cite{liuLQWTcvpr16DeepFashion} provided by its authors.

\subsection{Quantitative Comparison}
\label{sec:quanitative}
\begin{table}[t]\setlength{\tabcolsep}{4pt}
	\centering%
\caption{
\textbf{Quantitative comparisons with baseline methods on the THuman2.0 dataset~\cite{tao2021function4d}.}
}
\small
\vspace{-2mm}

	\begin{tabular}{lccccc}
		\toprule
    	   Methods          	         & PSNR$\uparrow$ & SSIM$\uparrow$ &  FID$\downarrow$ &  LPIPS$\downarrow$ & CLIP-score$\uparrow$   \\
            \midrule
            PwS baseline  & \underline{17.8003} & 0.8888 & 132.4511 & \underline{0.1320} & 0.7733 \\
            PIFu & \textbf{18.0934} & \textbf{0.9117} & 150.6622 & 0.1372 & 0.7721 \\
  Impersonator++ & 16.4791 & \underline{0.9012} & \textbf{106.5753} & 0.1468 & \textbf{0.8168} \\
  TEXTure& 16.7869 & 0.8740 & 215.7078 & 0.1435 & 0.7272 \\
  Magic123& 14.5013 & 0.8768 & 137.1108 & 0.1880 & \underline{0.7996} \\
  S3F& 14.1212 & 0.8840 & 165.9806 & 0.1868 & 0.7475 \\
            \emph{Ours}               & 17.3651 & 0.8946 & \underline{115.9918} & \textbf{0.1300} & 0.7992    \\ 
		\bottomrule
		\label{tab:thuman}
	\end{tabular}
\end{table}

To quantify the quality of our results, we measure peak signal-to-noise ratio (PSNR), structural similarity index measure (SSIM), Frechet Inception Distance (FID)~\cite{parmar2021cleanfid}, learned perceptual image patch similarity (LPIPS)~\cite{zhang2018perceptual}, and CLIP-score. CLIP-score measures the cosine similarity between the CLIP embeddings of an input image and each of the synthesized views. We use a total of 90 synthesized views with 4$^\circ$ spacing. 
We compare these metrics on the THuman2.0 dataset~\cite{tao2021function4d} with other baselines in Table~\ref{tab:thuman}. Quantitative results show that existing metrics are not consistent in evaluating 3D textured humans. PSNR favors blurry images as in PIFu~\cite{saito2019pifu}, and FID does not provide accurate results for sparse view distributions. 
To quantitatively compare with ELICIT~\cite{huang2022one}, we compute the CLIP-score (where higher values indicate better performance) on their provided subset of the DeepFashion dataset~\cite{liuLQWTcvpr16DeepFashion}. Our method achieved a CLIP-score of 0.7732, surpassing their score of 0.7236.

\subsection{Qualitative Comparison}
\label{sec:visual_comparison}
\begin{figure*}[t]
\centering

\mpage{0.03}{\raisebox{0pt}{\rotatebox{90}{Input}}}  \hfill
\mpage{0.05}{}\hfill
\mpage{0.088}{\includegraphics[width=\linewidth, trim=125 0 125 0, clip]{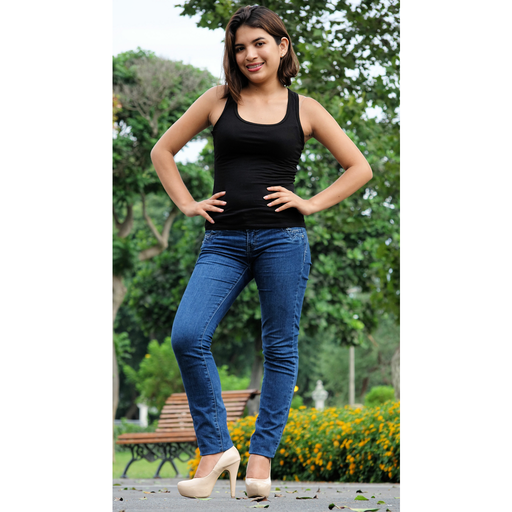}}\hfill
\mpage{0.05}{}\hfill
\mpage{0.05}{}\hfill
\mpage{0.07}{\includegraphics[width=\linewidth, trim=170 0 130 0, clip]{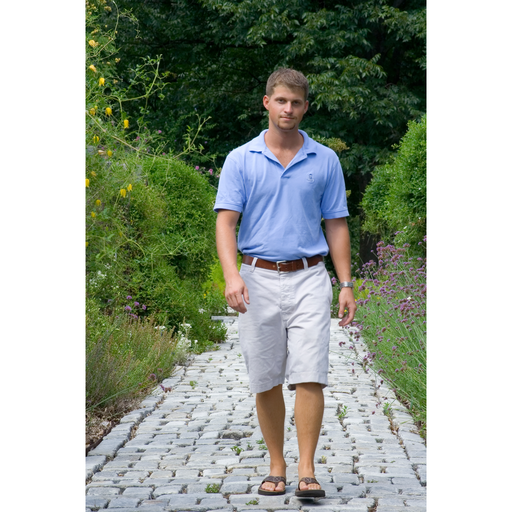}}\hfill
\mpage{0.05}{}\hfill
\mpage{0.05}{}\hfill
\mpage{0.07}{\includegraphics[width=\linewidth, trim=150 0 150 0, clip]{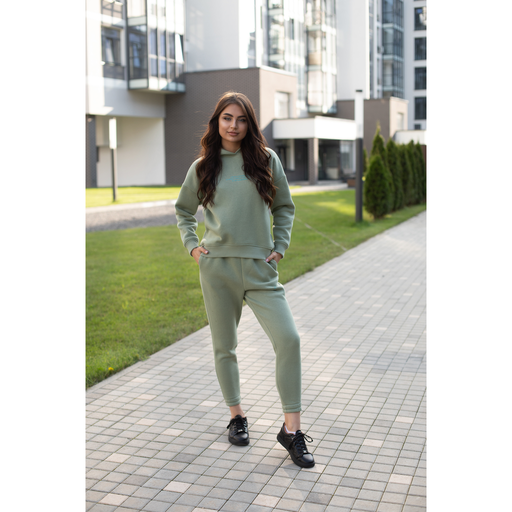}}\hfill
\mpage{0.05}{}\hfill
\mpage{0.05}{}\hfill
\mpage{0.07}{\includegraphics[width=\linewidth, trim=150 0 150 0, clip]{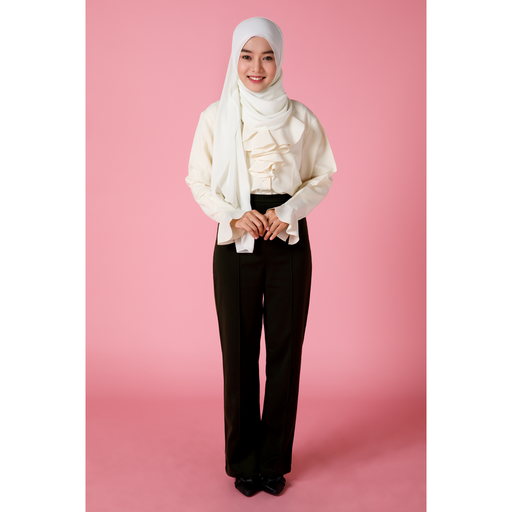}}\hfill
\mpage{0.05}{}\hfill
\mpage{0.05}{}\hfill
\mpage{0.07}{\includegraphics[width=\linewidth, trim=150 0 150 0, clip]{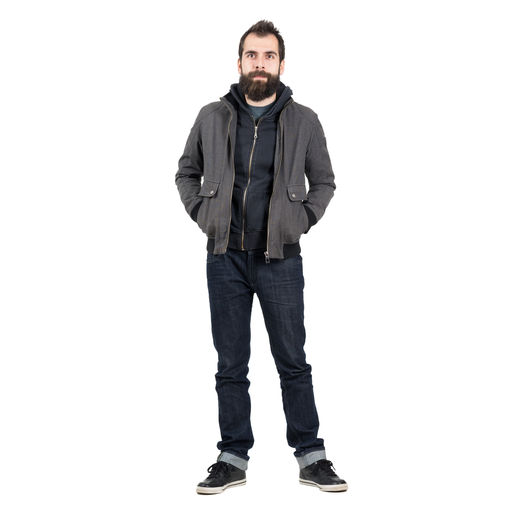}}\hfill
\mpage{0.05}{}\\

\mpage{0.03}{\raisebox{0pt}{\rotatebox{90}{PwS baseline~\shortcite{albahar2021pose}}}}  \hfill
\mpage{0.05}{\includegraphics[width=\linewidth, trim=175 0 175 0, clip]{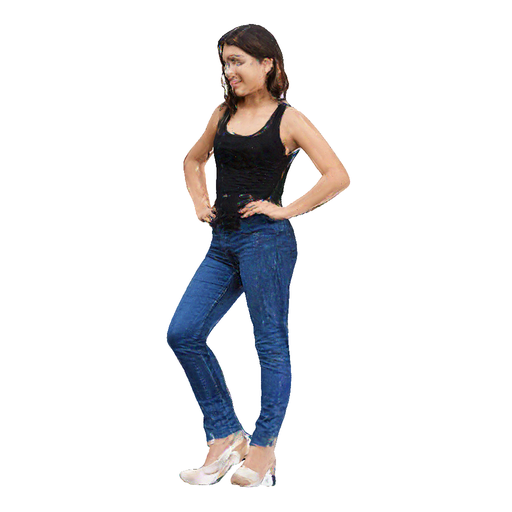}}\hfill
\mpage{0.05}{\includegraphics[width=\linewidth, trim=175 0 175 0, clip]{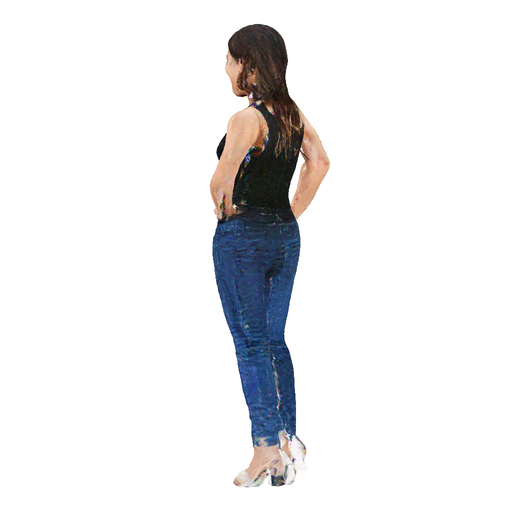}}\hfill
\mpage{0.05}{\includegraphics[width=\linewidth, trim=175 0 175 0, clip]{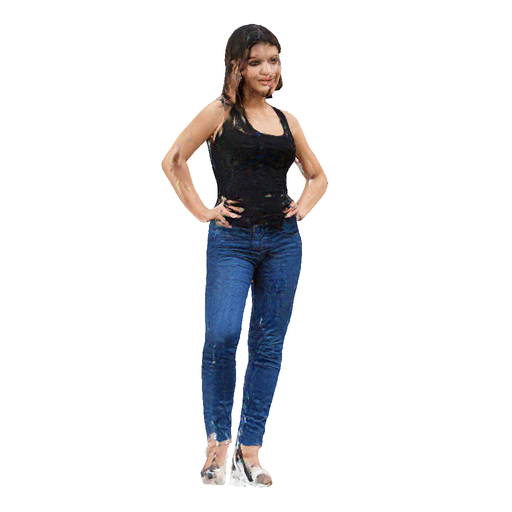}}\hfill
\hspace{2mm}
\mpage{0.05}{\includegraphics[width=\linewidth, trim=175 0 175 0, clip]{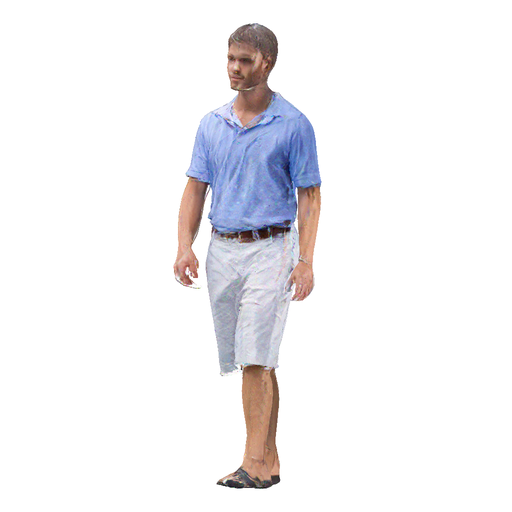}}\hfill
\mpage{0.05}{\includegraphics[width=\linewidth, trim=175 0 175 0, clip]{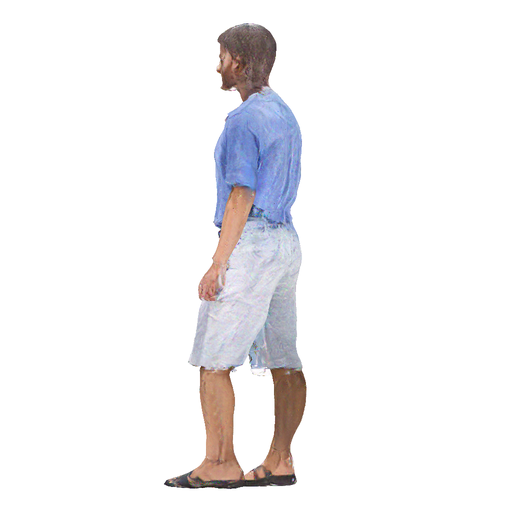}}\hfill
\mpage{0.05}{\includegraphics[width=\linewidth, trim=175 0 175 0, clip]{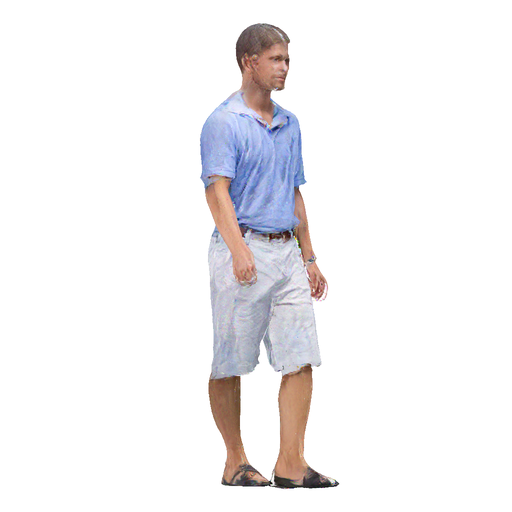}}\hfill
\hspace{2mm}
\mpage{0.05}{\includegraphics[width=\linewidth, trim=175 0 175 0, clip]{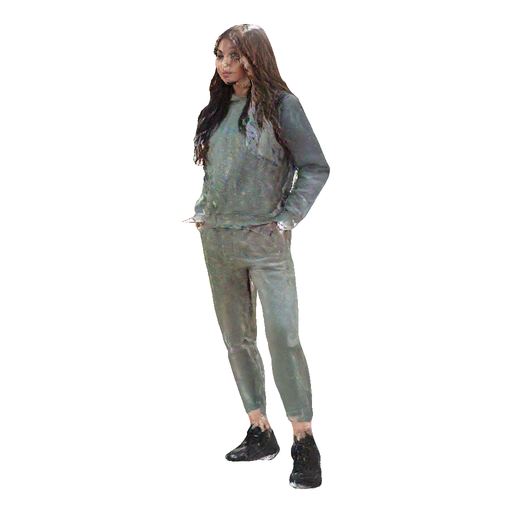}}\hfill
\mpage{0.05}{\includegraphics[width=\linewidth, trim=175 0 175 0, clip]{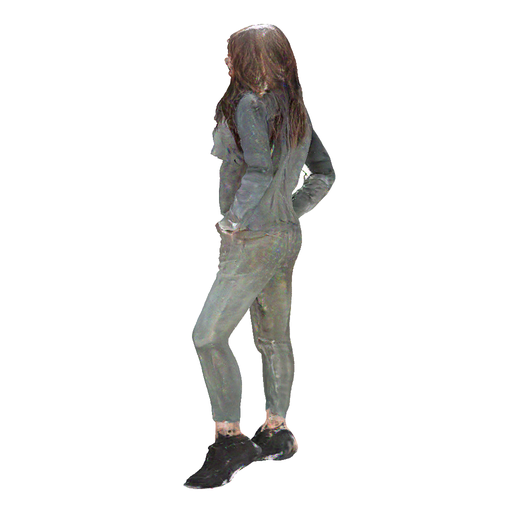}}\hfill
\mpage{0.05}{\includegraphics[width=\linewidth, trim=175 0 175 0, clip]{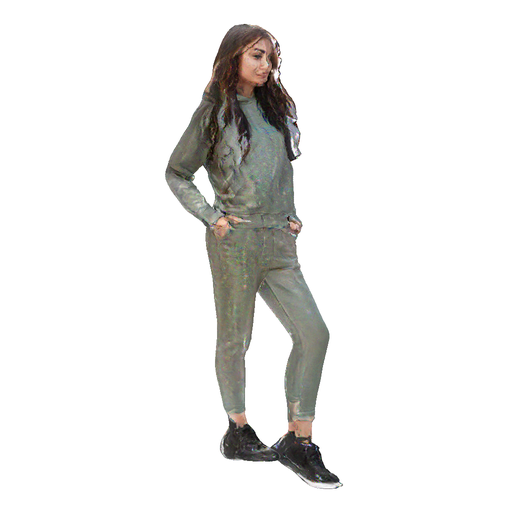}}\hfill
\hspace{2mm}
\mpage{0.05}{\includegraphics[width=\linewidth, trim=175 0 175 0, clip]{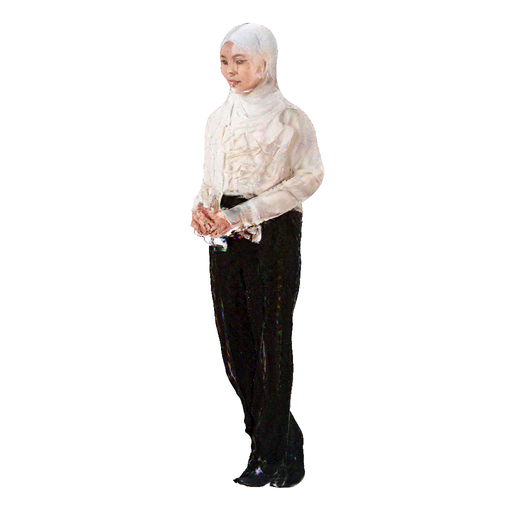}}\hfill
\mpage{0.05}{\includegraphics[width=\linewidth, trim=175 0 175 0, clip]{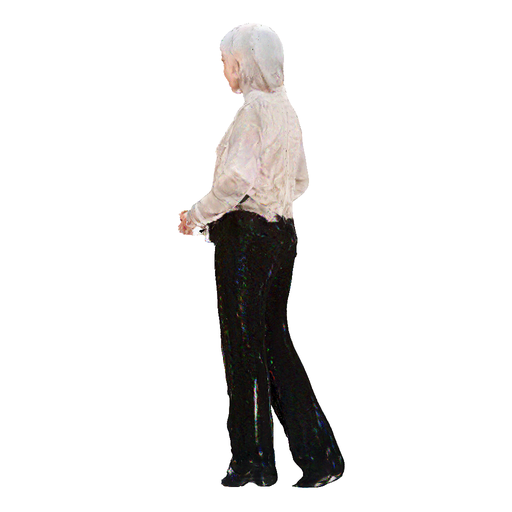}}\hfill
\mpage{0.05}{\includegraphics[width=\linewidth, trim=175 0 175 0, clip]{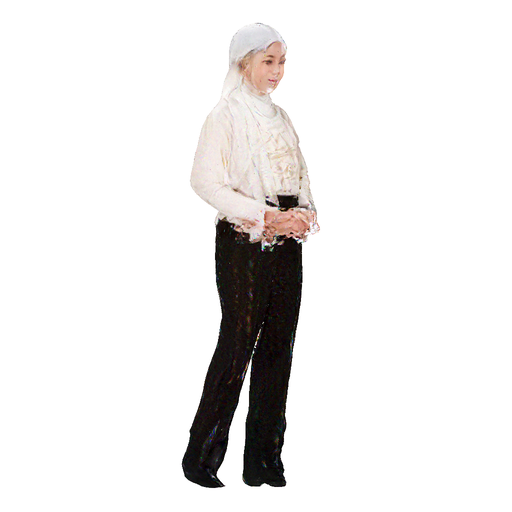}}\hfill
\hspace{2mm}
\mpage{0.05}{\includegraphics[width=\linewidth, trim=175 0 175 0, clip]{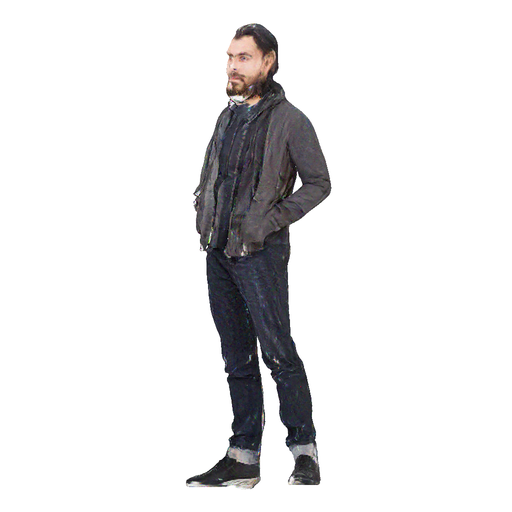}}\hfill
\mpage{0.05}{\includegraphics[width=\linewidth, trim=175 0 175 0, clip]{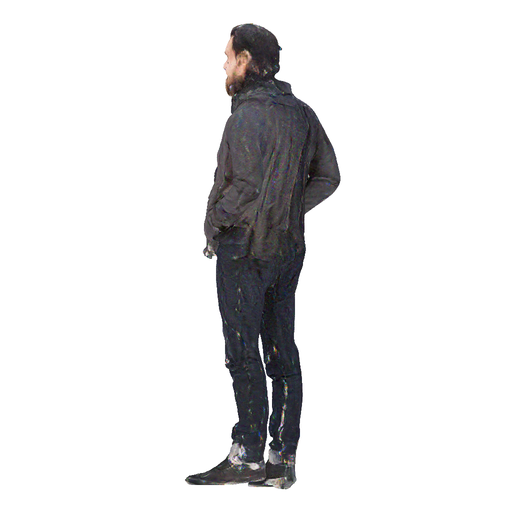}}\hfill
\mpage{0.05}{\includegraphics[width=\linewidth, trim=175 0 175 0, clip]{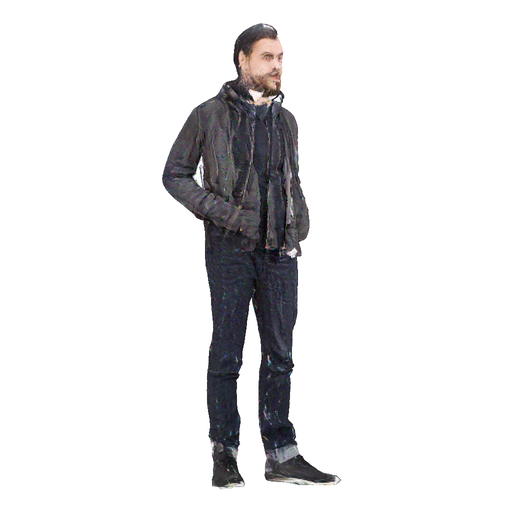}}\\

\mpage{0.03}{\raisebox{0pt}{\rotatebox{90}{PIFu~\shortcite{saito2019pifu}}}}  \hfill
\mpage{0.05}{\includegraphics[width=\linewidth, trim=175 0 175 0, clip]{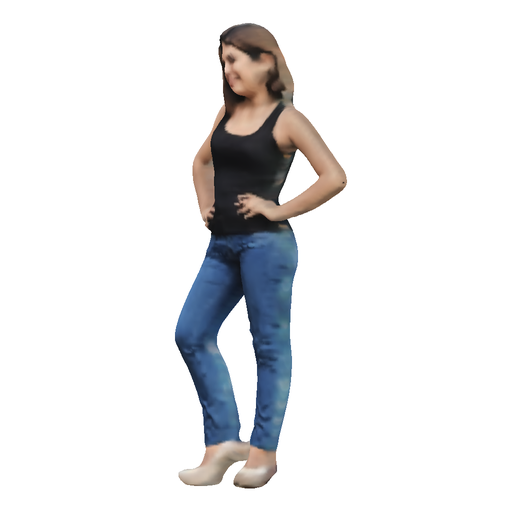}}\hfill
\mpage{0.05}{\includegraphics[width=\linewidth, trim=175 0 175 0, clip]{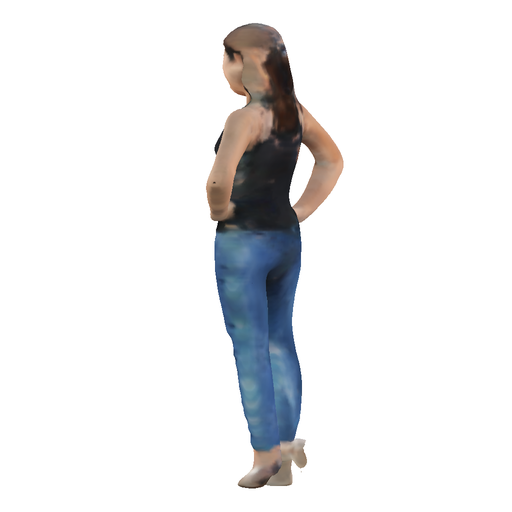}}\hfill
\mpage{0.05}{\includegraphics[width=\linewidth, trim=175 0 175 0, clip]{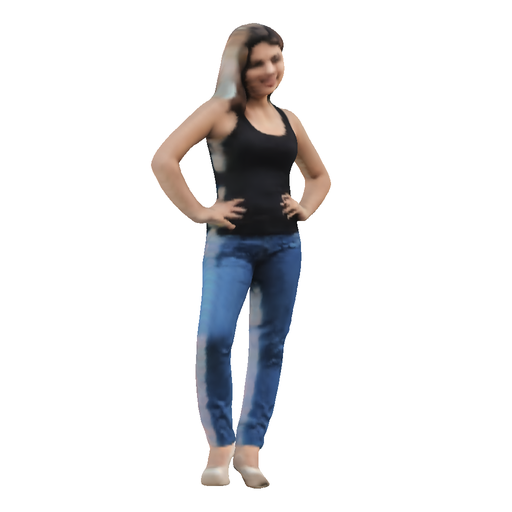}}\hfill
\hspace{2mm}
\mpage{0.05}{\includegraphics[width=\linewidth, trim=175 0 175 0, clip]{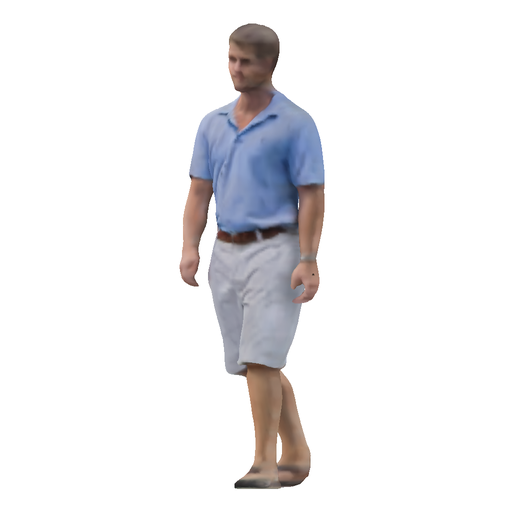}}\hfill
\mpage{0.05}{\includegraphics[width=\linewidth, trim=175 0 175 0, clip]{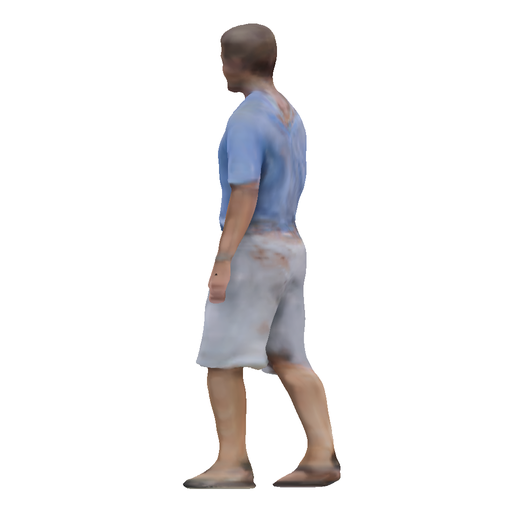}}\hfill
\mpage{0.05}{\includegraphics[width=\linewidth, trim=175 0 175 0, clip]{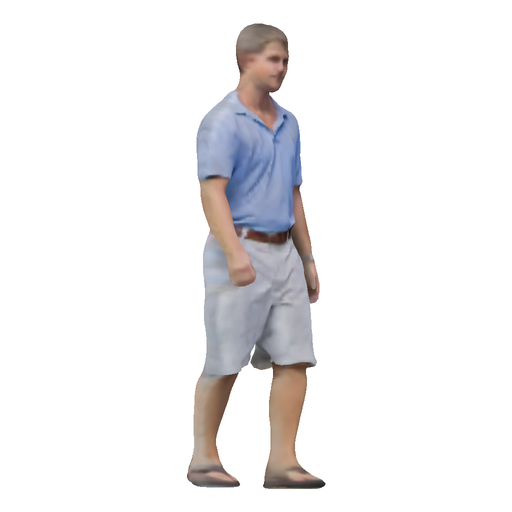}}\hfill
\hspace{2mm}
\mpage{0.05}{\includegraphics[width=\linewidth, trim=175 0 175 0, clip]{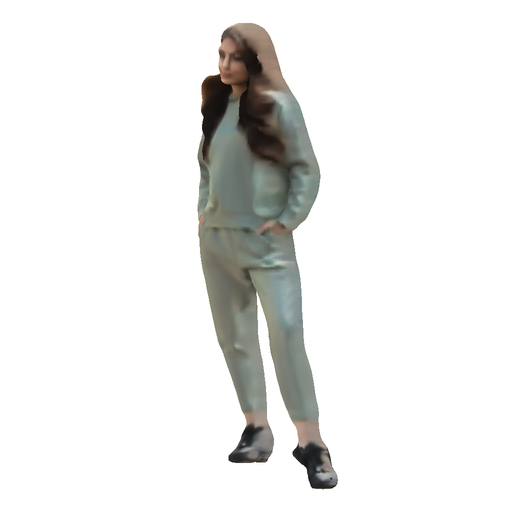}}\hfill
\mpage{0.05}{\includegraphics[width=\linewidth, trim=175 0 175 0, clip]{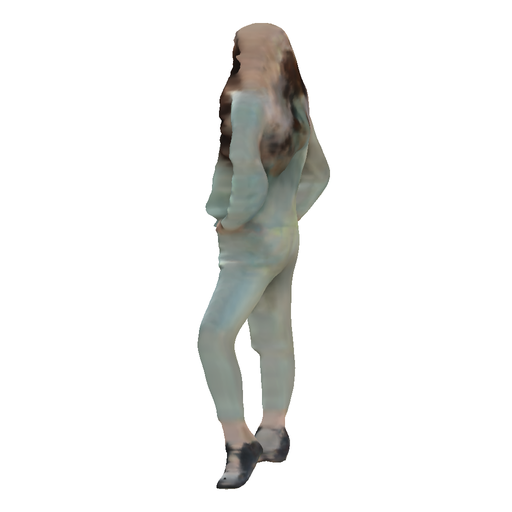}}\hfill
\mpage{0.05}{\includegraphics[width=\linewidth, trim=175 0 175 0, clip]{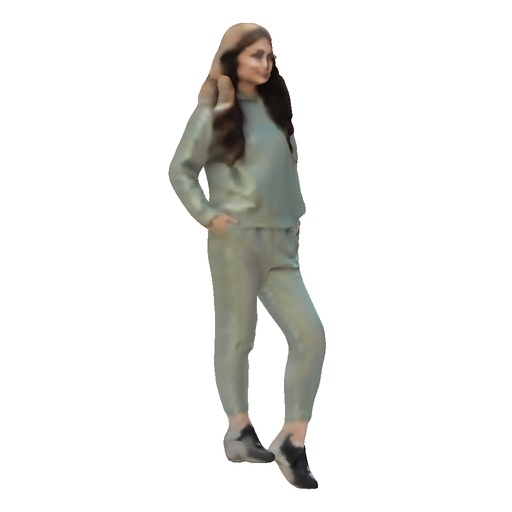}}\hfill
\hspace{2mm}
\mpage{0.05}{\includegraphics[width=\linewidth, trim=175 0 175 0, clip]{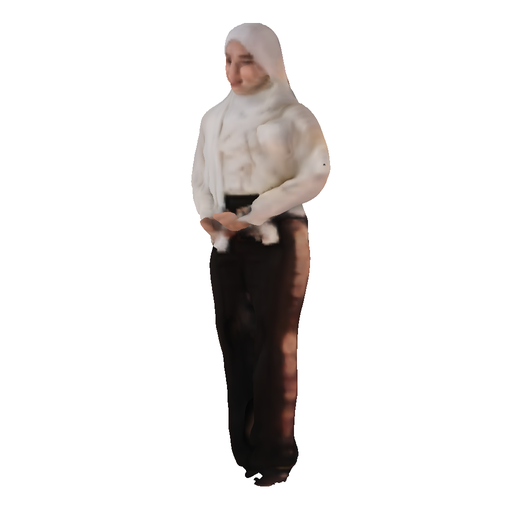}}\hfill
\mpage{0.05}{\includegraphics[width=\linewidth, trim=175 0 175 0, clip]{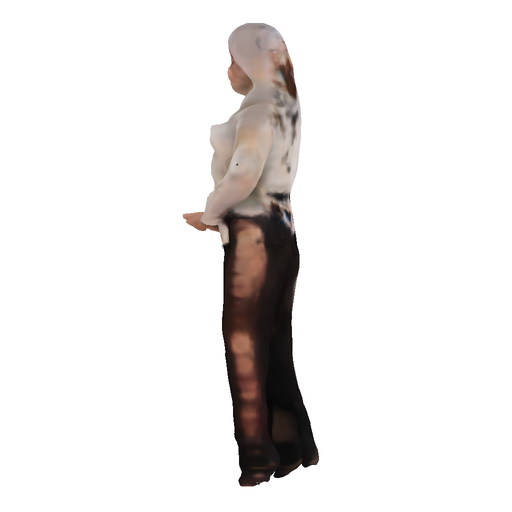}}\hfill
\mpage{0.05}{\includegraphics[width=\linewidth, trim=175 0 175 0, clip]{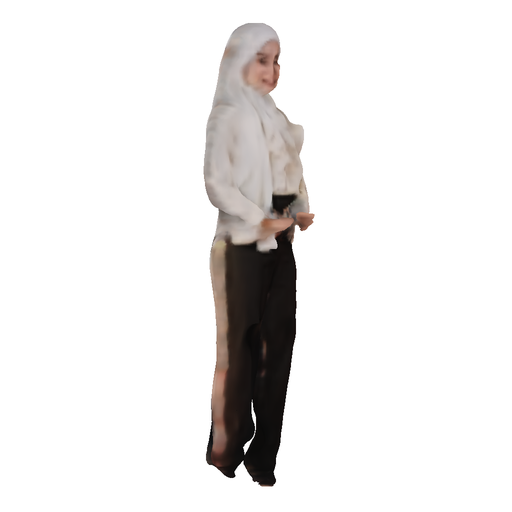}}\hfill
\hspace{2mm}
\mpage{0.05}{\includegraphics[width=\linewidth, trim=175 0 175 0, clip]{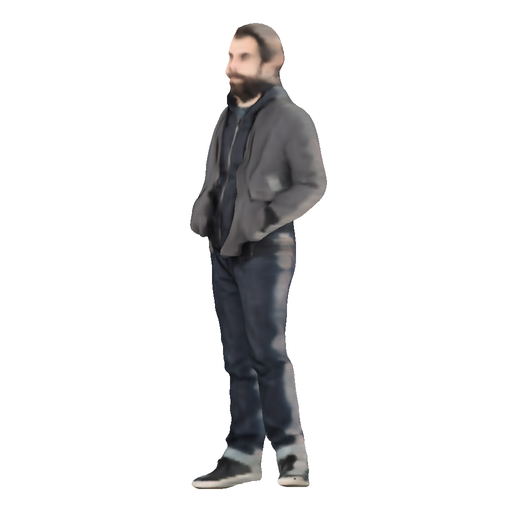}}\hfill
\mpage{0.05}{\includegraphics[width=\linewidth, trim=175 0 175 0, clip]{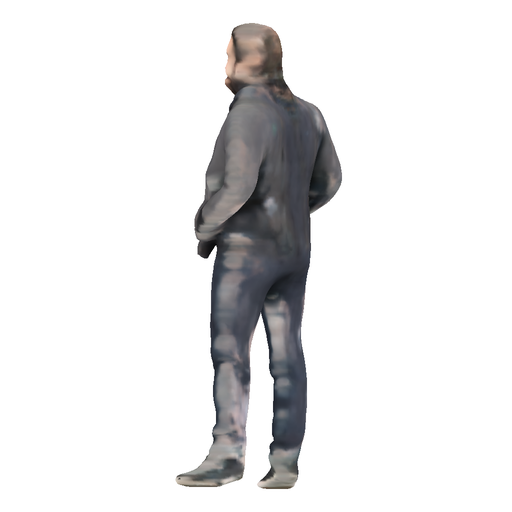}}\hfill
\mpage{0.05}{\includegraphics[width=\linewidth, trim=175 0 175 0, clip]{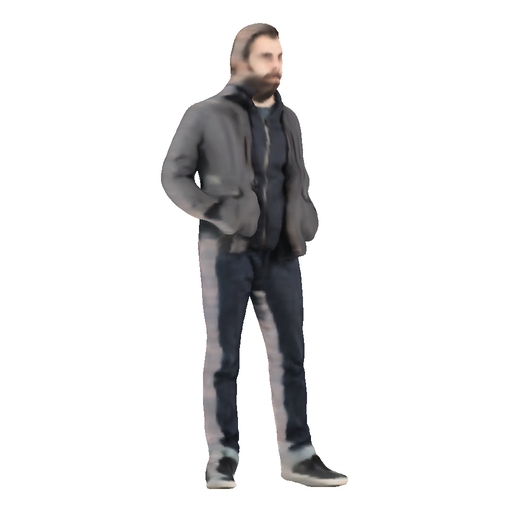}}\\

\mpage{0.03}{\raisebox{0pt}{\rotatebox{90}{Impersonator++~\shortcite{liu2021liquid}}}}  \hfill
\mpage{0.05}{\includegraphics[width=\linewidth, trim=175 0 175 0, clip]{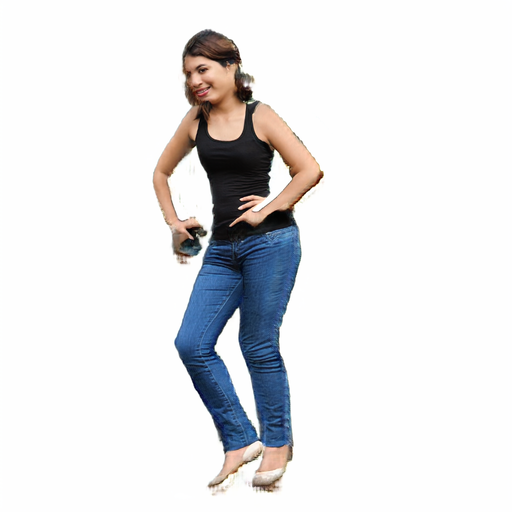}}\hfill
\mpage{0.05}{\includegraphics[width=\linewidth, trim=175 0 175 0, clip]{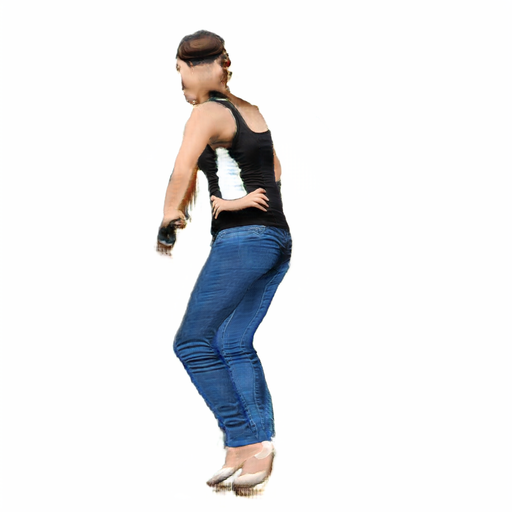}}\hfill
\mpage{0.05}{\includegraphics[width=\linewidth, trim=175 0 175 0, clip]{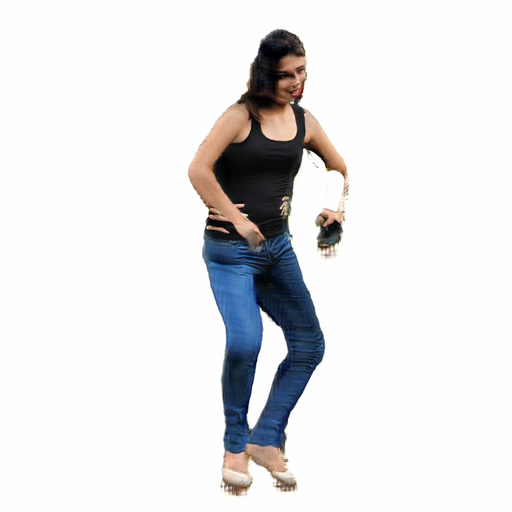}}\hfill
\hspace{2mm}
\mpage{0.05}{\includegraphics[width=\linewidth, trim=175 0 175 0, clip]{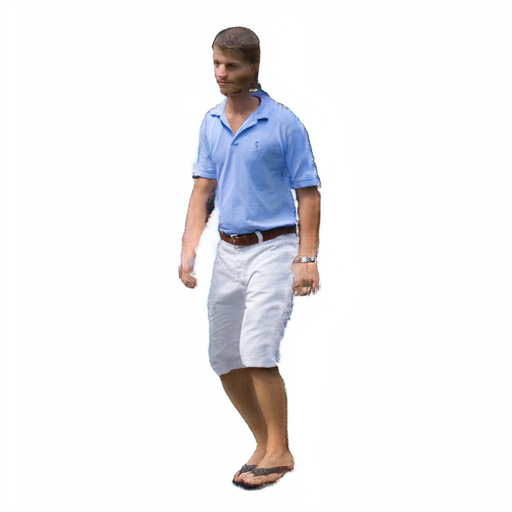}}\hfill
\mpage{0.05}{\includegraphics[width=\linewidth, trim=175 0 175 0, clip]{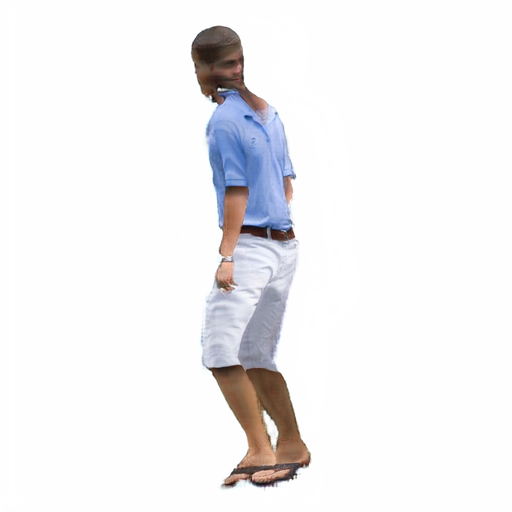}}\hfill
\mpage{0.05}{\includegraphics[width=\linewidth, trim=175 0 175 0, clip]{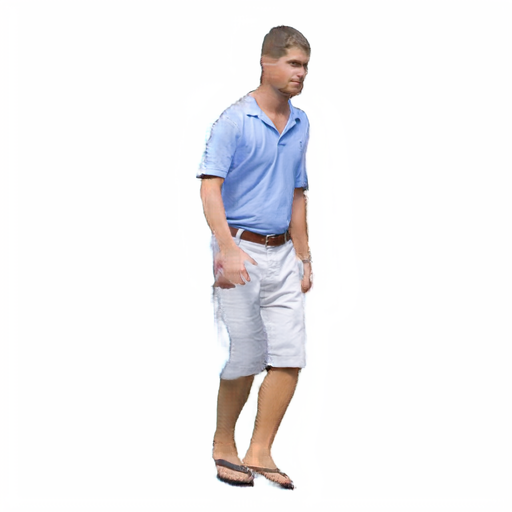}}\hfill
\hspace{2mm}
\mpage{0.05}{\includegraphics[width=\linewidth, trim=175 0 175 0, clip]{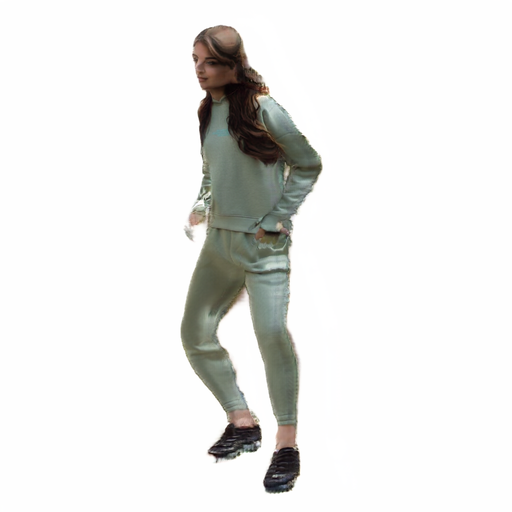}}\hfill
\mpage{0.05}{\includegraphics[width=\linewidth, trim=175 0 175 0, clip]{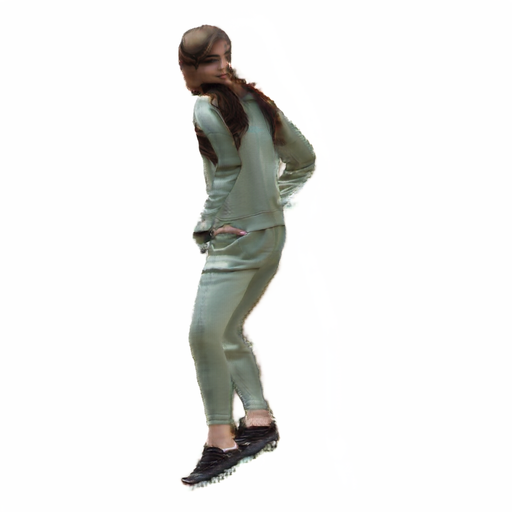}}\hfill
\mpage{0.05}{\includegraphics[width=\linewidth, trim=175 0 175 0, clip]{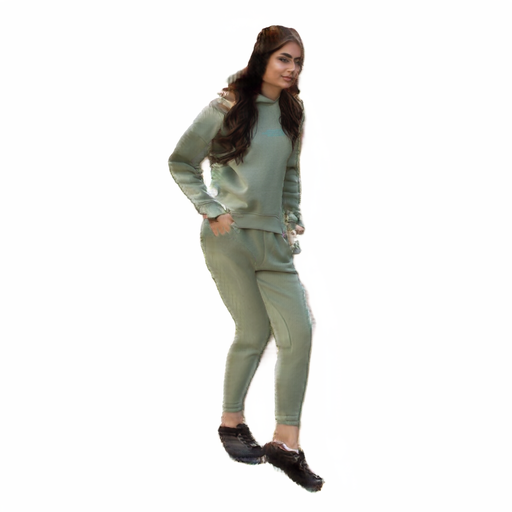}}\hfill
\hspace{2mm}
\mpage{0.05}{\includegraphics[width=\linewidth, trim=175 0 175 0, clip]{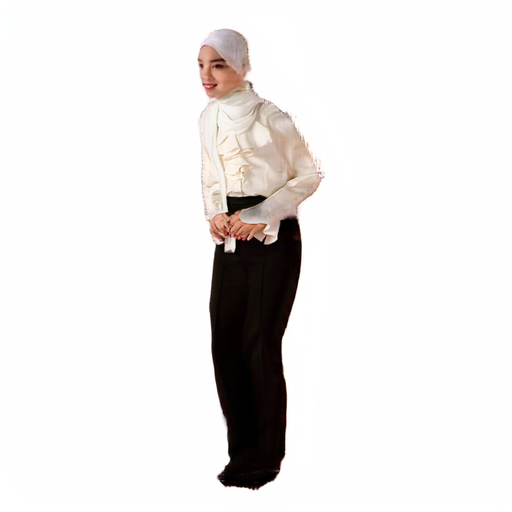}}\hfill
\mpage{0.05}{\includegraphics[width=\linewidth, trim=175 0 175 0, clip]{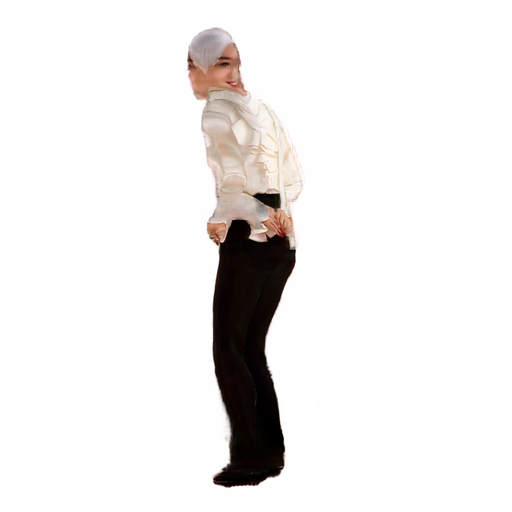}}\hfill
\mpage{0.05}{\includegraphics[width=\linewidth, trim=175 0 175 0, clip]{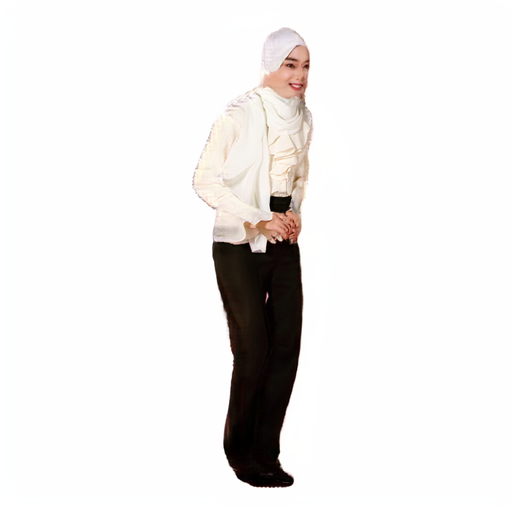}}\hfill
\hspace{2mm}
\mpage{0.05}{\includegraphics[width=\linewidth, trim=175 0 175 0, clip]{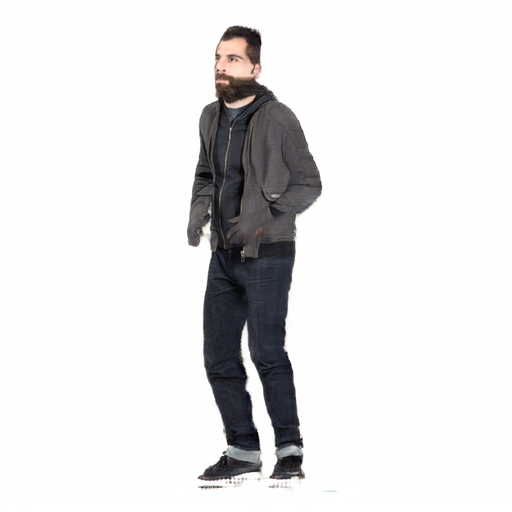}}\hfill
\mpage{0.05}{\includegraphics[width=\linewidth, trim=175 0 175 0, clip]{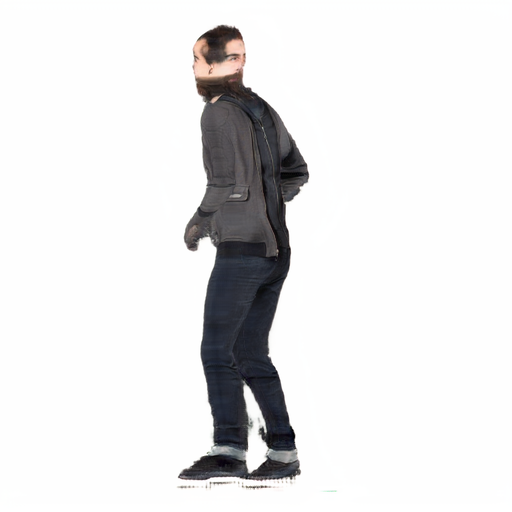}}\hfill
\mpage{0.05}{\includegraphics[width=\linewidth, trim=175 0 175 0, clip]{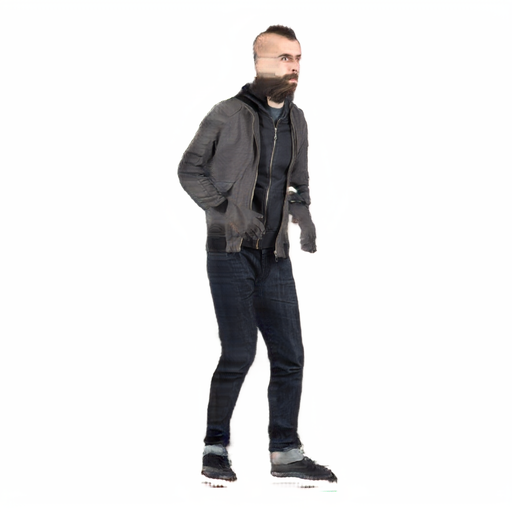}}\\

\mpage{0.03}{\raisebox{0pt}{\rotatebox{90}{TEXTure~\shortcite{richardson2023texture}}}}  \hfill
\mpage{0.05}{\includegraphics[width=\linewidth, trim=175 0 175 0, clip]{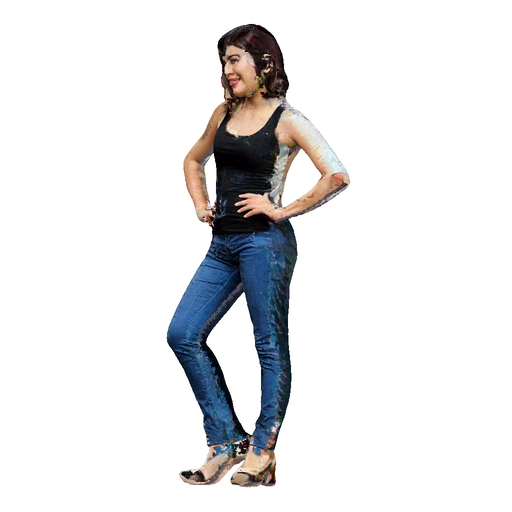}}\hfill
\mpage{0.05}{\includegraphics[width=\linewidth, trim=175 0 175 0, clip]{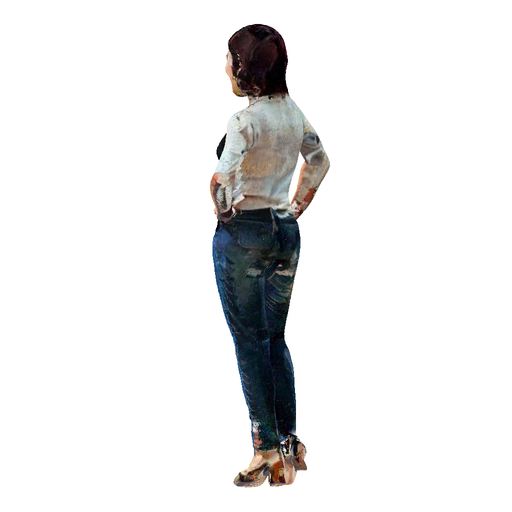}}\hfill
\mpage{0.05}{\includegraphics[width=\linewidth, trim=175 0 175 0, clip]{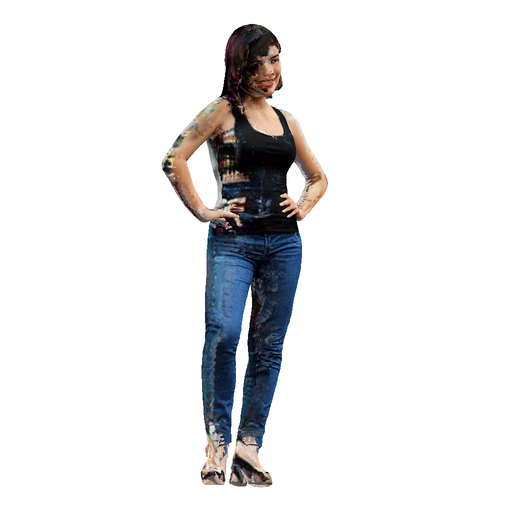}}\hfill
\hspace{2mm}
\mpage{0.05}{\includegraphics[width=\linewidth, trim=175 0 175 0, clip]{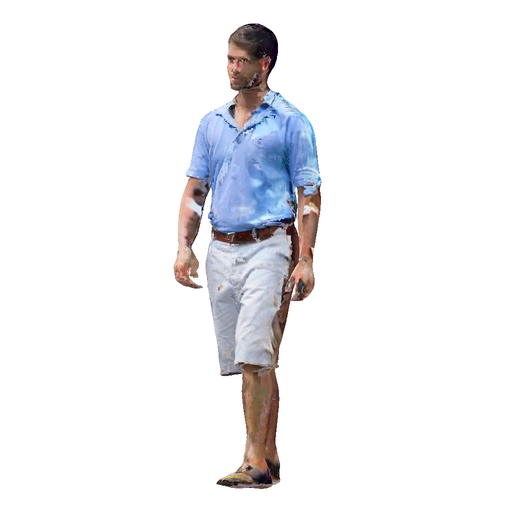}}\hfill
\mpage{0.05}{\includegraphics[width=\linewidth, trim=175 0 175 0, clip]{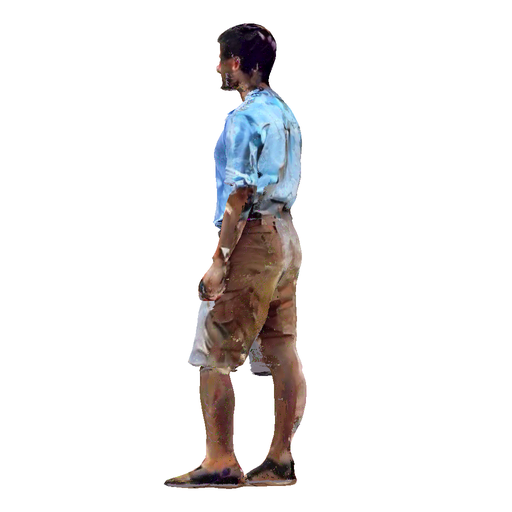}}\hfill
\mpage{0.05}{\includegraphics[width=\linewidth, trim=175 0 175 0, clip]{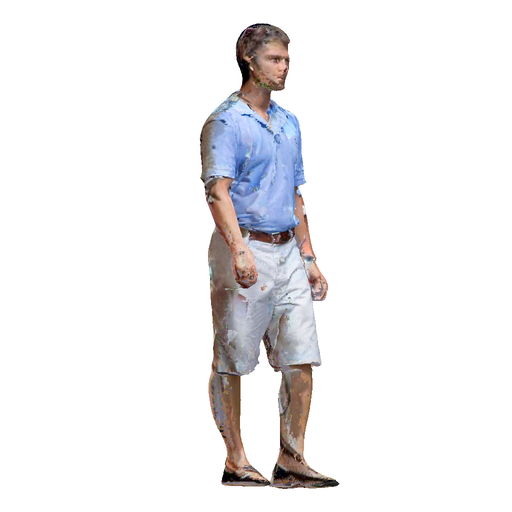}}\hfill
\hspace{2mm}
\mpage{0.05}{\includegraphics[width=\linewidth, trim=175 0 175 0, clip]{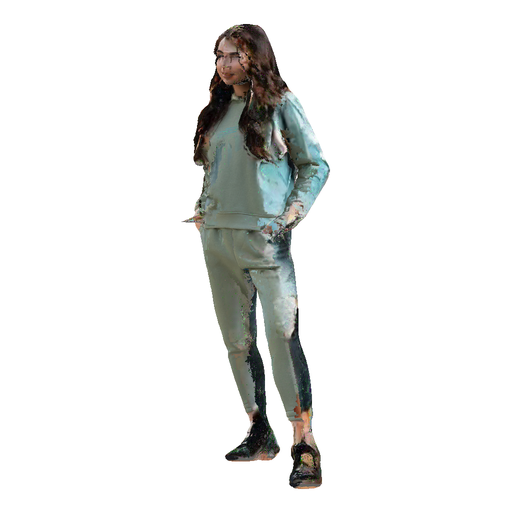}}\hfill
\mpage{0.05}{\includegraphics[width=\linewidth, trim=175 0 175 0, clip]{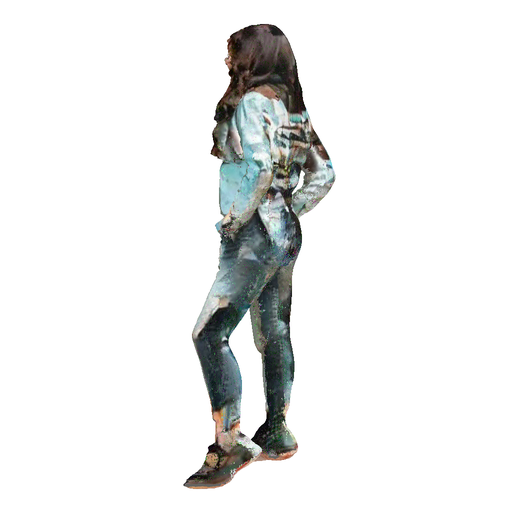}}\hfill
\mpage{0.05}{\includegraphics[width=\linewidth, trim=175 0 175 0, clip]{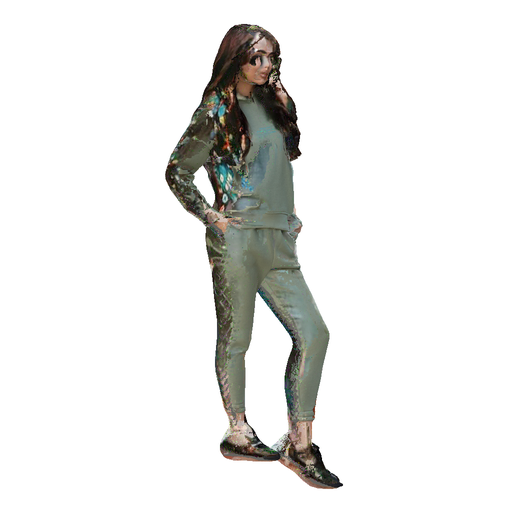}}\hfill
\hspace{2mm}
\mpage{0.05}{\includegraphics[width=\linewidth, trim=175 0 175 0, clip]{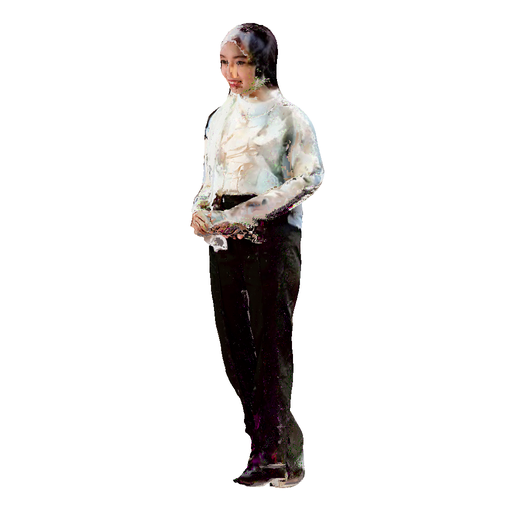}}\hfill
\mpage{0.05}{\includegraphics[width=\linewidth, trim=175 0 175 0, clip]{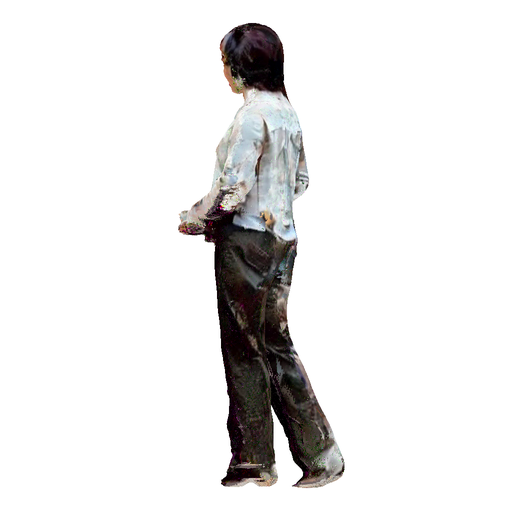}}\hfill
\mpage{0.05}{\includegraphics[width=\linewidth, trim=175 0 175 0, clip]{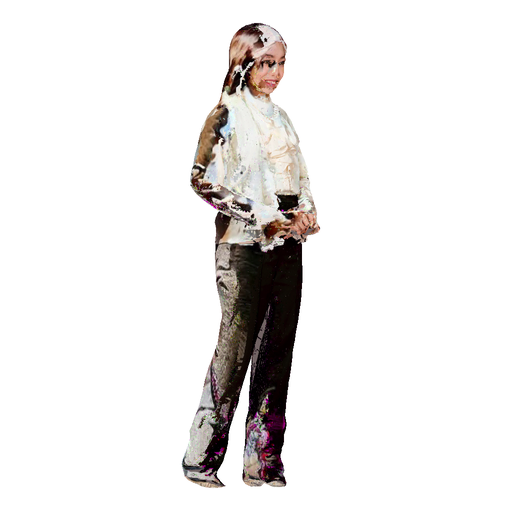}}\hfill
\hspace{2mm}
\mpage{0.05}{\includegraphics[width=\linewidth, trim=175 0 175 0, clip]{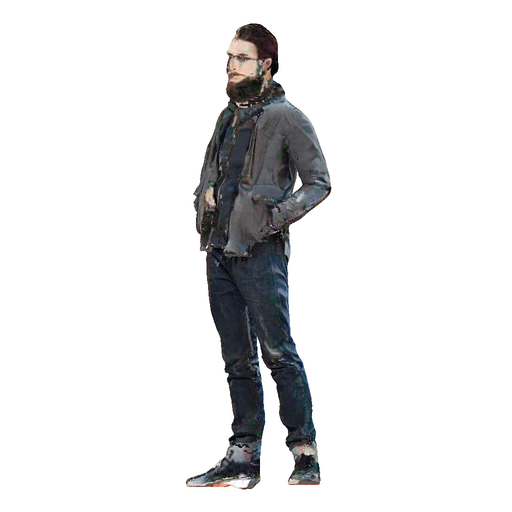}}\hfill
\mpage{0.05}{\includegraphics[width=\linewidth, trim=175 0 175 0, clip]{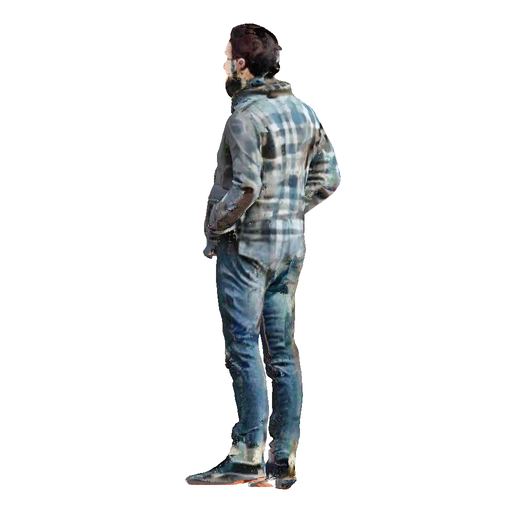}}\hfill
\mpage{0.05}{\includegraphics[width=\linewidth, trim=175 0 175 0, clip]{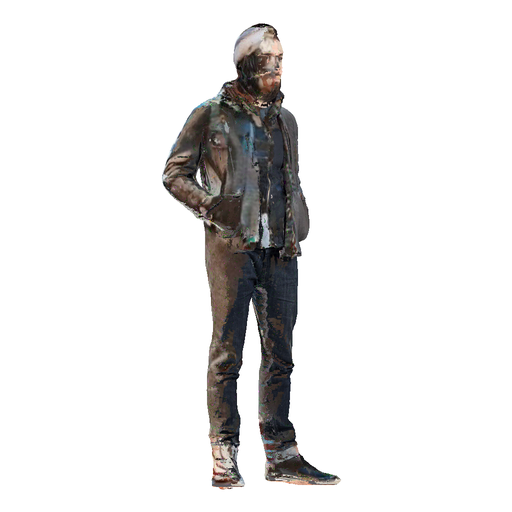}}\\

\mpage{0.03}{\raisebox{0pt}{\rotatebox{90}{Magic123~\shortcite{qian2023magic123}}}}  \hfill
\mpage{0.05}{\includegraphics[width=\linewidth, trim=175 0 175 0, clip]{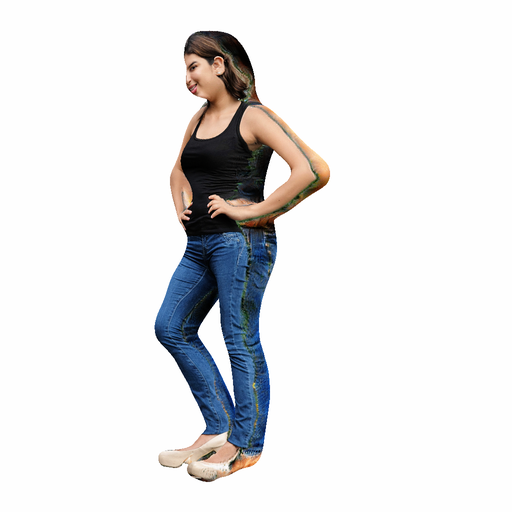}}\hfill
\mpage{0.05}{\includegraphics[width=\linewidth, trim=160 0 179 0, clip]{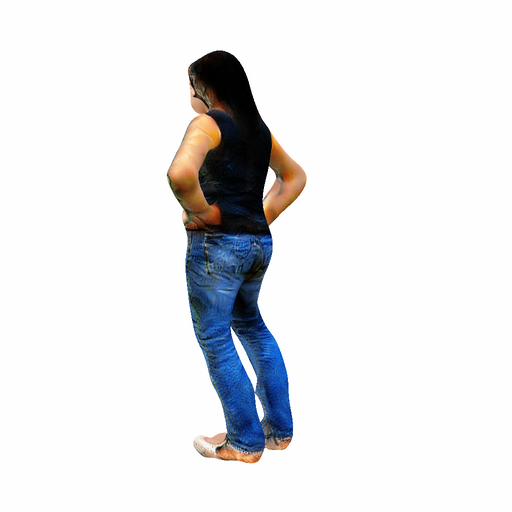}}\hfill
\mpage{0.05}{\includegraphics[width=\linewidth, trim=175 0 175 0, clip]{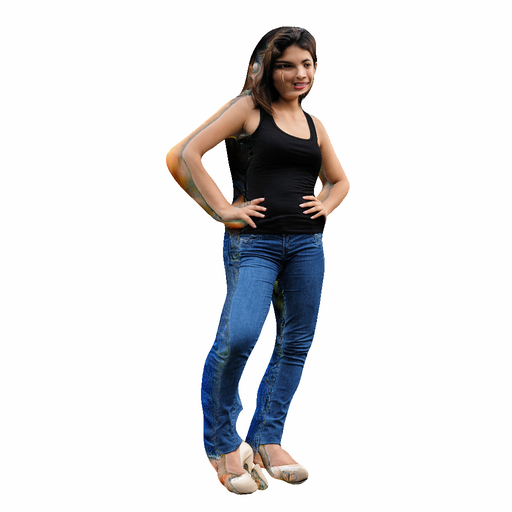}}\hfill
\hspace{2mm}
\mpage{0.05}{\includegraphics[width=\linewidth, trim=175 0 175 0, clip]{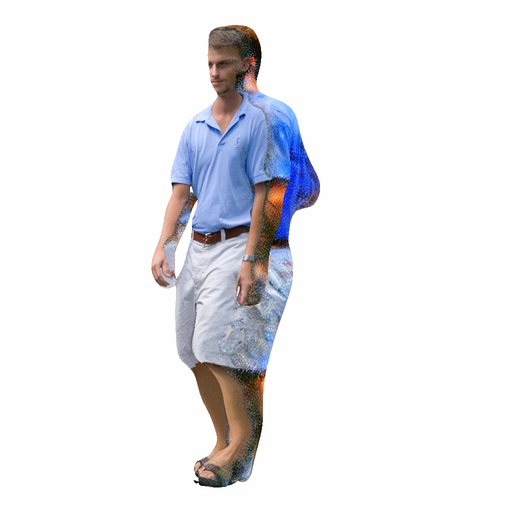}}\hfill
\mpage{0.05}{\includegraphics[width=\linewidth, trim=160 0 190 0, clip]{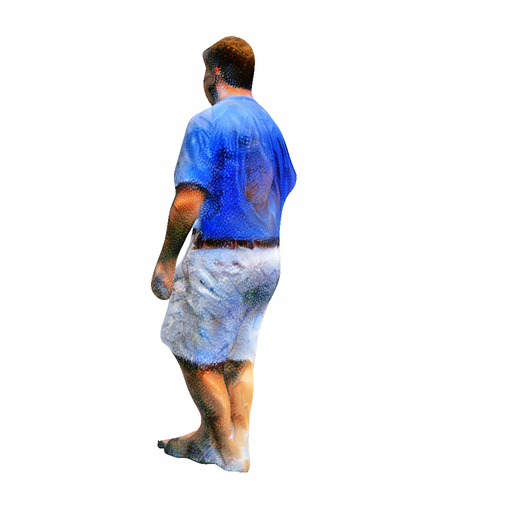}}\hfill
\mpage{0.05}{\includegraphics[width=\linewidth, trim=175 0 175 0, clip]{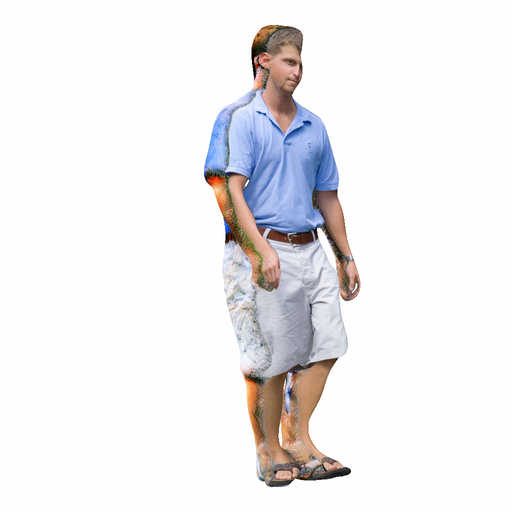}}\hfill
\hspace{2mm}
\mpage{0.05}{\includegraphics[width=\linewidth, trim=175 0 175 0, clip]{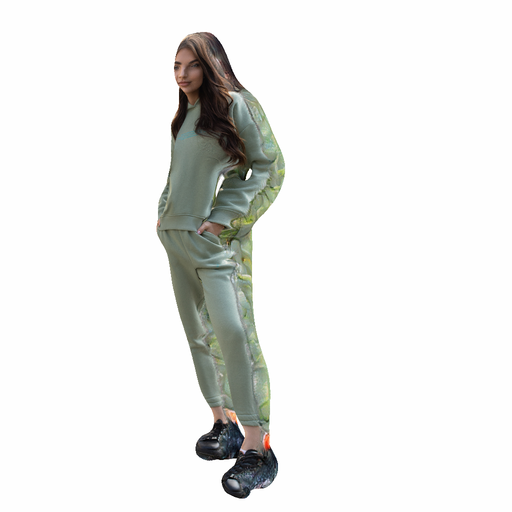}}\hfill
\mpage{0.05}{\includegraphics[width=\linewidth, trim=160 0 190 0, clip]{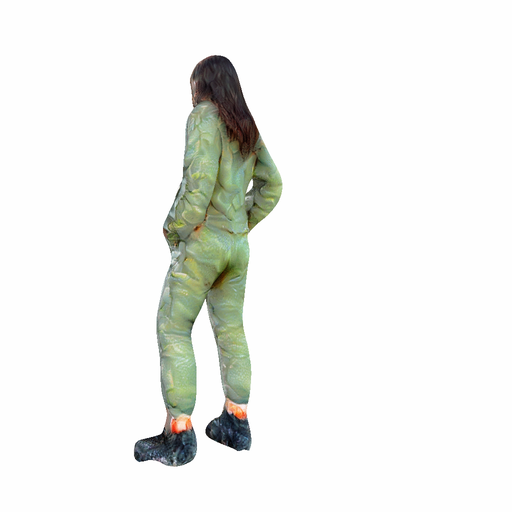}}\hfill
\mpage{0.05}{\includegraphics[width=\linewidth, trim=190 0 160 0, clip]{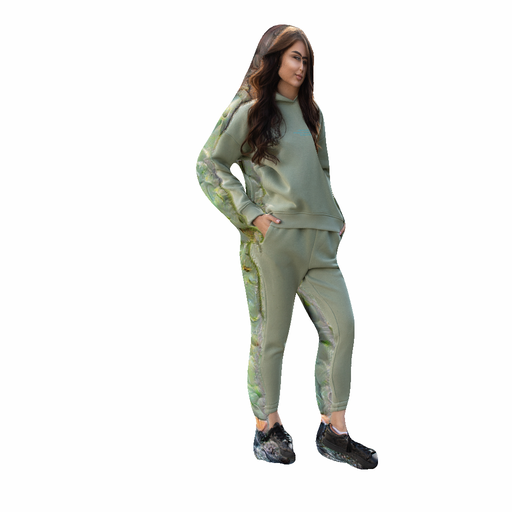}}\hfill
\hspace{2mm}
\mpage{0.05}{\includegraphics[width=\linewidth, trim=175 0 175 0, clip]{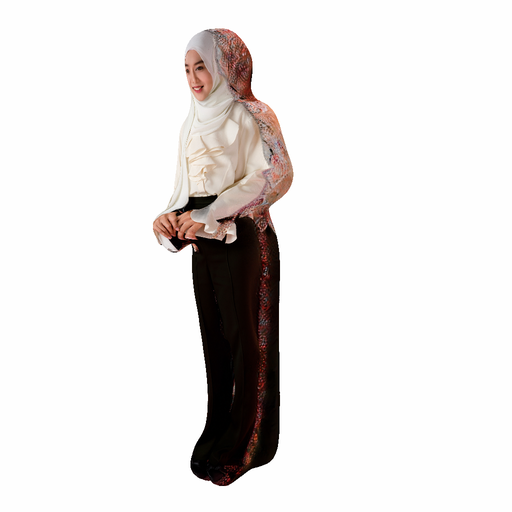}}\hfill
\mpage{0.05}{\includegraphics[width=\linewidth, trim=160 0 190 0, clip]{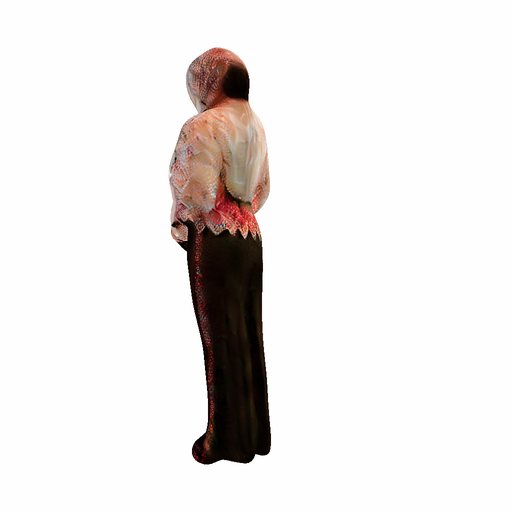}}\hfill
\mpage{0.05}{\includegraphics[width=\linewidth, trim=175 0 175 0, clip]{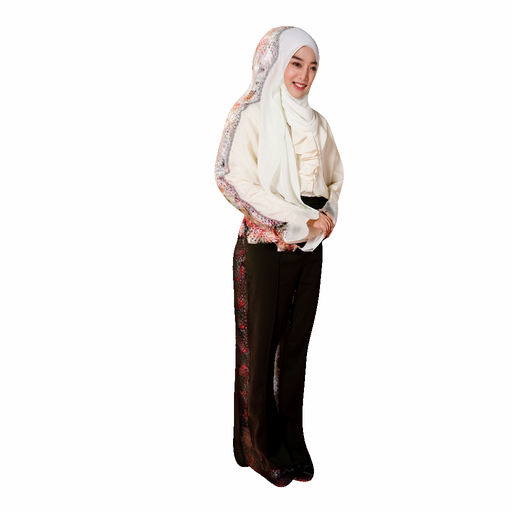}}\hfill
\hspace{2mm}
\mpage{0.05}{\includegraphics[width=\linewidth, trim=175 0 175 0, clip]{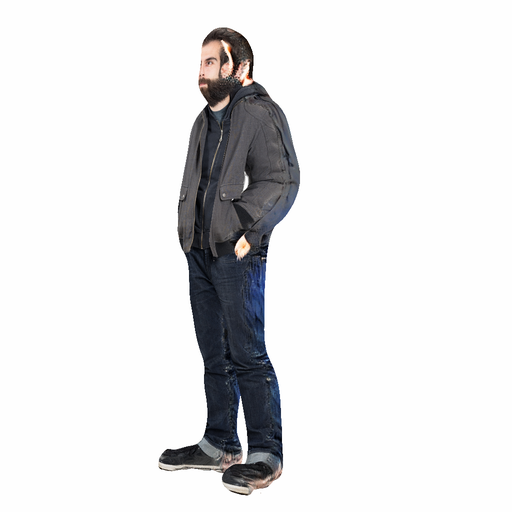}}\hfill
\mpage{0.05}{\includegraphics[width=\linewidth, trim=175 0 175 0, clip]{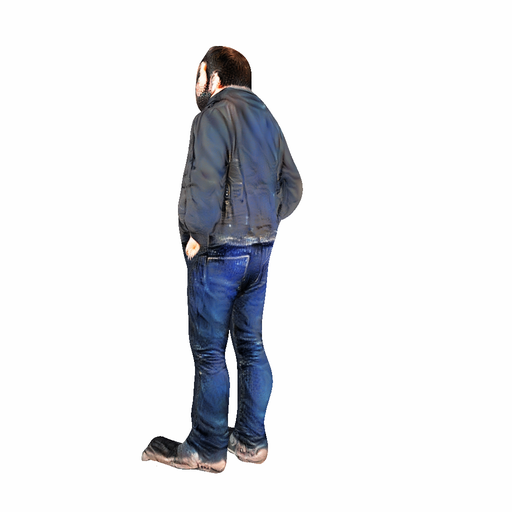}}\hfill
\mpage{0.05}{\includegraphics[width=\linewidth, trim=160 0 190 0, clip]{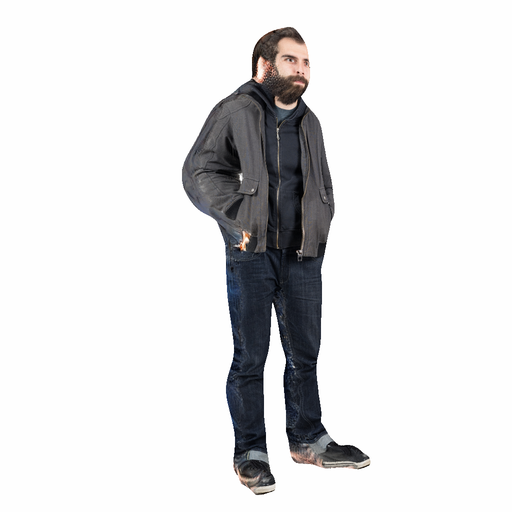}}\\

\mpage{0.03}{\raisebox{0pt}{\rotatebox{90}{Ours}}}  \hfill
\mpage{0.05}{\includegraphics[width=\linewidth, trim=175 0 175 0, clip]{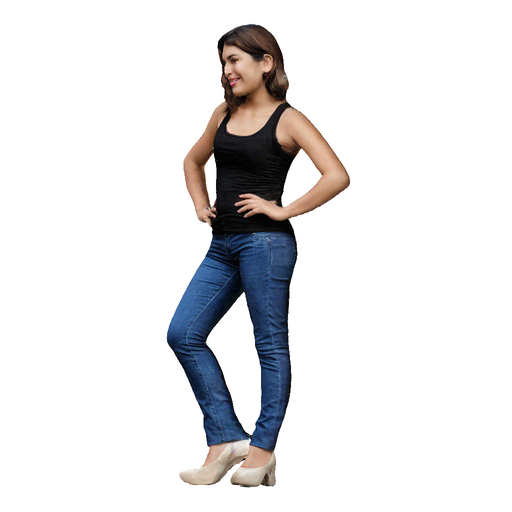}}\hfill
\mpage{0.05}{\includegraphics[width=\linewidth, trim=175 0 175 0, clip]{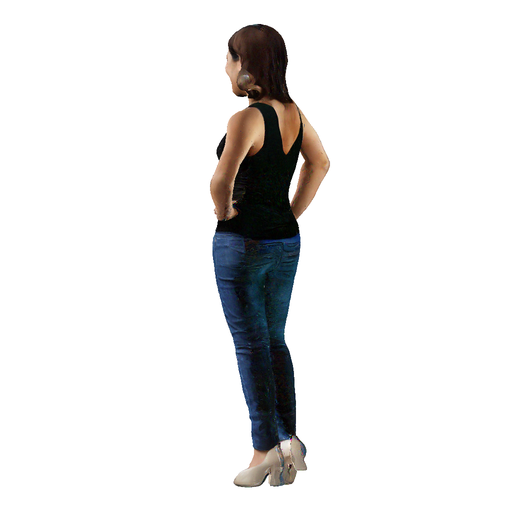}}\hfill
\mpage{0.05}{\includegraphics[width=\linewidth, trim=175 0 175 0, clip]{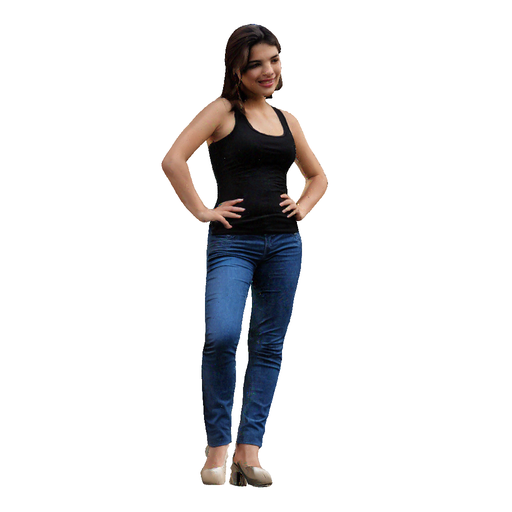}}\hfill
\hspace{2mm}
\mpage{0.05}{\includegraphics[width=\linewidth, trim=175 0 175 0, clip]{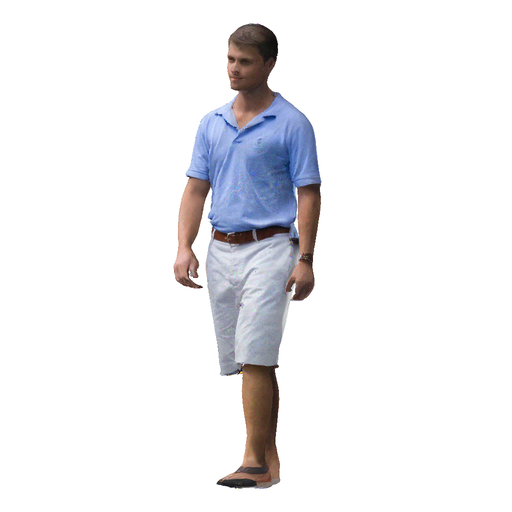}}\hfill
\mpage{0.05}{\includegraphics[width=\linewidth, trim=175 0 175 0, clip]{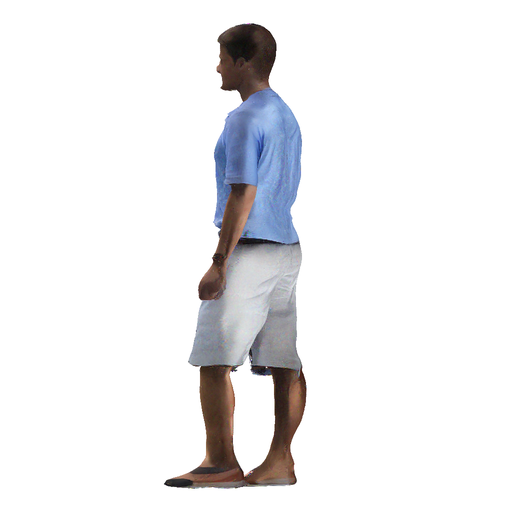}}\hfill
\mpage{0.05}{\includegraphics[width=\linewidth, trim=175 0 175 0, clip]{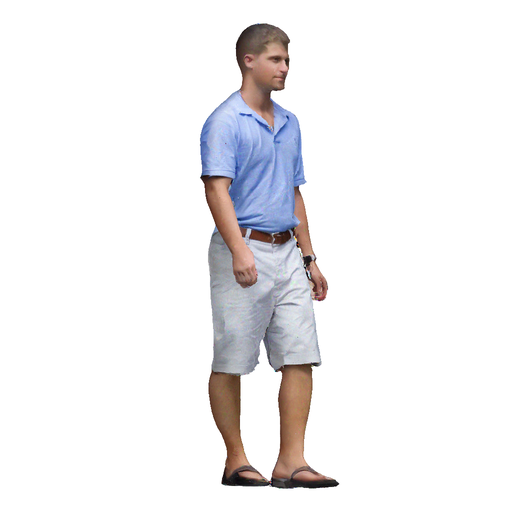}}\hfill
\hspace{2mm}
\mpage{0.05}{\includegraphics[width=\linewidth, trim=175 0 175 0, clip]{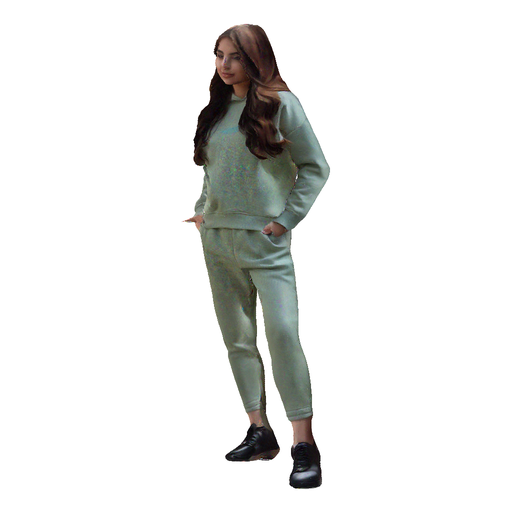}}\hfill
\mpage{0.05}{\includegraphics[width=\linewidth, trim=175 0 175 0, clip]{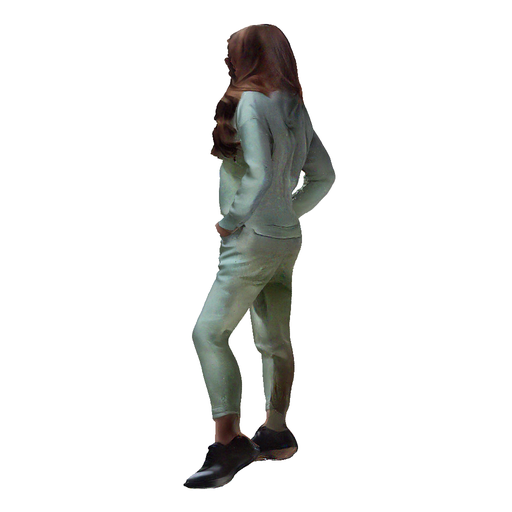}}\hfill
\mpage{0.05}{\includegraphics[width=\linewidth, trim=175 0 175 0, clip]{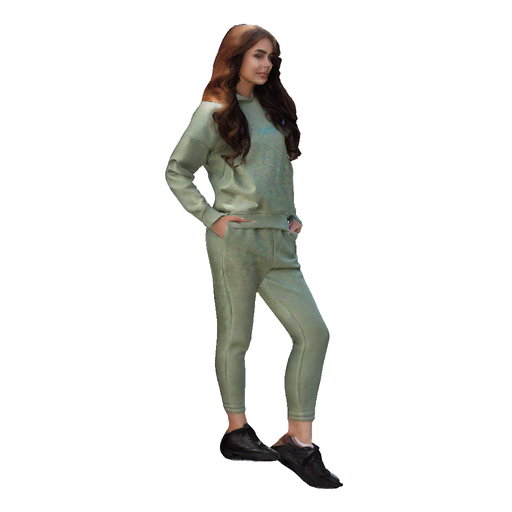}}\hfill
\hspace{2mm}
\mpage{0.05}{\includegraphics[width=\linewidth, trim=175 0 175 0, clip]{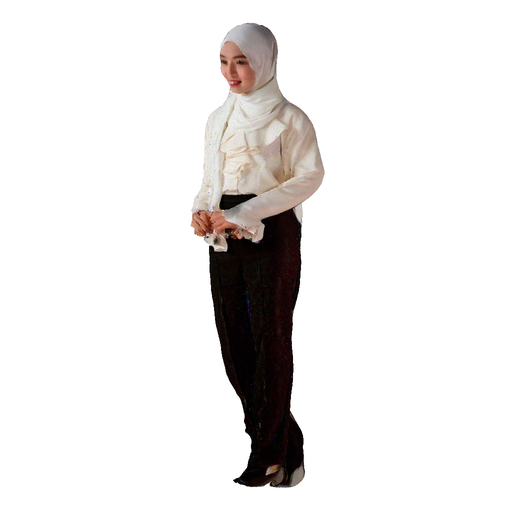}}\hfill
\mpage{0.05}{\includegraphics[width=\linewidth, trim=175 0 175 0, clip]{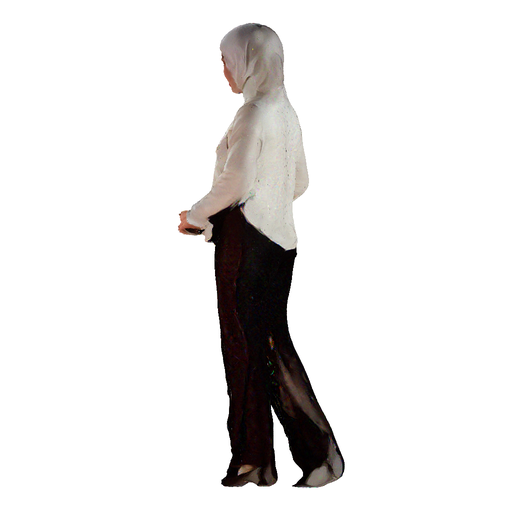}}\hfill
\mpage{0.05}{\includegraphics[width=\linewidth, trim=175 0 175 0, clip]{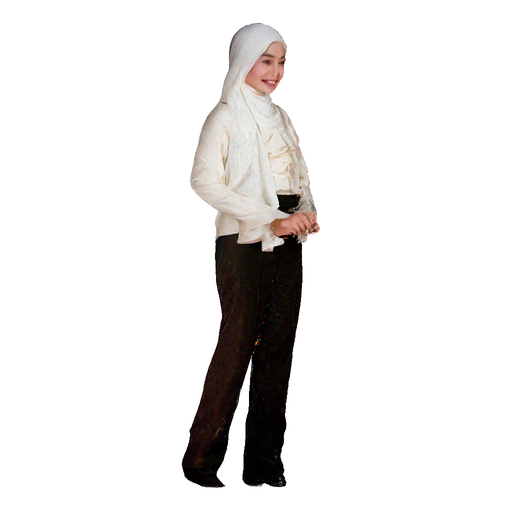}}\hfill
\hspace{2mm}
\mpage{0.05}{\includegraphics[width=\linewidth, trim=175 0 175 0, clip]{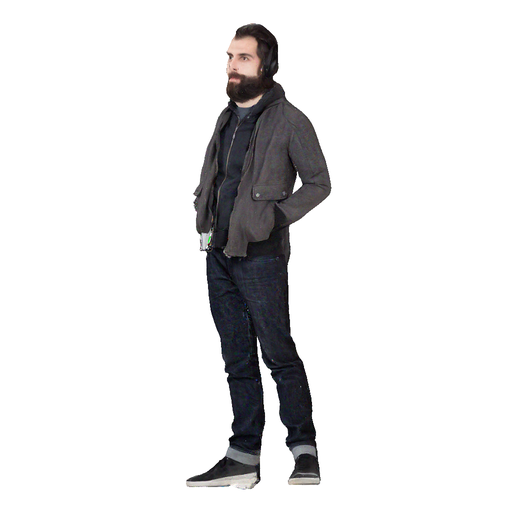}}\hfill
\mpage{0.05}{\includegraphics[width=\linewidth, trim=175 0 175 0, clip]{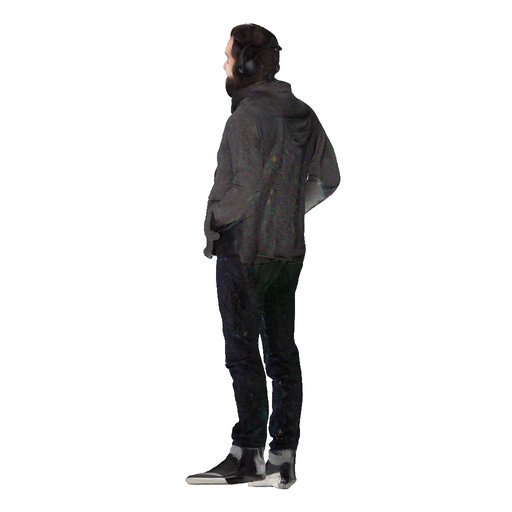}}\hfill
\mpage{0.05}{\includegraphics[width=\linewidth, trim=175 0 175 0, clip]{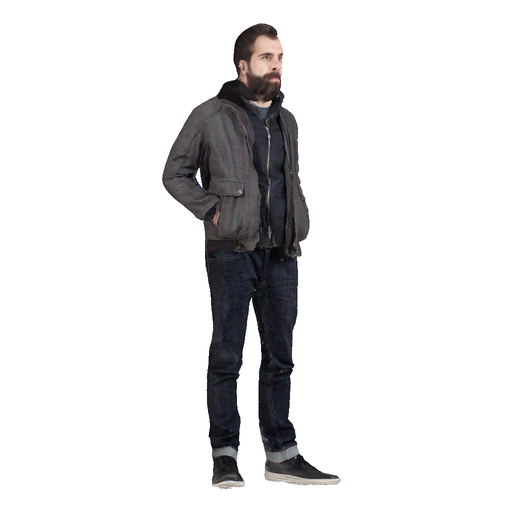}}\\

\caption{\textbf{Visual comparison on in-the-wild images from Adobe Stock.}
We compare our 3D human digitization approach with prior methods~\cite{saito2019pifu,liu2021liquid,richardson2023texture,corona2022structured,qian2023magic123} on images in-the-wild to showcase the generalizability of our approach.
Our approach demonstrates high-resolution photorealistic results that preserve the appearance of the input image.
}
\label{fig:unsplash}
\end{figure*}

\begin{figure*}[t]
\centering

\mpage{0.03}{\raisebox{0pt}{\rotatebox{90}{Input}}}  \hfill
\mpage{0.061}{}\hfill
\mpage{0.068}{\includegraphics[width=\linewidth, trim=125 0 145 0, clip]{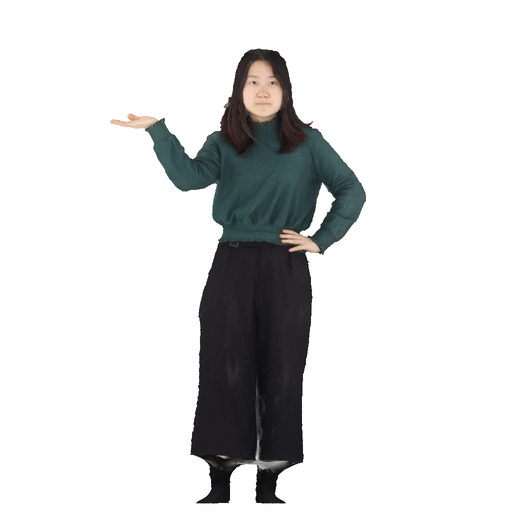}}\hfill
\mpage{0.061}{}\hfill
\mpage{0.0725}{}\hfill
\mpage{0.045}{\includegraphics[width=\linewidth, trim=175 0 175 0, clip]{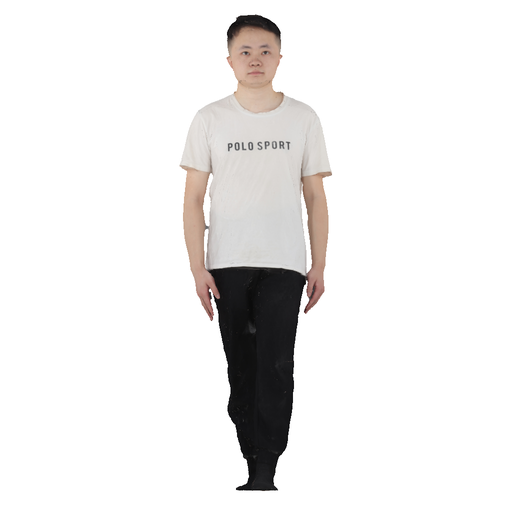}}\hfill
\mpage{0.0725}{}\hfill
\mpage{0.0725}{}\hfill
\mpage{0.045}{\includegraphics[width=\linewidth, trim=175 0 175 0, clip]{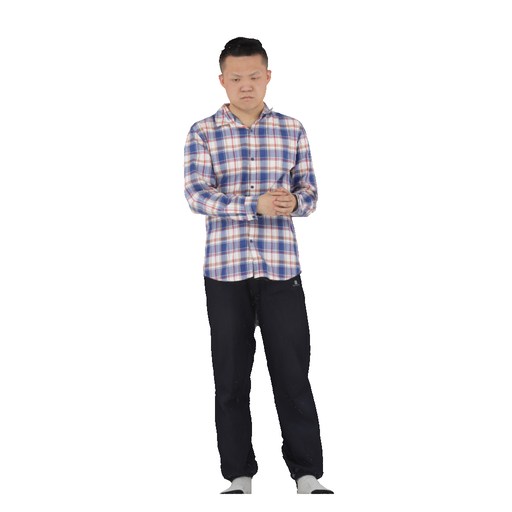}}\hfill
\mpage{0.0725}{}\hfill
\mpage{0.0675}{}\hfill
\mpage{0.055}{\includegraphics[width=\linewidth, trim=175 0 155 0, clip]{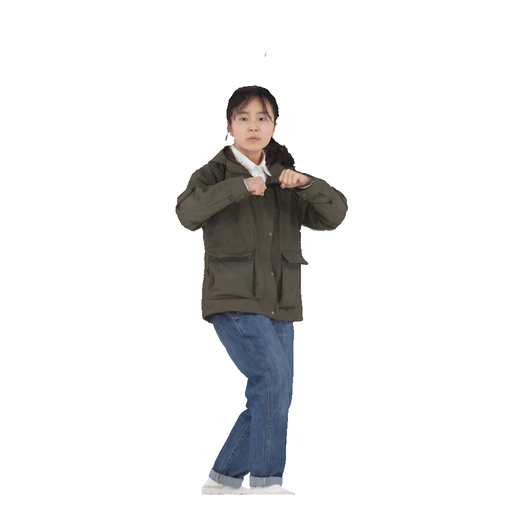}}\hfill
\mpage{0.0675}{}\hfill
\mpage{0.06}{}\hfill
\mpage{0.07}{\includegraphics[width=\linewidth, trim=140 0 125 0, clip]{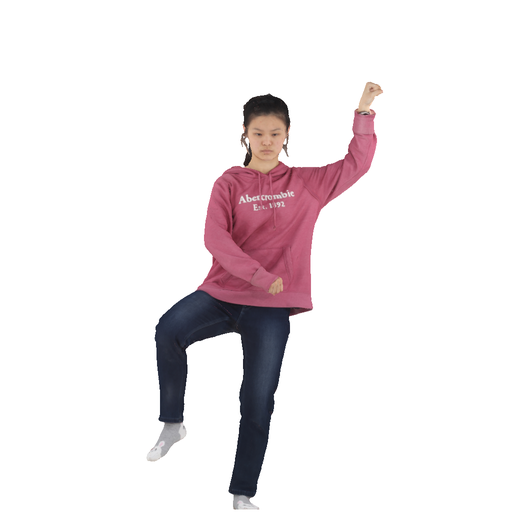}}\hfill
\hspace{5mm}
\mpage{0.06}{}\\

\vspace{-1mm}

\mpage{0.03}{\raisebox{0pt}{\rotatebox{90}{PwS baseline}}}  \hfill
\mpage{0.068}{\includegraphics[width=\linewidth, trim=125 0 145 0, clip]{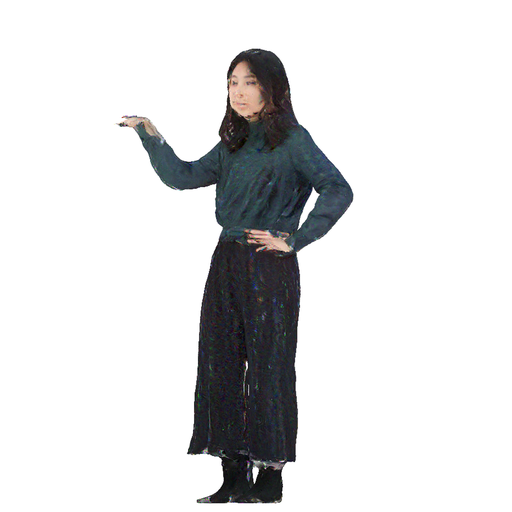}}\hfill
\hspace{-3mm}
\mpage{0.068}{\includegraphics[width=\linewidth, trim=125 0 145 0, clip]{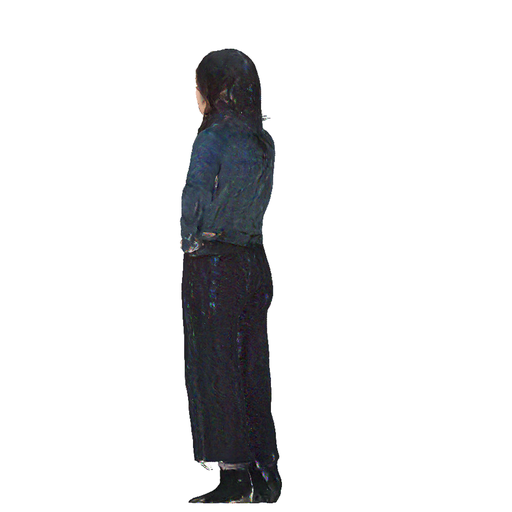}}\hfill
\hspace{-5mm}
\mpage{0.068}{\includegraphics[width=\linewidth, trim=125 0 145 0, clip]{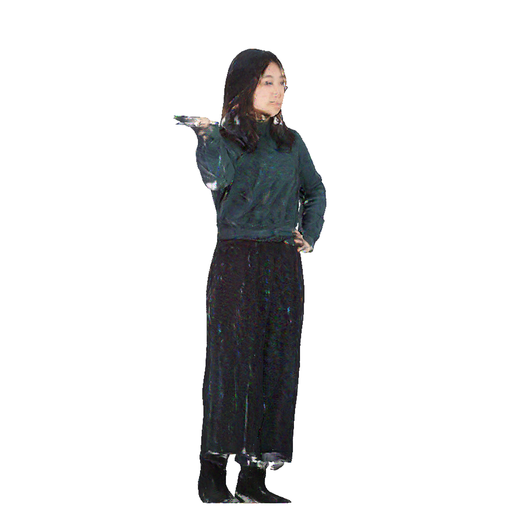}}\hfill
\hspace{2mm}
\mpage{0.045}{\includegraphics[width=\linewidth, trim=175 0 175 0, clip]{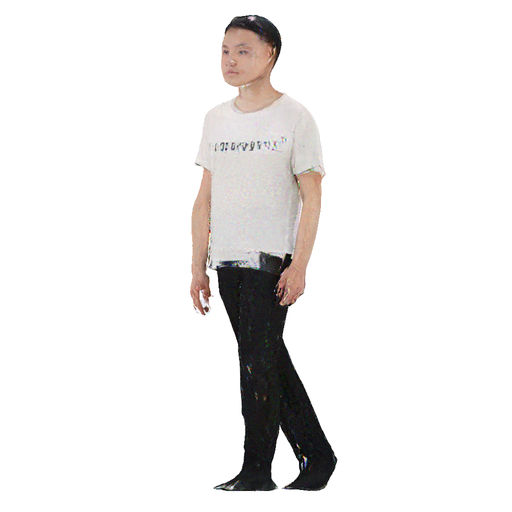}}\hfill
\mpage{0.045}{\includegraphics[width=\linewidth, trim=175 0 175 0, clip]{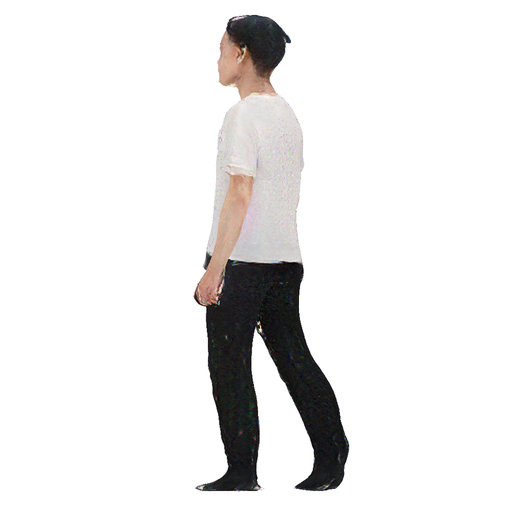}}\hfill
\mpage{0.045}{\includegraphics[width=\linewidth, trim=175 0 175 0, clip]{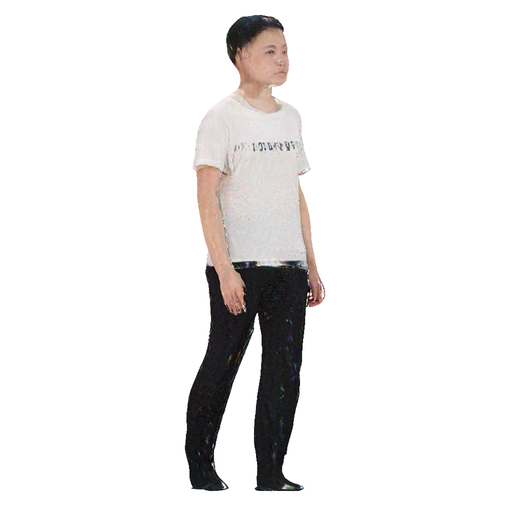}}\hfill
\hspace{2mm}
\mpage{0.045}{\includegraphics[width=\linewidth, trim=175 0 175 0, clip]{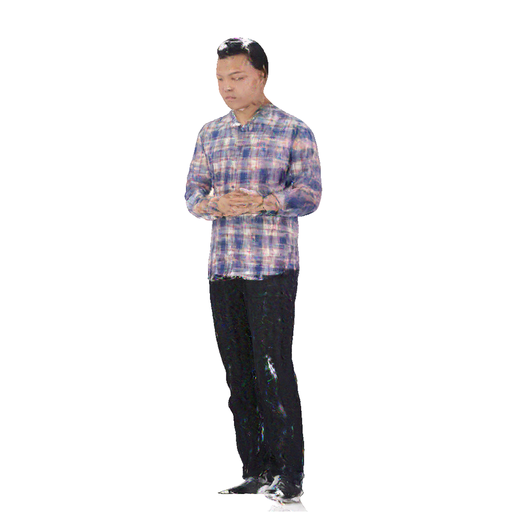}}\hfill
\mpage{0.045}{\includegraphics[width=\linewidth, trim=175 0 175 0, clip]{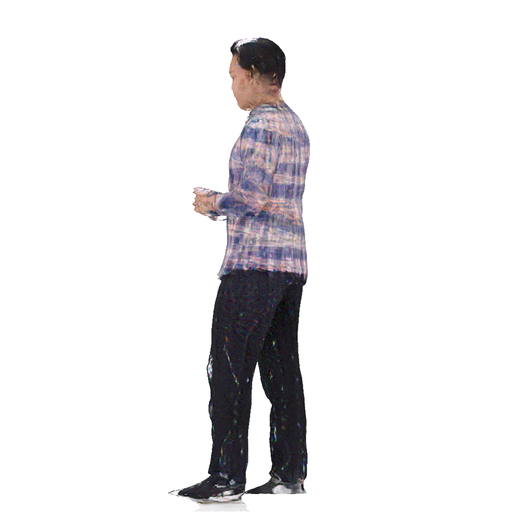}}\hfill
\mpage{0.045}{\includegraphics[width=\linewidth, trim=175 0 175 0, clip]{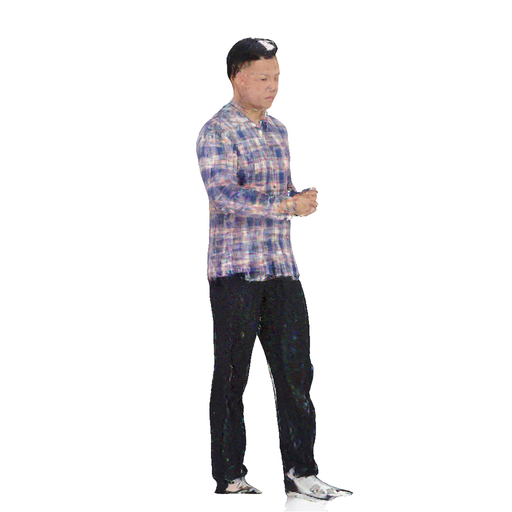}}\hfill
\hspace{2mm}
\mpage{0.06}{\includegraphics[width=\linewidth, trim=150 0 150 0, clip]{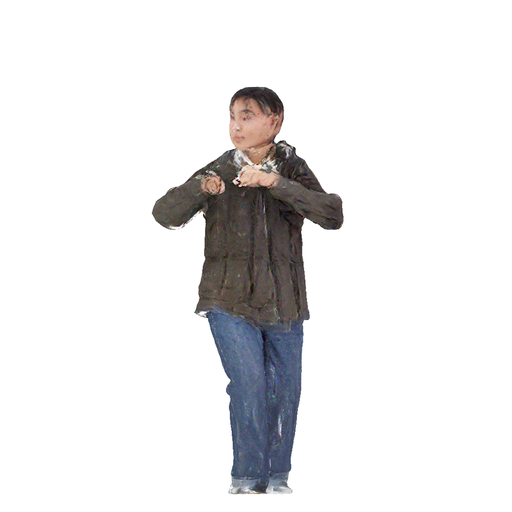}}\hfill
\hspace{-2mm}
\mpage{0.045}{\includegraphics[width=\linewidth, trim=175 0 175 0, clip]{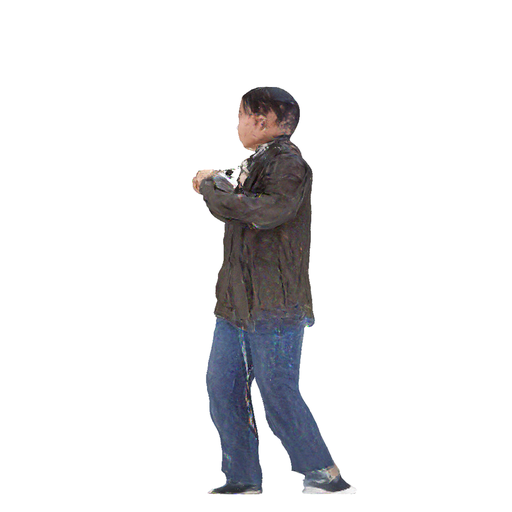}}\hfill
\mpage{0.045}{\includegraphics[width=\linewidth, trim=175 0 175 0, clip]{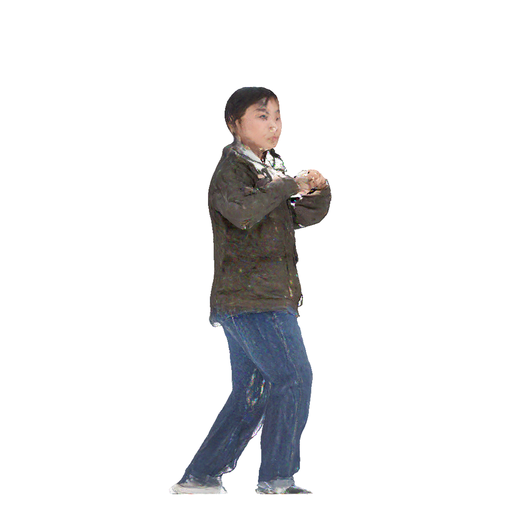}}\hfill
\hspace{2mm}
\mpage{0.07}{\includegraphics[width=\linewidth, trim=140 0 125 0, clip]{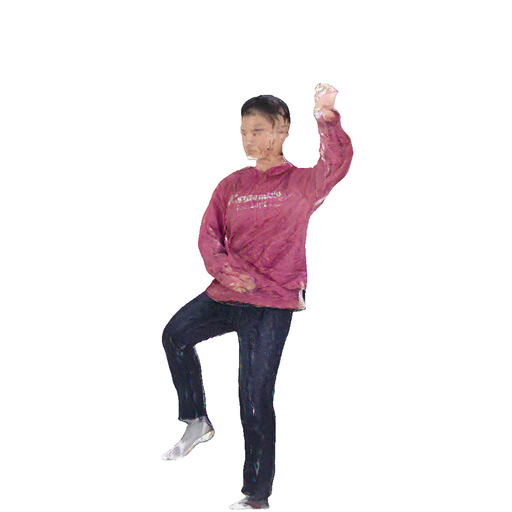}}\hfill
\mpage{0.045}{\includegraphics[width=\linewidth, trim=175 0 175 0, clip]{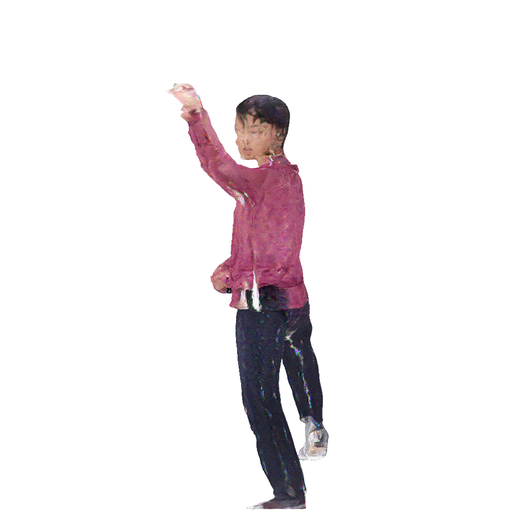}}\hfill
\mpage{0.07}{\includegraphics[width=\linewidth, trim=140 0 125 0, clip]{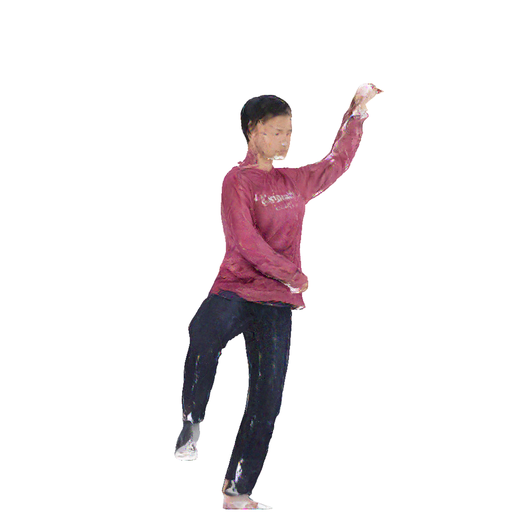}}\\

\vspace{-1mm}

\mpage{0.03}{\raisebox{0pt}{\rotatebox{90}{PIFu}}}  \hfill
\mpage{0.068}{\includegraphics[width=\linewidth, trim=125 0 145 0, clip]{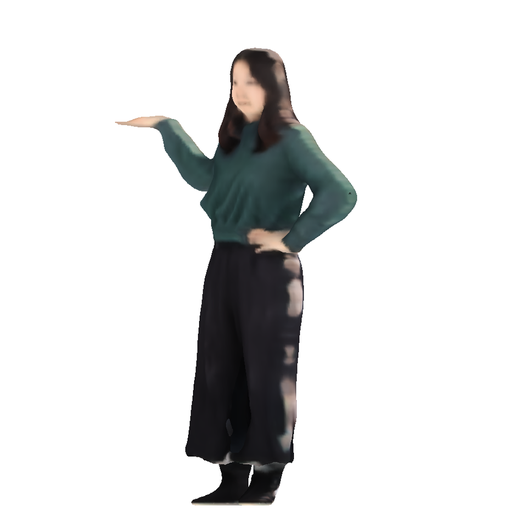}}\hfill
\hspace{-3mm}
\mpage{0.068}{\includegraphics[width=\linewidth, trim=125 0 145 0, clip]{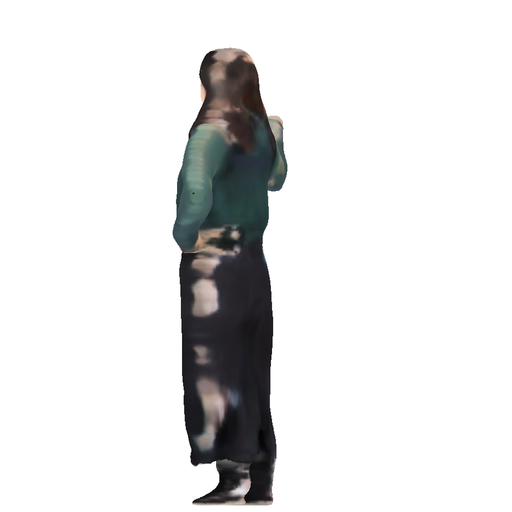}}\hfill
\hspace{-5mm}
\mpage{0.068}{\includegraphics[width=\linewidth, trim=125 0 145 0 0, clip]{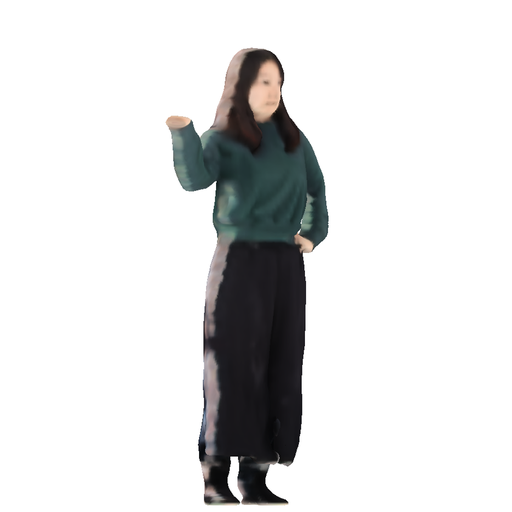}}\hfill
\hspace{2mm}
\mpage{0.045}{\includegraphics[width=\linewidth, trim=175 0 175 0, clip]{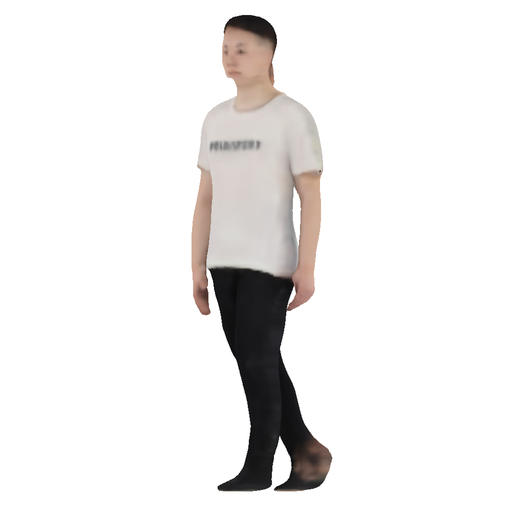}}\hfill
\mpage{0.045}{\includegraphics[width=\linewidth, trim=175 0 175 0, clip]{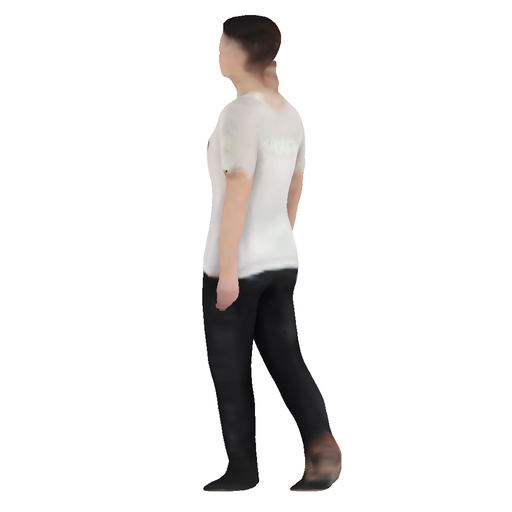}}\hfill
\mpage{0.045}{\includegraphics[width=\linewidth, trim=175 0 175 0, clip]{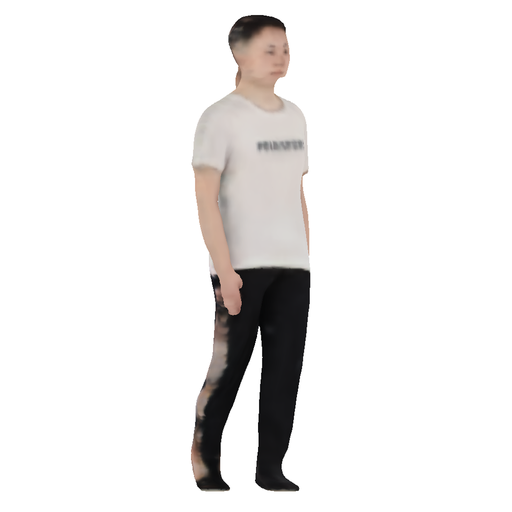}}\hfill
\hspace{2mm}
\mpage{0.045}{\includegraphics[width=\linewidth, trim=175 0 175 0, clip]{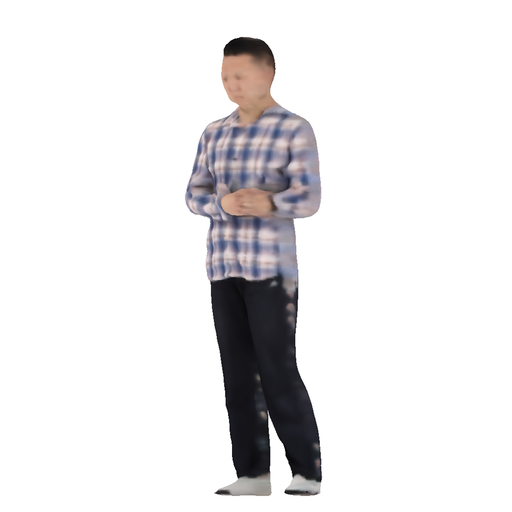}}\hfill
\mpage{0.045}{\includegraphics[width=\linewidth, trim=175 0 175 0, clip]{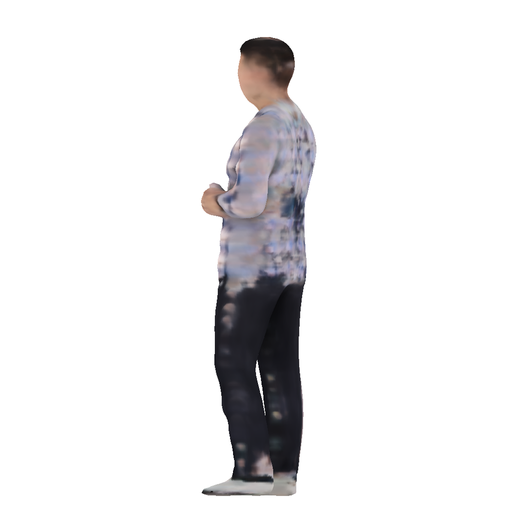}}\hfill
\mpage{0.045}{\includegraphics[width=\linewidth, trim=175 0 175 0, clip]{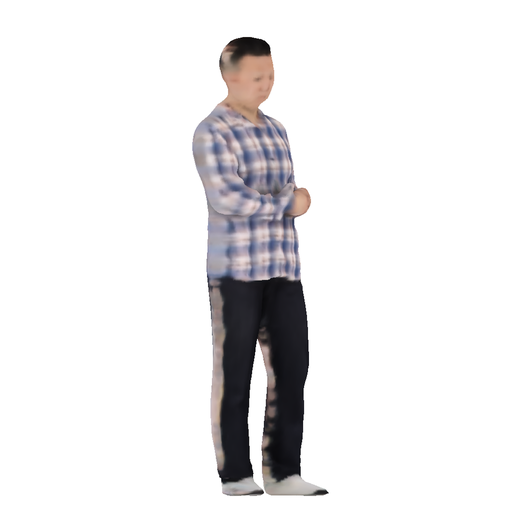}}\hfill
\hspace{2mm}
\mpage{0.06}{\includegraphics[width=\linewidth, trim=150 0 150 0, clip]{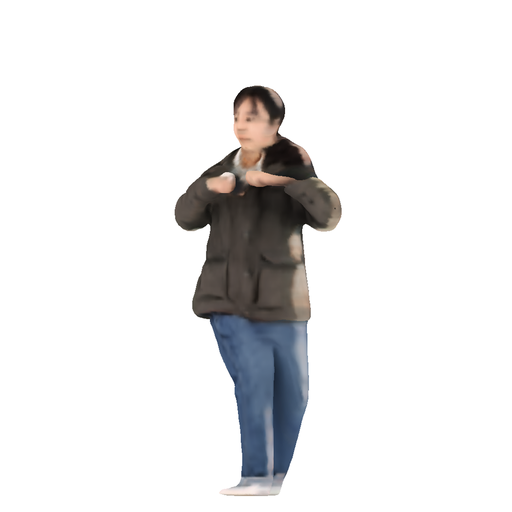}}\hfill
\hspace{-2mm}
\mpage{0.045}{\includegraphics[width=\linewidth, trim=175 0 175 0, clip]{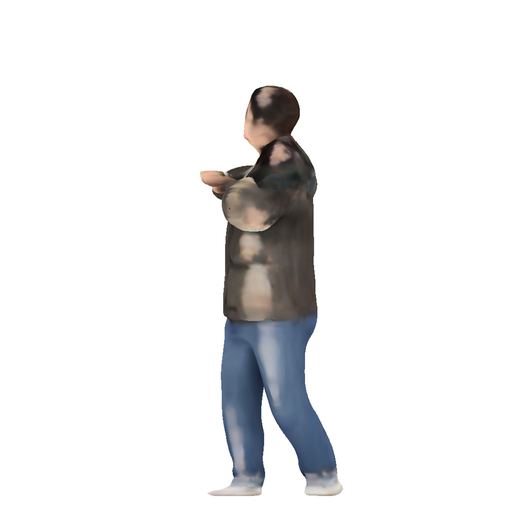}}\hfill
\mpage{0.045}{\includegraphics[width=\linewidth, trim=175 0 175 0, clip]{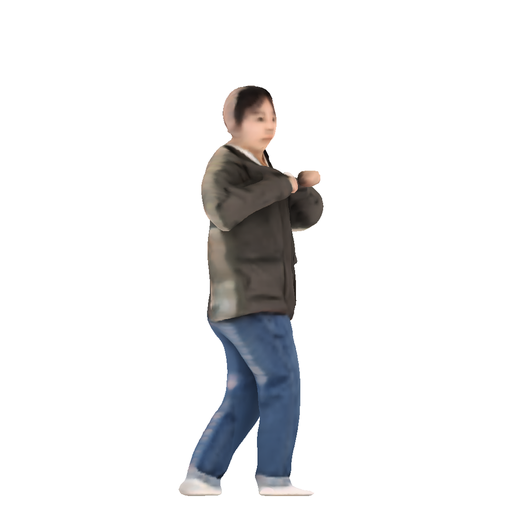}}\hfill
\hspace{2mm}
\mpage{0.07}{\includegraphics[width=\linewidth, trim=140 0 125 0, clip]{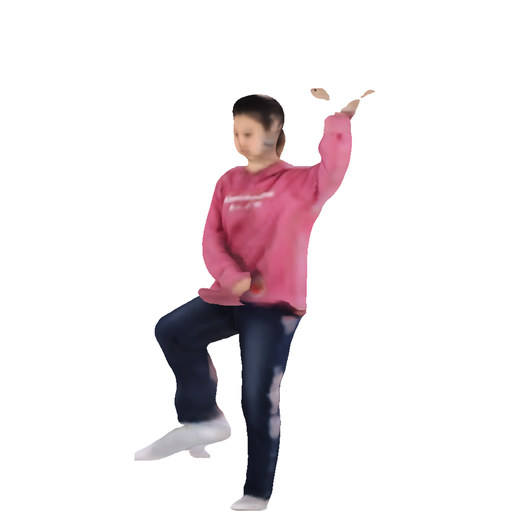}}\hfill
\mpage{0.045}{\includegraphics[width=\linewidth, trim=175 0 175 0, clip]{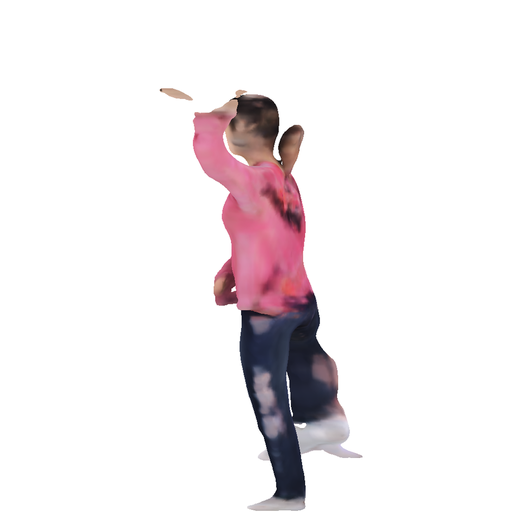}}\hfill
\mpage{0.07}{\includegraphics[width=\linewidth, trim=140 0 125 0, clip]{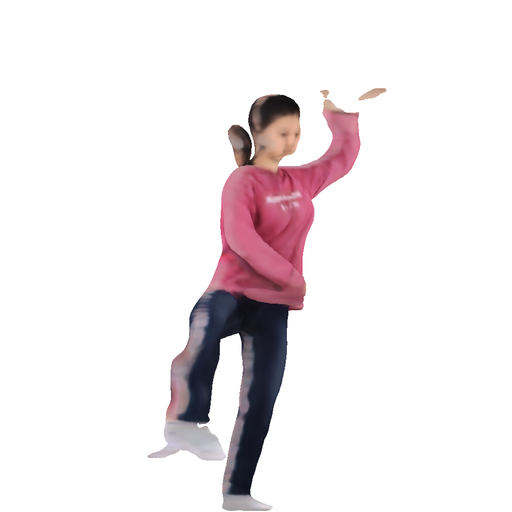}}\\

\vspace{-0.6mm}

\mpage{0.03}{\raisebox{0pt}{\rotatebox{90}{Impersonator++}}}  \hfill
\mpage{0.068}{\includegraphics[width=\linewidth, trim=125 0 145 0, clip]{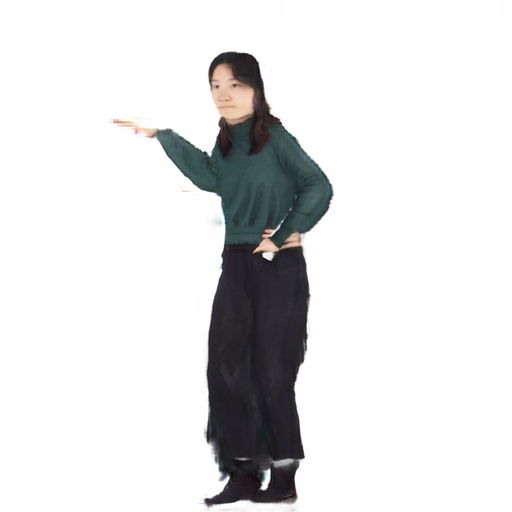}}\hfill
\hspace{-3mm}
\mpage{0.068}{\includegraphics[width=\linewidth, trim=125 0 145 0, clip]{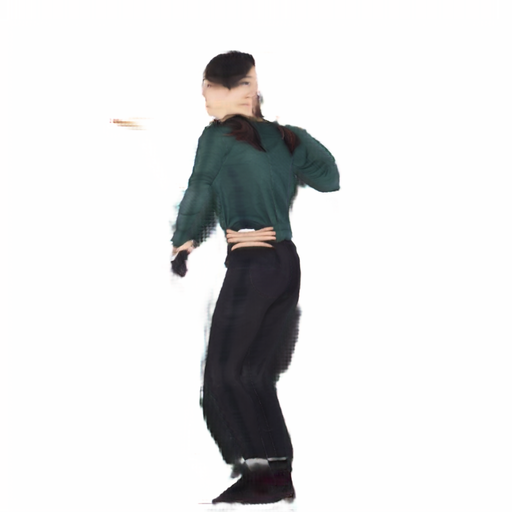}}\hfill
\hspace{-5mm}
\mpage{0.068}{\includegraphics[width=\linewidth, trim=125 0 145 0, clip]{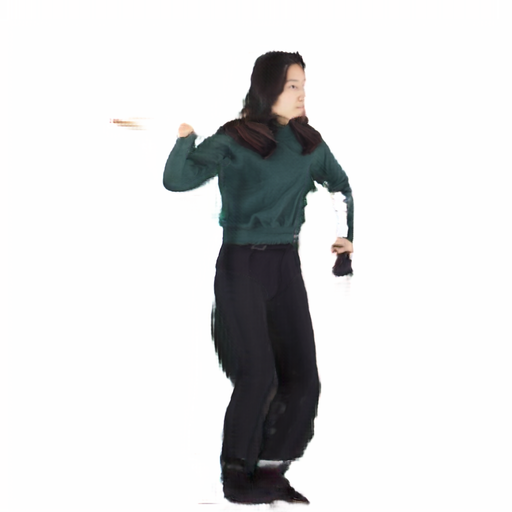}}\hfill
\hspace{2mm}
\mpage{0.045}{\includegraphics[width=\linewidth, trim=175 0 175 0, clip]{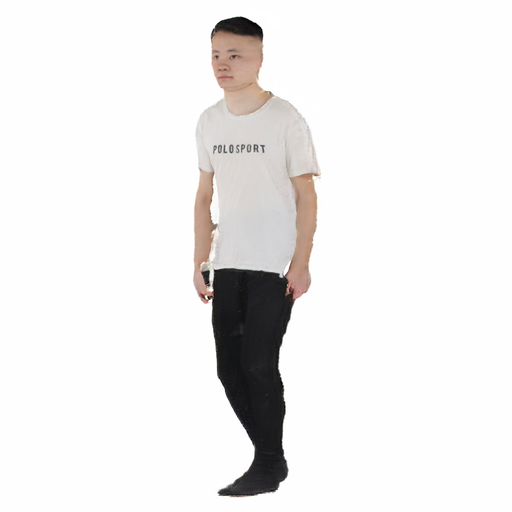}}\hfill
\mpage{0.045}{\includegraphics[width=\linewidth, trim=175 0 175 0, clip]{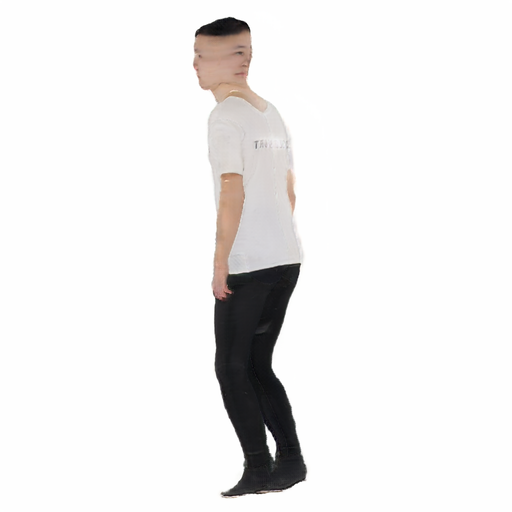}}\hfill
\mpage{0.045}{\includegraphics[width=\linewidth, trim=175 0 175 0, clip]{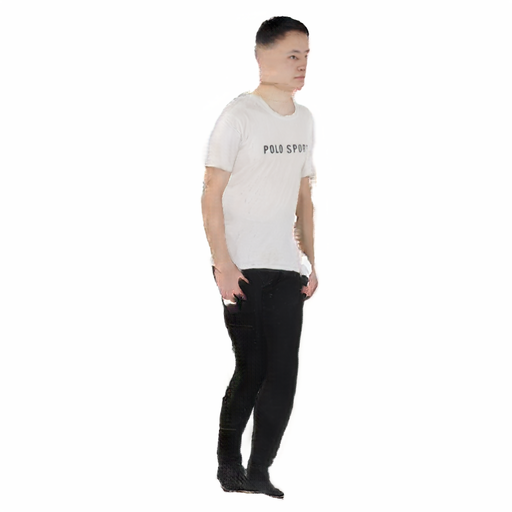}}\hfill
\hspace{2mm}
\mpage{0.045}{\includegraphics[width=\linewidth, trim=175 0 175 0, clip]{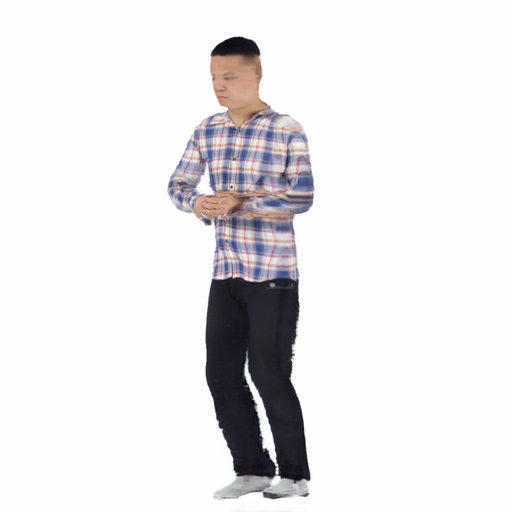}}\hfill
\mpage{0.045}{\includegraphics[width=\linewidth, trim=175 0 175 0, clip]{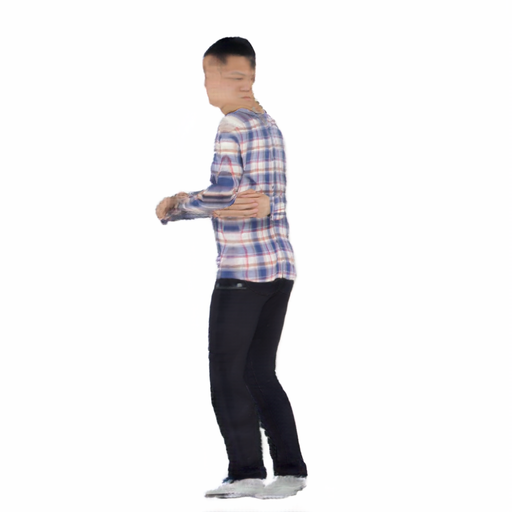}}\hfill
\mpage{0.045}{\includegraphics[width=\linewidth, trim=175 0 175 0, clip]{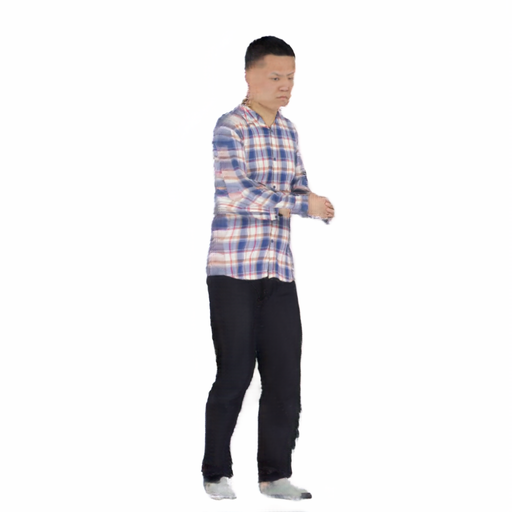}}\hfill
\hspace{2mm}
\mpage{0.06}{\includegraphics[width=\linewidth, trim=150 0 150 0, clip]{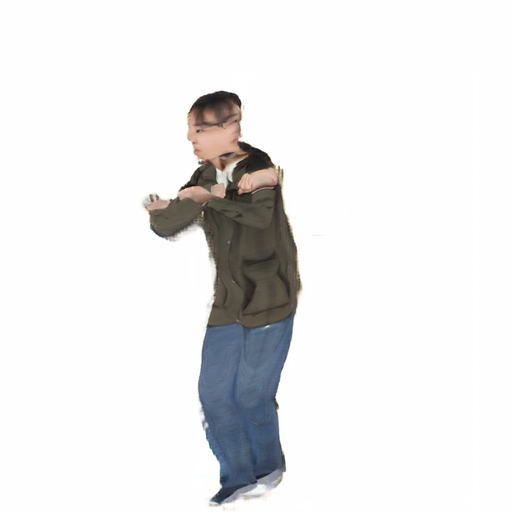}}\hfill
\hspace{-2mm}
\mpage{0.045}{\includegraphics[width=\linewidth, trim=175 0 175 0, clip]{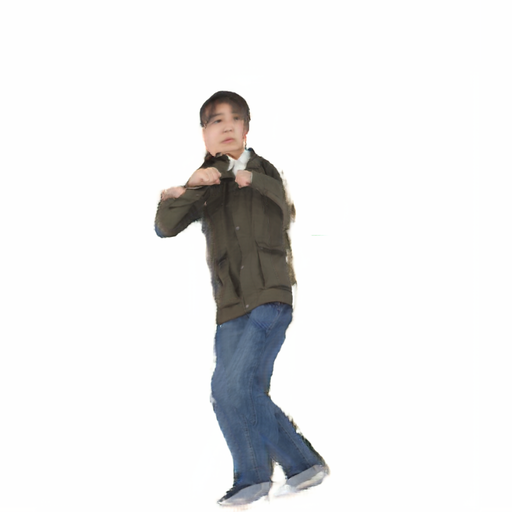}}\hfill
\mpage{0.045}{\includegraphics[width=\linewidth, trim=175 0 175 0, clip]{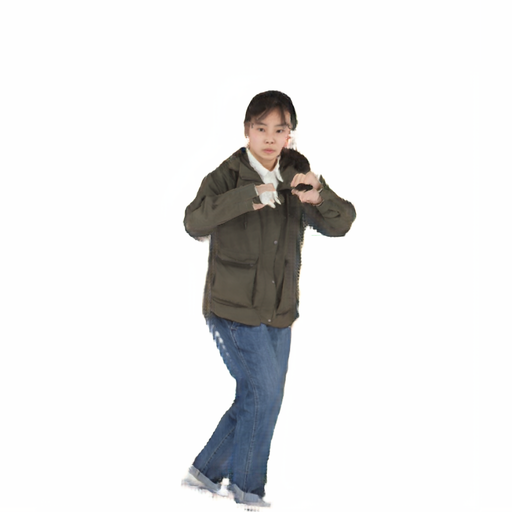}}\hfill
\hspace{2mm}
\mpage{0.07}{\includegraphics[width=\linewidth, trim=140 0 125 0, clip]{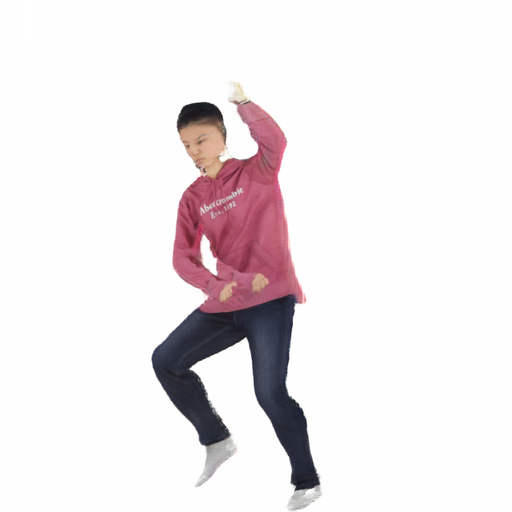}}\hfill
\mpage{0.045}{\includegraphics[width=\linewidth, trim=175 0 175 0, clip]{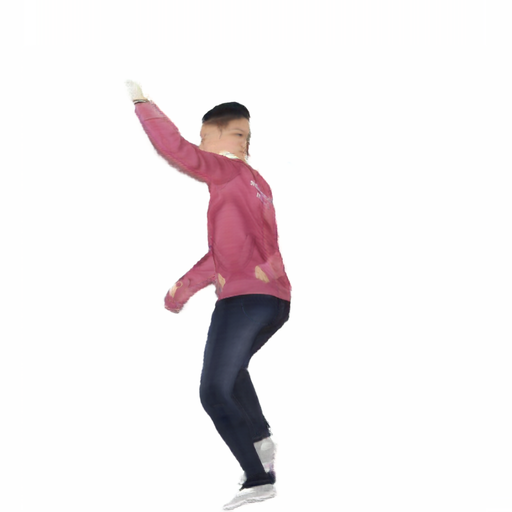}}\hfill
\mpage{0.07}{\includegraphics[width=\linewidth, trim=140 0 125 0, clip]{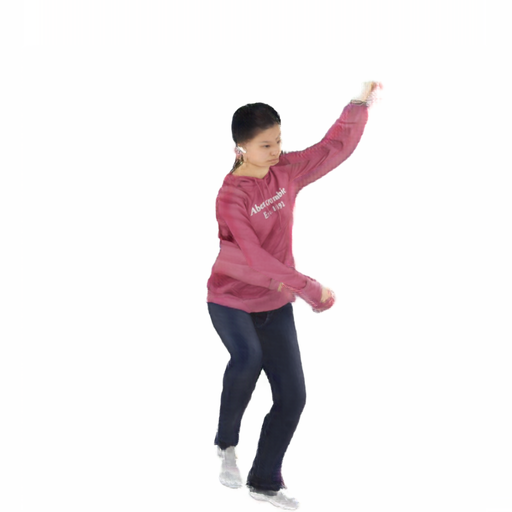}}\\

\vspace{-0.6mm}

\mpage{0.03}{\raisebox{0pt}{\rotatebox{90}{TEXTure}}}  \hfill
\mpage{0.068}{\includegraphics[width=\linewidth, trim=125 0 145 0, clip]{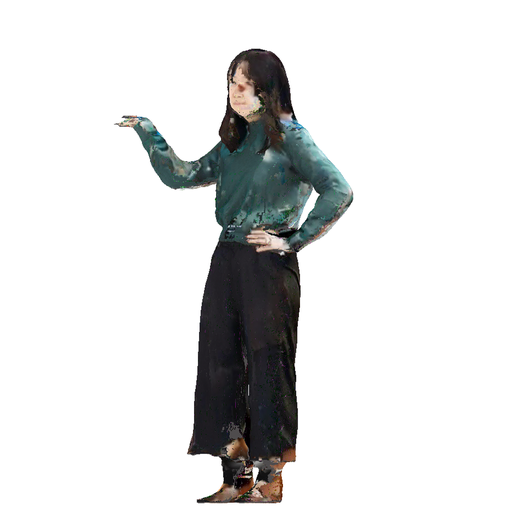}}\hfill
\hspace{-3mm}
\mpage{0.068}{\includegraphics[width=\linewidth, trim=125 0 145 0, clip]{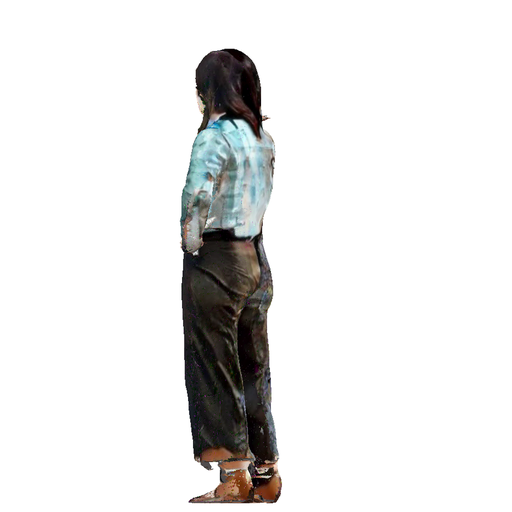}}\hfill
\hspace{-5mm}
\mpage{0.068}{\includegraphics[width=\linewidth, trim=125 0 145 0, clip]{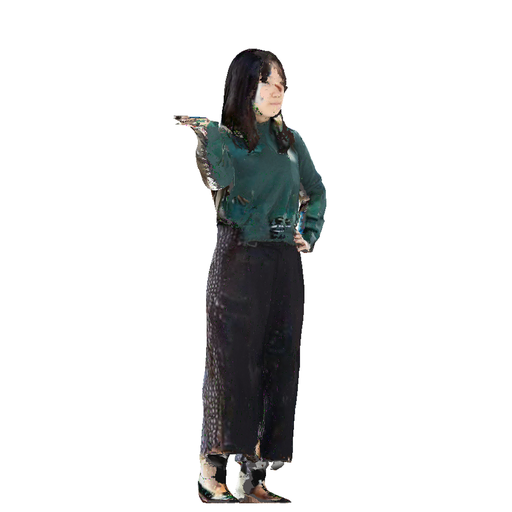}}\hfill
\hspace{2mm}
\mpage{0.045}{\includegraphics[width=\linewidth, trim=175 0 175 0, clip]{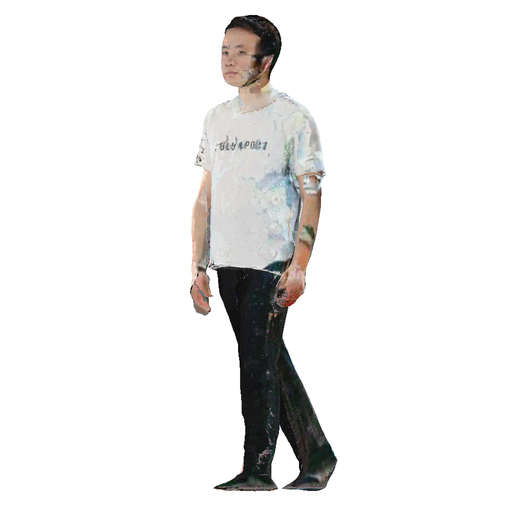}}\hfill
\mpage{0.045}{\includegraphics[width=\linewidth, trim=175 0 175 0, clip]{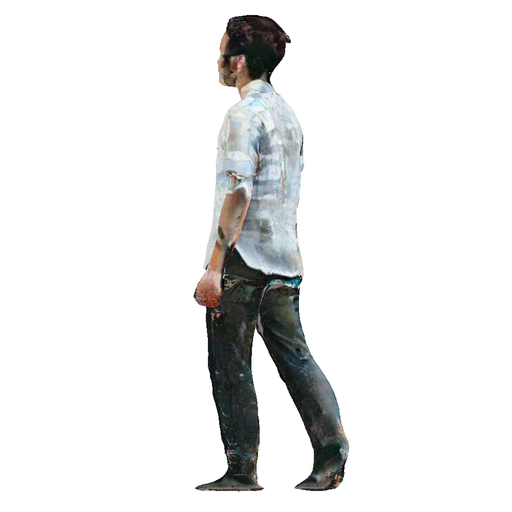}}\hfill
\mpage{0.045}{\includegraphics[width=\linewidth, trim=175 0 175 0, clip]{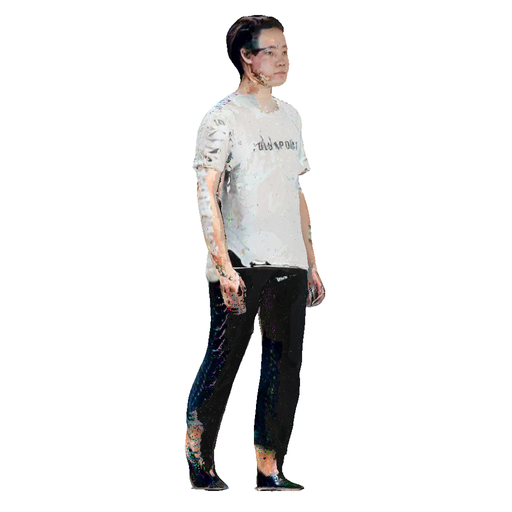}}\hfill
\hspace{2mm}
\mpage{0.045}{\includegraphics[width=\linewidth, trim=175 0 175 0, clip]{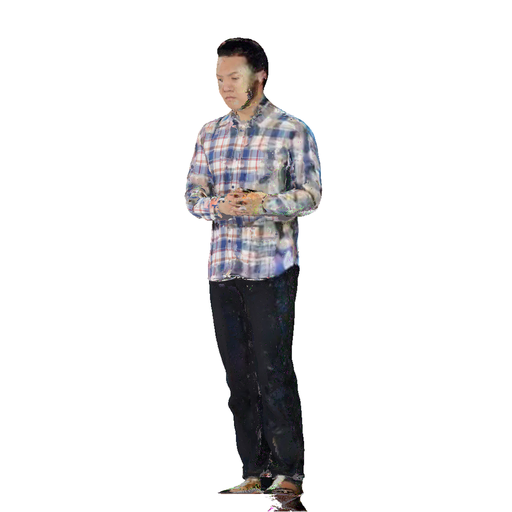}}\hfill
\mpage{0.045}{\includegraphics[width=\linewidth, trim=175 0 175 0, clip]{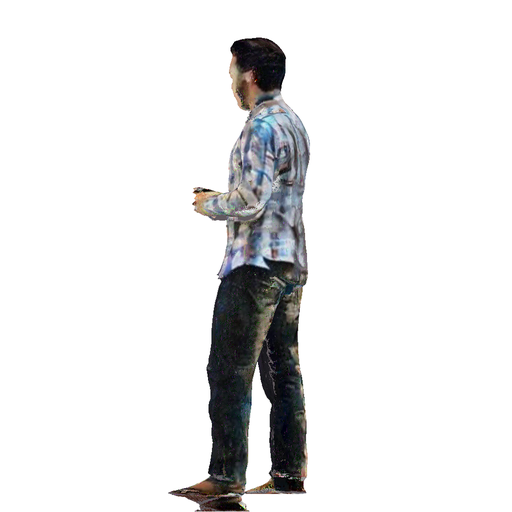}}\hfill
\mpage{0.045}{\includegraphics[width=\linewidth, trim=175 0 175 0, clip]{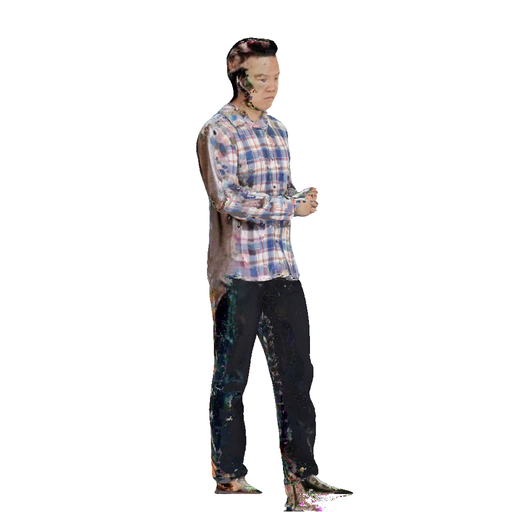}}\hfill
\hspace{2mm}
\mpage{0.06}{\includegraphics[width=\linewidth, trim=150 0 150 0, clip]{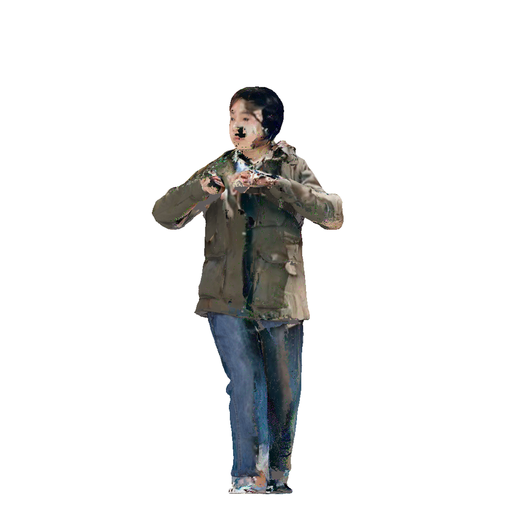}}\hfill
\hspace{-2mm}
\mpage{0.045}{\includegraphics[width=\linewidth, trim=175 0 175 0, clip]{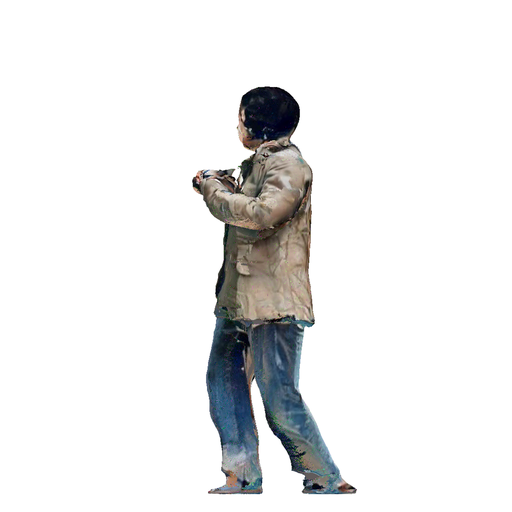}}\hfill
\mpage{0.045}{\includegraphics[width=\linewidth, trim=175 0 175 0, clip]{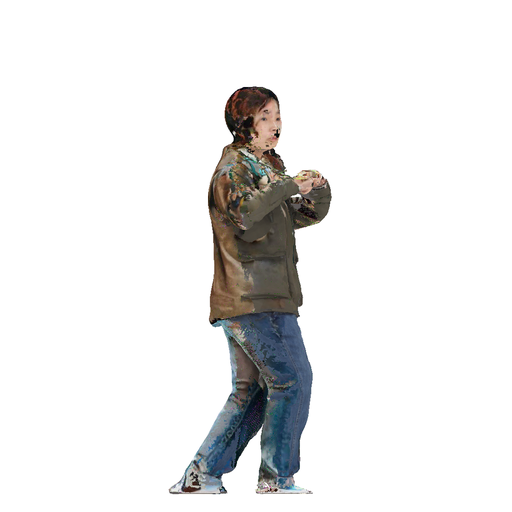}}\hfill
\hspace{2mm}
\mpage{0.07}{\includegraphics[width=\linewidth, trim=140 0 125 0, clip]{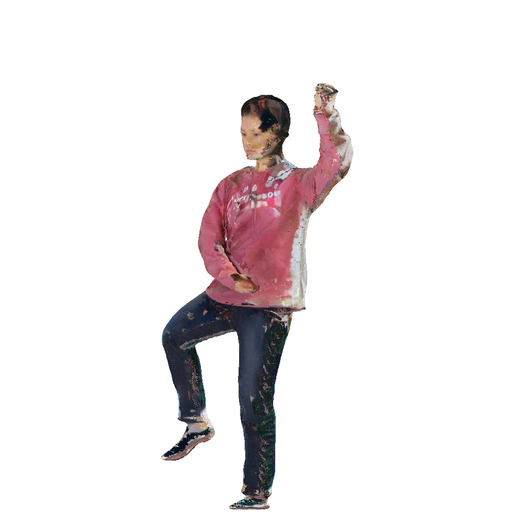}}\hfill
\mpage{0.045}{\includegraphics[width=\linewidth, trim=175 0 175 0, clip]{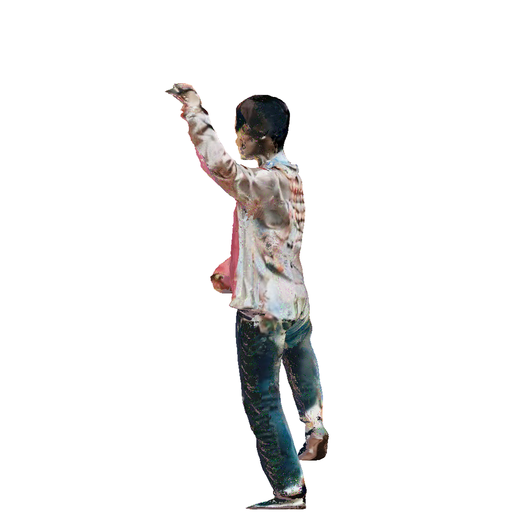}}\hfill
\mpage{0.07}{\includegraphics[width=\linewidth, trim=140 0 125 0, clip]{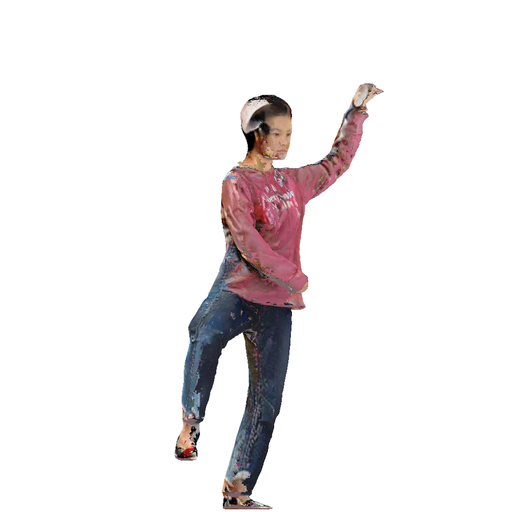}}\\

\vspace{-0.6mm}

\mpage{0.03}{\raisebox{0pt}{\rotatebox{90}{Magic123}}}  \hfill
\mpage{0.068}{\includegraphics[width=\linewidth, trim=125 0 145 0, clip]{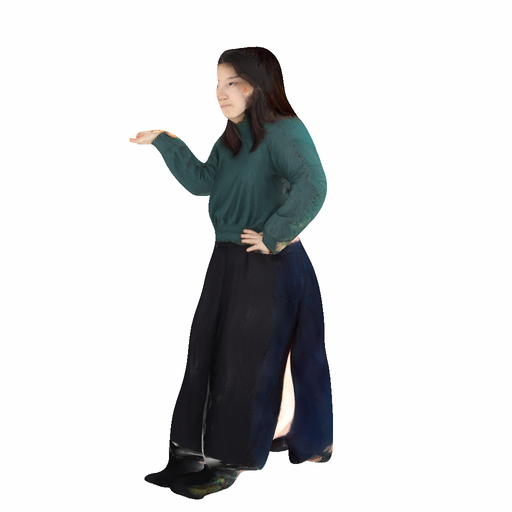}}\hfill
\hspace{-3mm}
\mpage{0.068}{\includegraphics[width=\linewidth, trim=125 0 145 0, clip]{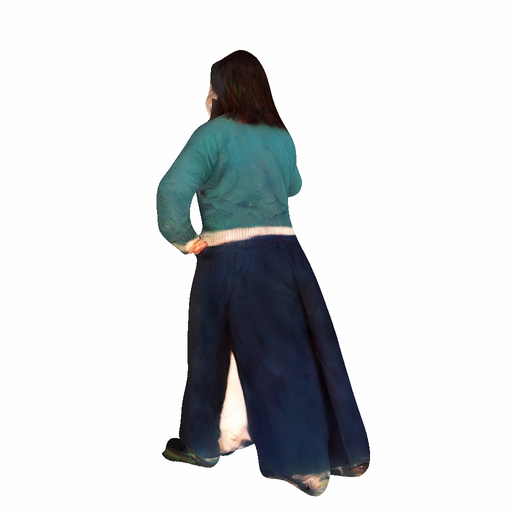}}\hfill
\hspace{-5mm}
\mpage{0.068}{\includegraphics[width=\linewidth, trim=125 0 145 0, clip]{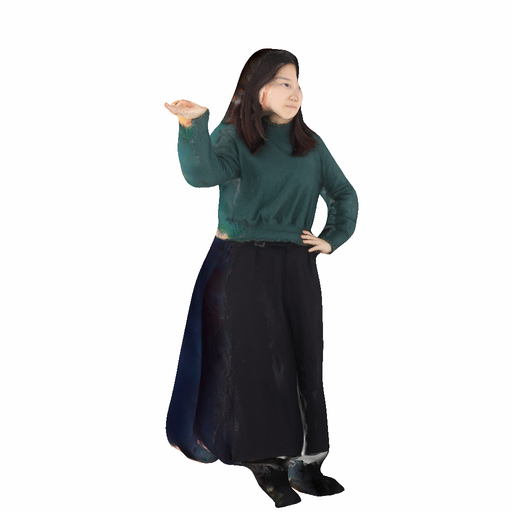}}\hfill
\hspace{2mm}
\mpage{0.045}{\includegraphics[width=\linewidth, trim=175 0 175 0, clip]{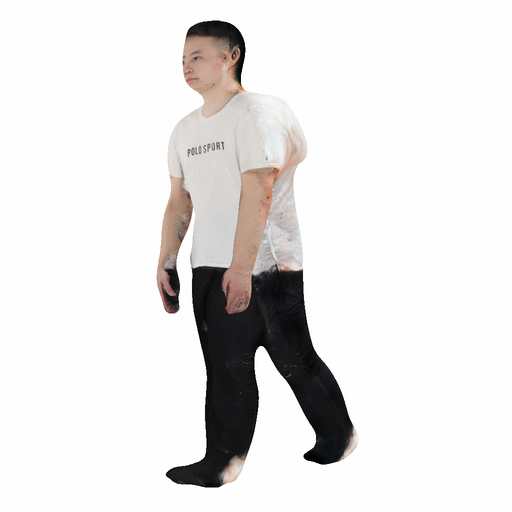}}\hfill
\mpage{0.045}{\includegraphics[width=\linewidth, trim=175 0 175 0, clip]{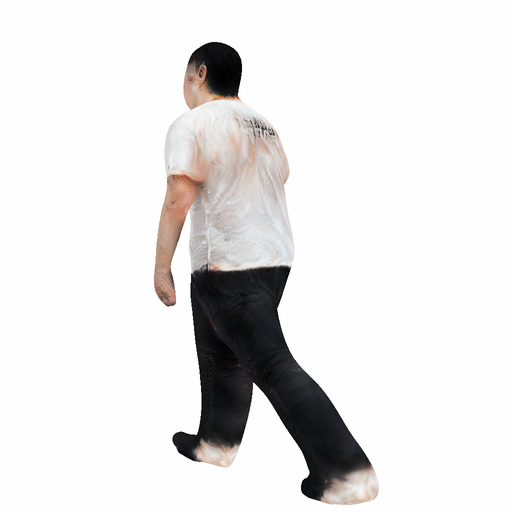}}\hfill
\mpage{0.045}{\includegraphics[width=\linewidth, trim=175 0 175 0, clip]{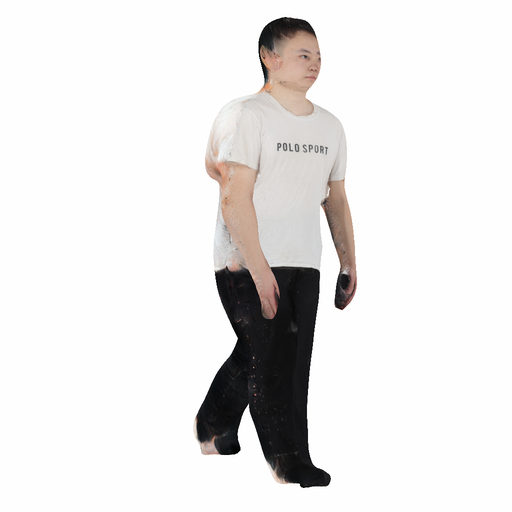}}\hfill
\hspace{2mm}
\mpage{0.045}{\includegraphics[width=\linewidth, trim=175 0 175 0, clip]{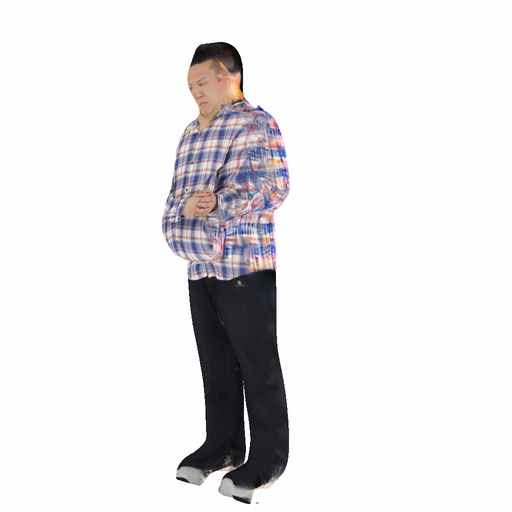}}\hfill
\mpage{0.045}{\includegraphics[width=\linewidth, trim=175 0 175 0, clip]{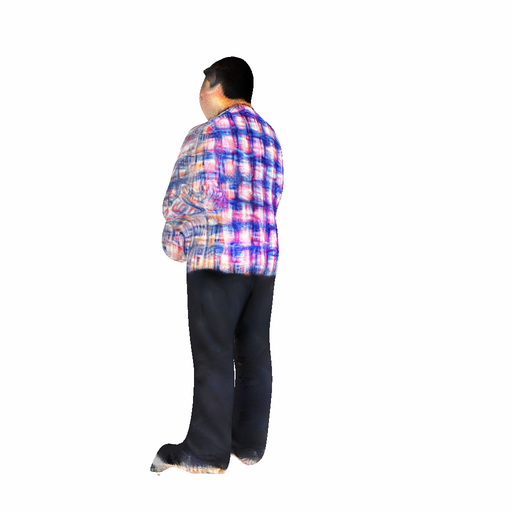}}\hfill
\mpage{0.045}{\includegraphics[width=\linewidth, trim=175 0 175 0, clip]{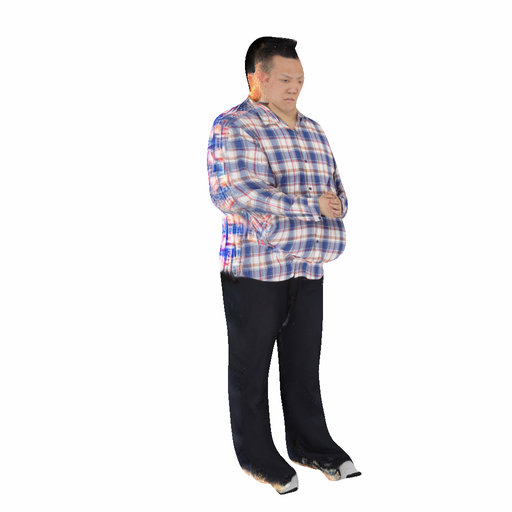}}\hfill
\hspace{2mm}
\mpage{0.045}{\includegraphics[width=\linewidth, trim=175 0 175 0, clip]{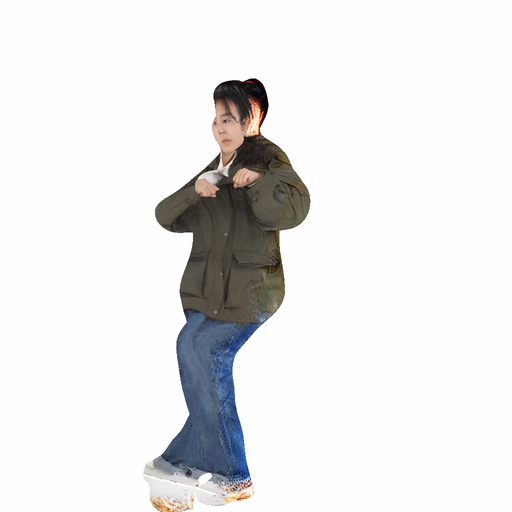}}\hfill
\mpage{0.045}{\includegraphics[width=\linewidth, trim=175 0 175 0, clip]{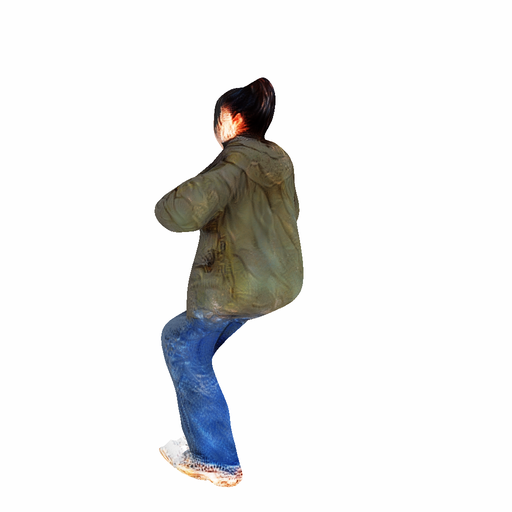}}\hfill
\mpage{0.045}{\includegraphics[width=\linewidth, trim=175 0 175 0, clip]{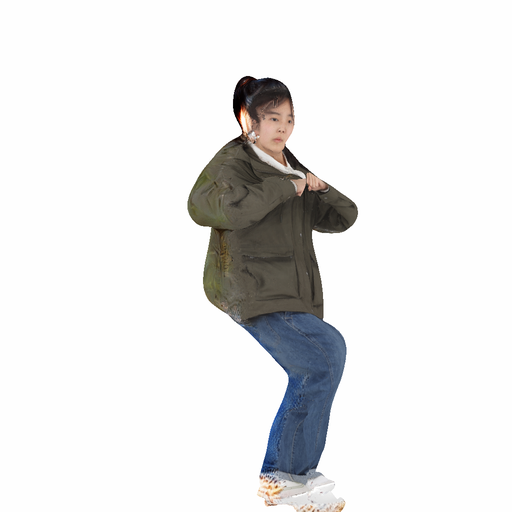}}\hfill
\hspace{2mm}
\mpage{0.07}{\includegraphics[width=\linewidth, trim=140 0 125 0, clip]{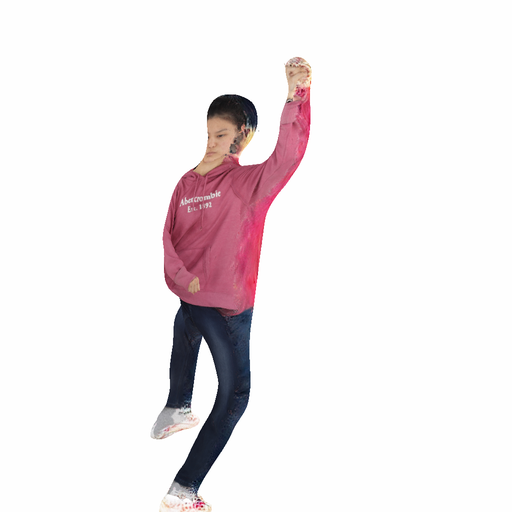}}\hfill
\mpage{0.045}{\includegraphics[width=\linewidth, trim=175 0 175 0, clip]{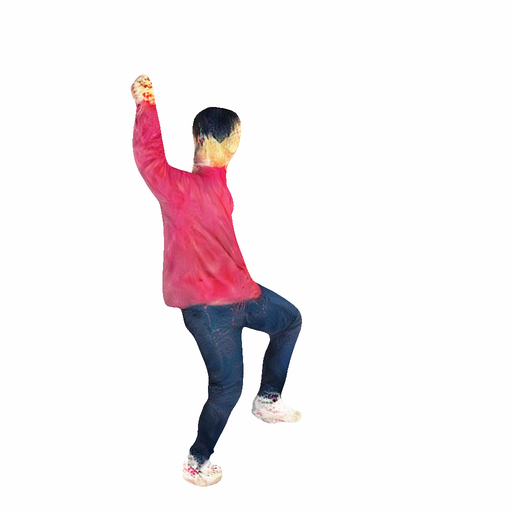}}\hfill
\mpage{0.07}{\includegraphics[width=\linewidth, trim=140 0 125 0, clip]{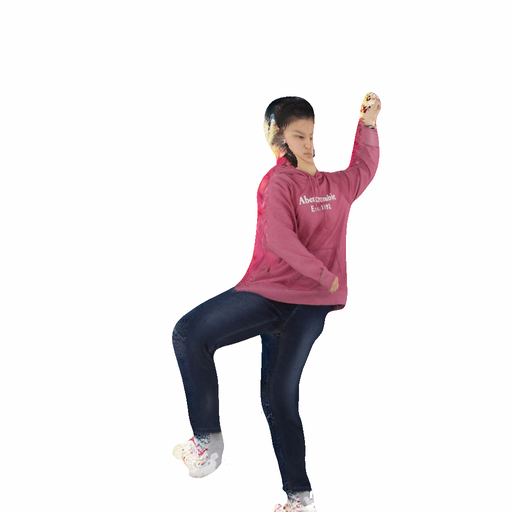}}\\

\vspace{-0.6mm}

\mpage{0.03}{\raisebox{0pt}{\rotatebox{90}{S3F}}}  \hfill
\mpage{0.068}{\includegraphics[width=\linewidth, trim=125 0 145 0, clip]{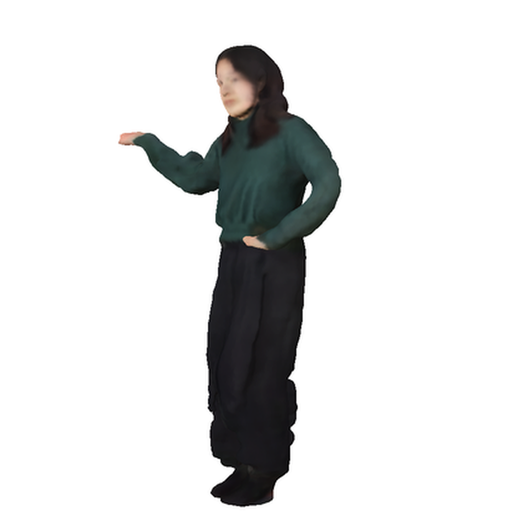}}\hfill
\hspace{-3mm}
\mpage{0.068}{\includegraphics[width=\linewidth, trim=125 0 145 0, clip]{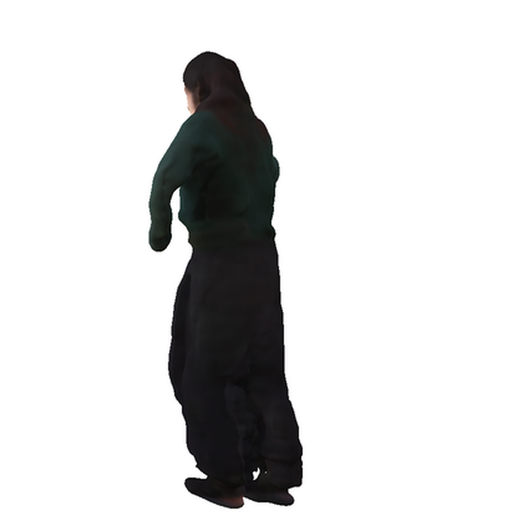}}\hfill
\hspace{-5mm}
\mpage{0.068}{\includegraphics[width=\linewidth, trim=125 0 145 0, clip]{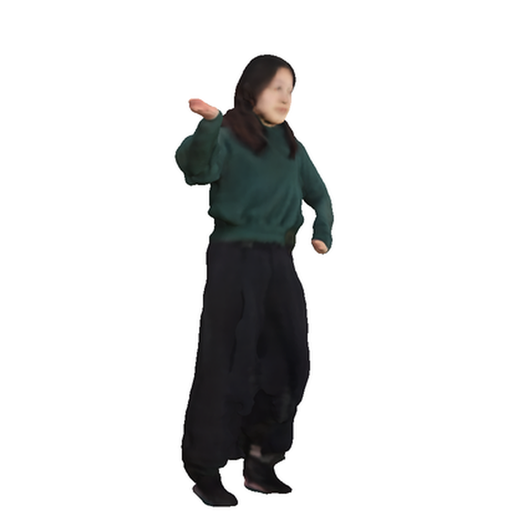}}\hfill
\hspace{2mm}
\mpage{0.045}{\includegraphics[width=\linewidth, trim=175 0 175 0, clip]{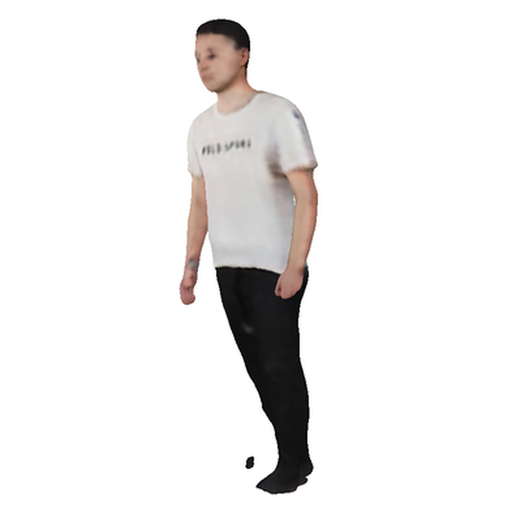}}\hfill
\mpage{0.045}{\includegraphics[width=\linewidth, trim=175 0 175 0, clip]{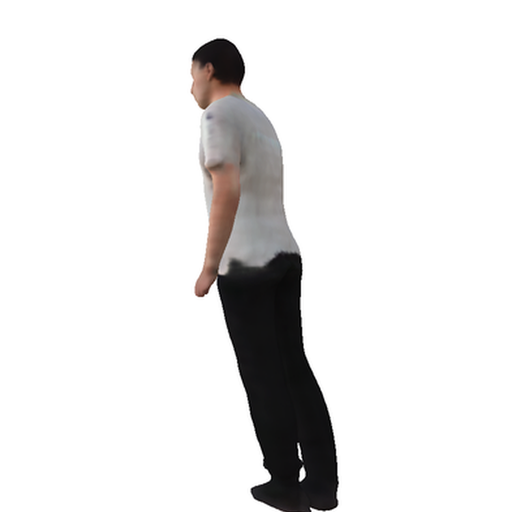}}\hfill
\mpage{0.045}{\includegraphics[width=\linewidth, trim=175 0 175 0, clip]{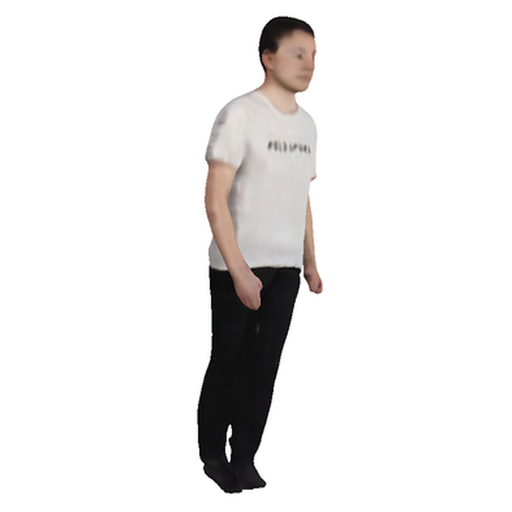}}\hfill
\hspace{2mm}
\mpage{0.045}{\includegraphics[width=\linewidth, trim=175 0 175 0, clip]{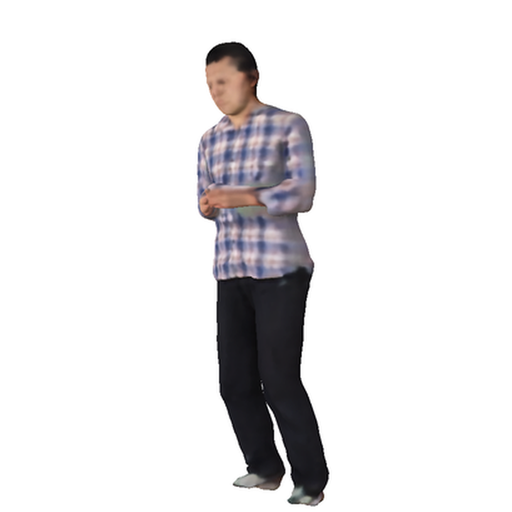}}\hfill
\mpage{0.045}{\includegraphics[width=\linewidth, trim=175 0 175 0, clip]{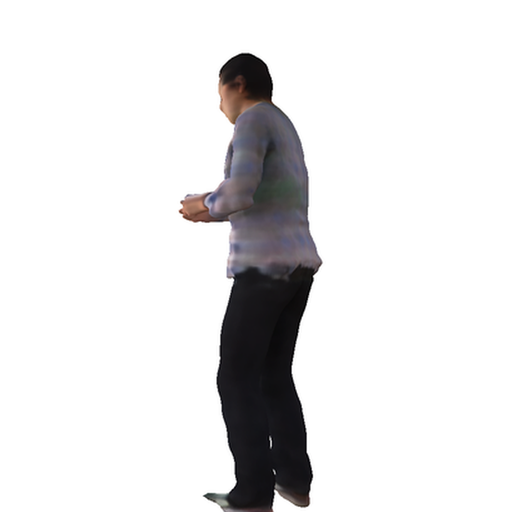}}\hfill
\mpage{0.045}{\includegraphics[width=\linewidth, trim=175 0 175 0, clip]{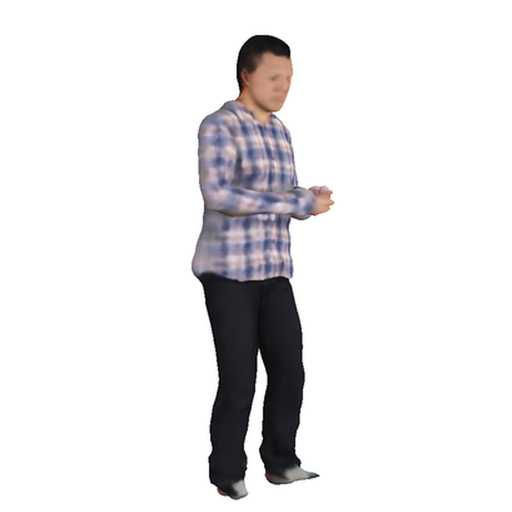}}\hfill
\hspace{2mm}
\mpage{0.06}{\includegraphics[width=\linewidth, trim=150 0 150 0, clip]{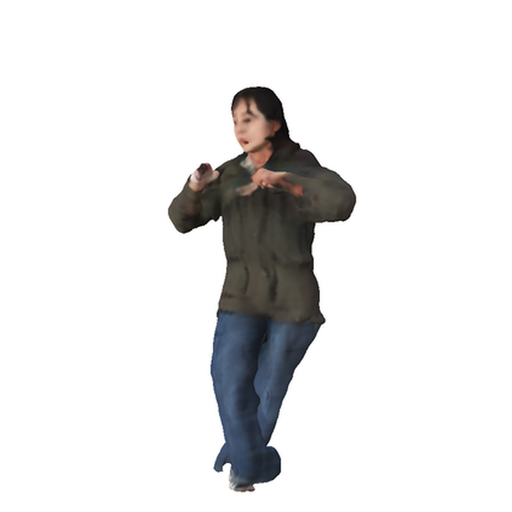}}\hfill
\hspace{-2mm}
\mpage{0.045}{\includegraphics[width=\linewidth, trim=175 0 175 0, clip]{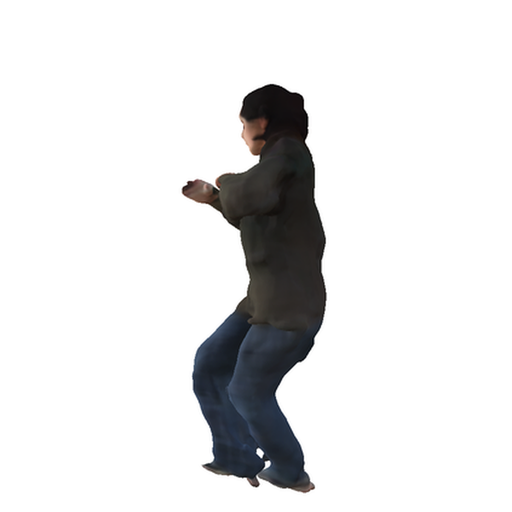}}\hfill
\mpage{0.045}{\includegraphics[width=\linewidth, trim=175 0 175 0, clip]{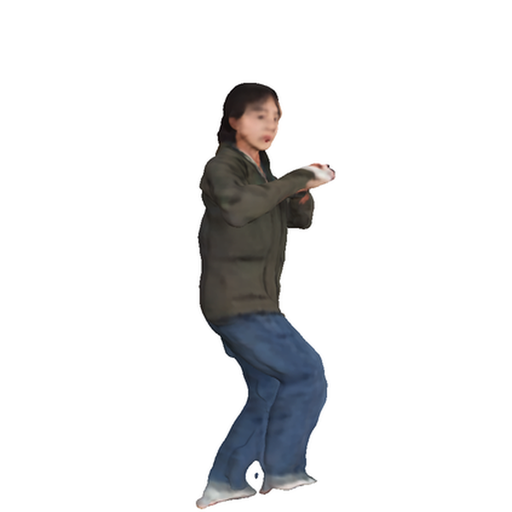}}\hfill
\hspace{2mm}
\mpage{0.07}{\includegraphics[width=\linewidth, trim=140 0 125 0, clip]{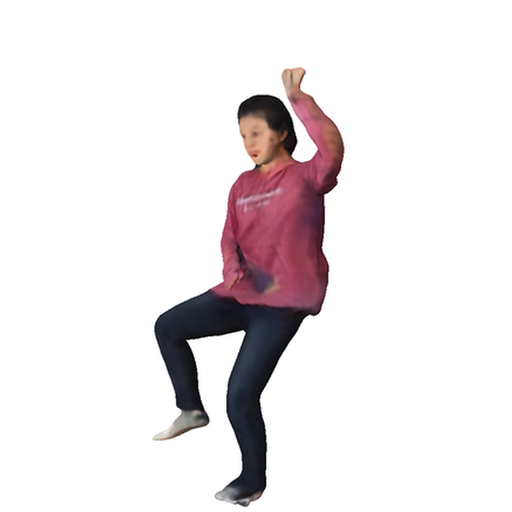}}\hfill
\mpage{0.045}{\includegraphics[width=\linewidth, trim=175 0 175 0, clip]{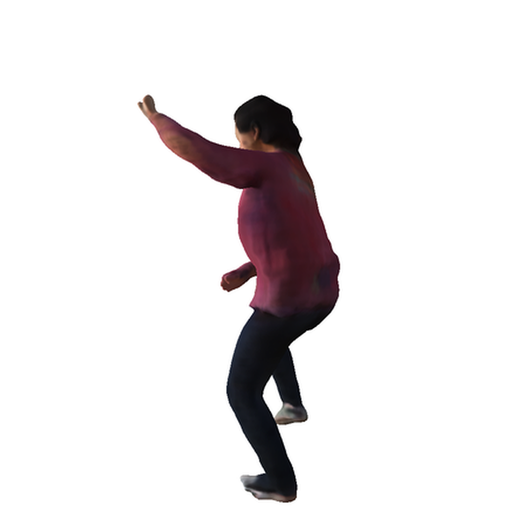}}\hfill
\mpage{0.07}{\includegraphics[width=\linewidth, trim=140 0 125 0, clip]{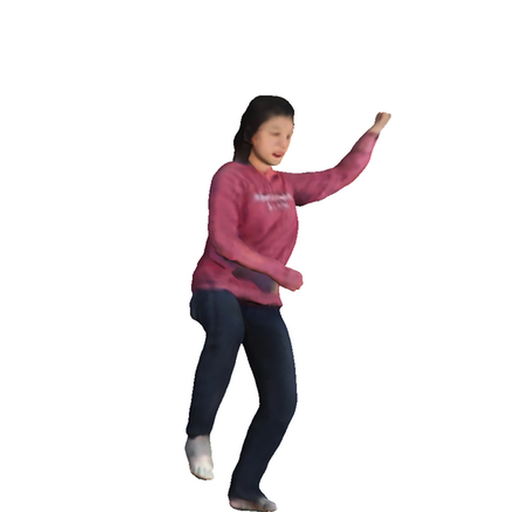}}\\

\vspace{-0.6mm}

\mpage{0.03}{\raisebox{0pt}{\rotatebox{90}{Ours}}}  \hfill
\mpage{0.068}{\includegraphics[width=\linewidth, trim=125 0 145 0, clip]{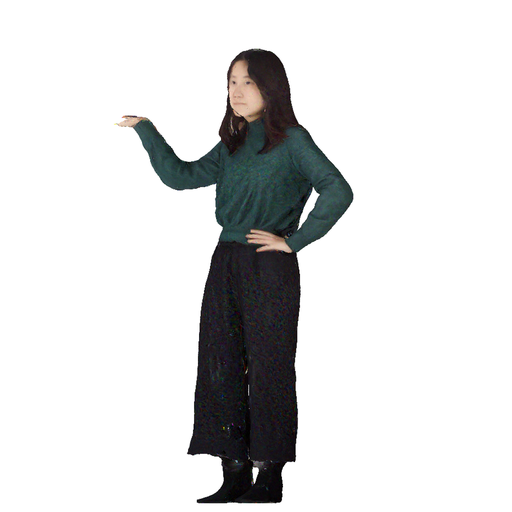}}\hfill
\hspace{-3mm}
\mpage{0.068}{\includegraphics[width=\linewidth, trim=125 0 145 0, clip]{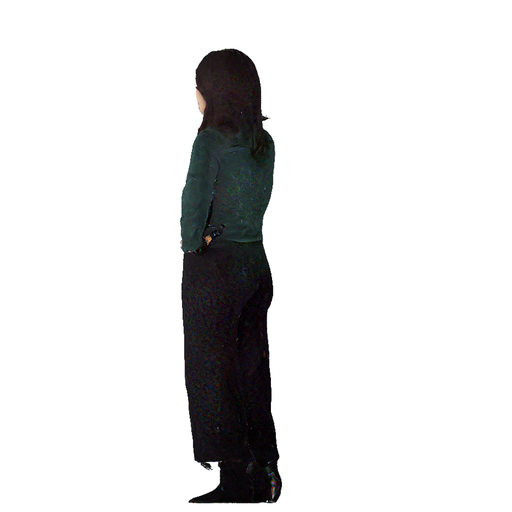}}\hfill
\hspace{-5mm}
\mpage{0.068}{\includegraphics[width=\linewidth, trim=125 0 145 0, clip]{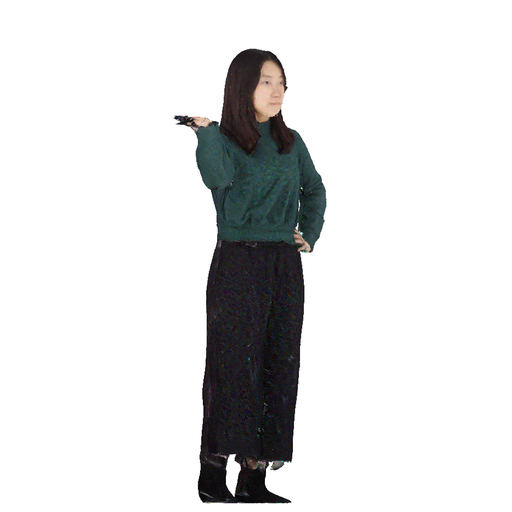}}\hfill
\hspace{2mm}
\mpage{0.045}{\includegraphics[width=\linewidth, trim=175 0 175 0, clip]{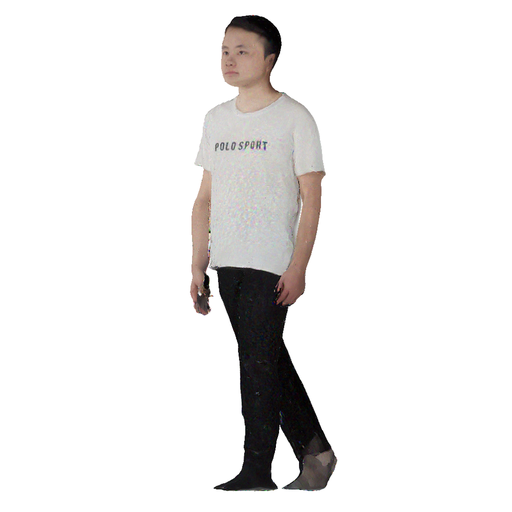}}\hfill
\mpage{0.045}{\includegraphics[width=\linewidth, trim=175 0 175 0, clip]{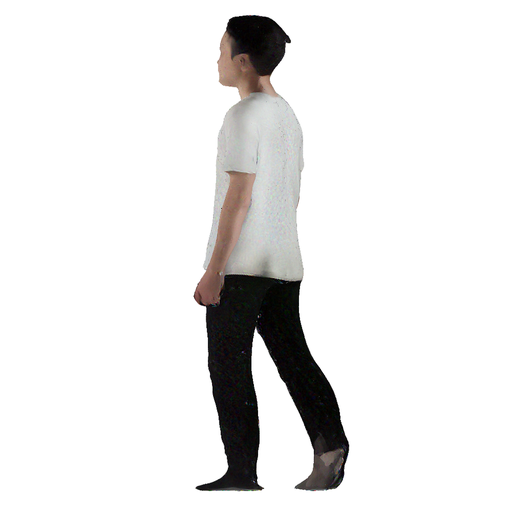}}\hfill
\mpage{0.045}{\includegraphics[width=\linewidth, trim=175 0 175 0, clip]{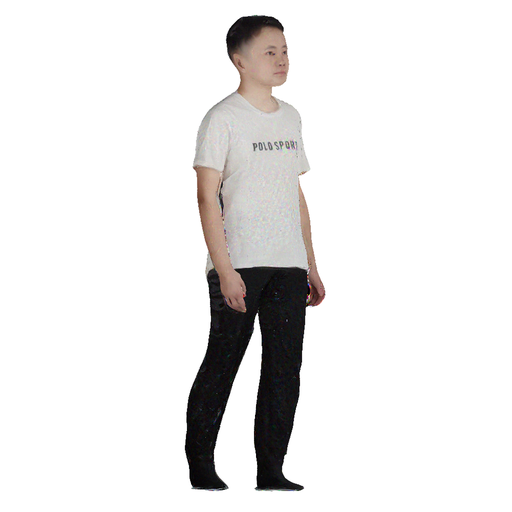}}\hfill
\hspace{2mm}
\mpage{0.045}{\includegraphics[width=\linewidth, trim=175 0 175 0, clip]{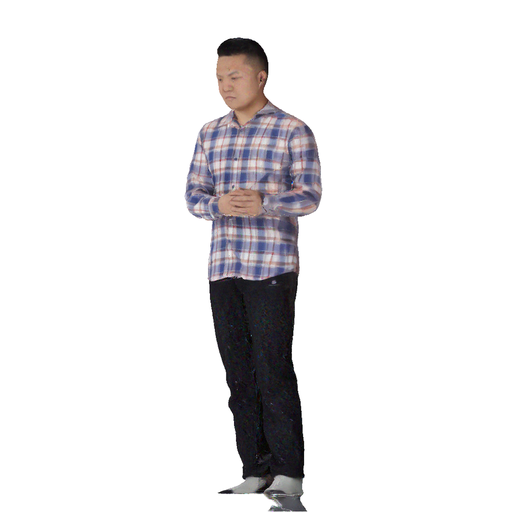}}\hfill
\mpage{0.045}{\includegraphics[width=\linewidth, trim=175 0 175 0, clip]{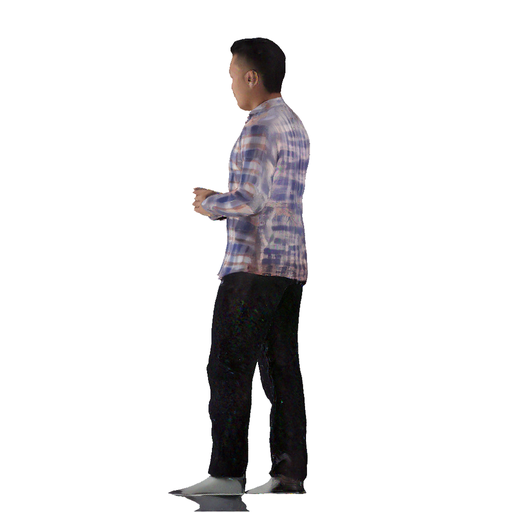}}\hfill
\mpage{0.045}{\includegraphics[width=\linewidth, trim=175 0 175 0, clip]{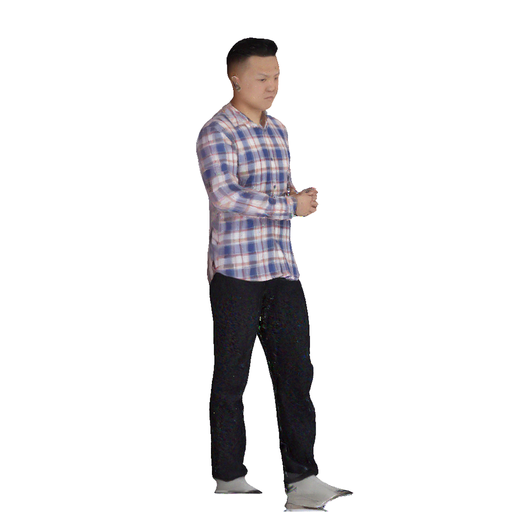}}\hfill
\hspace{2mm}
\mpage{0.06}{\includegraphics[width=\linewidth, trim=150 0 150 0, clip]{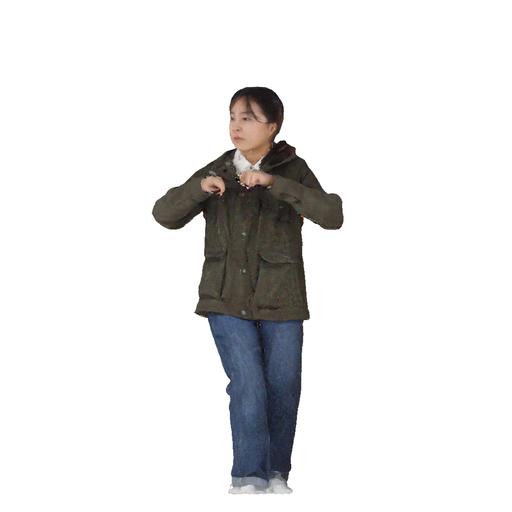}}\hfill
\hspace{-2mm}
\mpage{0.045}{\includegraphics[width=\linewidth, trim=175 0 175 0, clip]{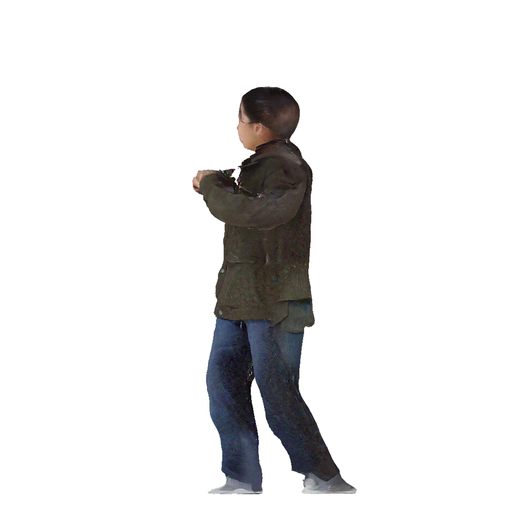}}\hfill
\mpage{0.045}{\includegraphics[width=\linewidth, trim=175 0 175 0, clip]{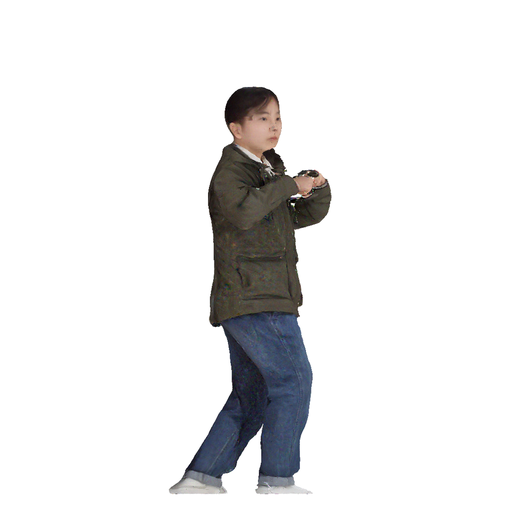}}\hfill
\hspace{2mm}
\mpage{0.07}{\includegraphics[width=\linewidth, trim=140 0 125 0, clip]{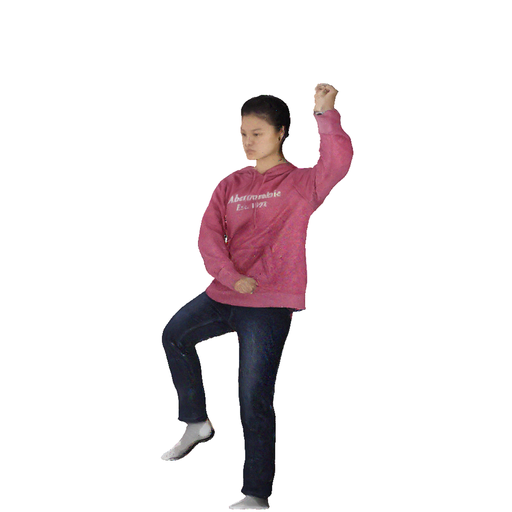}}\hfill
\mpage{0.045}{\includegraphics[width=\linewidth, trim=175 0 175 0, clip]{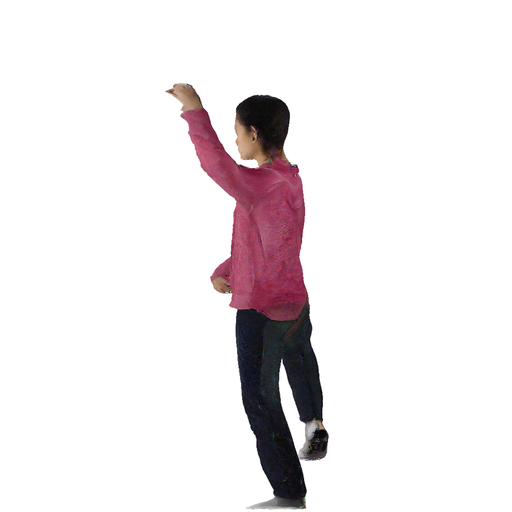}}\hfill
\mpage{0.07}{\includegraphics[width=\linewidth, trim=140 0 125 0, clip]{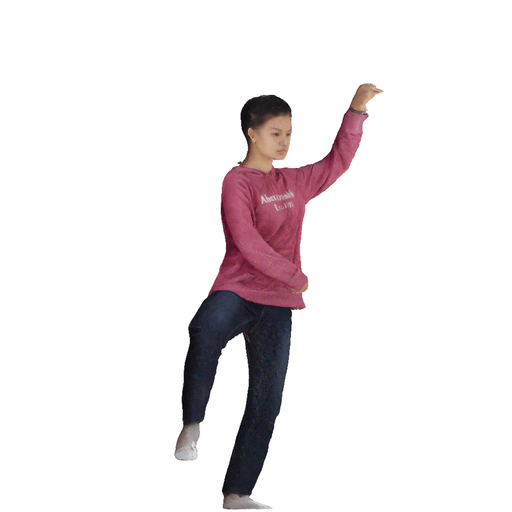}}\\

\vspace{-2mm}

\caption{\textbf{Visual comparisons on the THuman2.0 dataset.}
We compare our approach with prior methods~\cite{saito2019pifu,liu2021liquid,richardson2023texture,albahar2021pose,corona2022structured,qian2023magic123} on the THuman2.0 dataset~\cite{tao2021function4d}.
Our results showcase photorealistic images with consistent views that are consistent with the input images.
}
\label{fig:thuman}
\end{figure*}

We show visual comparisons of our results with the baselines on in-the-wild images from Adobe Stock in Figures~\ref{fig:teaser} and~\ref{fig:unsplash}, and on the THuman2.0 dataset~\cite{tao2021function4d} in~\figref{thuman}. 
These results demonstrate that our method produces high-resolution, photorealistic 3D human models that respect the appearance of the input, for a variety of input images.

\subsection{Ablation Study}
\label{sec:ablation}
\begin{table}[t]\setlength{\tabcolsep}{4pt}
	\centering%
\caption{
\textbf{Ablation study on the THuman2.0 dataset~\cite{tao2021function4d}.} We use the ground truth mesh to evaluate the effectiveness of initializing the back-view (B), and using normal (N) and silhouette (S) maps as guidance signals. 
}
\small
	\begin{tabular}{c|ccc|ccccc}
		\toprule
    ID & B & N & S  & PSNR$\uparrow$ & SSIM$\uparrow$ &  FID$\downarrow$ &  LPIPS$\downarrow$ & CLIP-score$\uparrow$   \\
            \midrule
            A & & \checkmark & \checkmark & 23.9463 & 0.9373 & 117.7447 & 0.0538 & 0.8013 \\
            B &\checkmark &  &            & 24.0494 & 0.9389 & 129.4944 & 0.0592 & 0.7896 \\
            C &\checkmark & \checkmark &  & \textbf{25.8709} & \underline{0.9449} & 108.5836 & 0.0506 & \underline{0.8041} \\
            D &\checkmark &  & \checkmark & 25.7199 & 0.9435 & \underline{101.3901} & \underline{0.0480} & 0.8013 \\
  E &\checkmark & \checkmark & \checkmark & \underline{25.8465} & \textbf{0.9453} & \textbf{98.9282}  & \textbf{0.0473} & \textbf{0.8069} \\
		\bottomrule
	\end{tabular}
 	\label{tab:ablation}
\end{table}

\subsubsection{Guidance signals.}
We validate our shape-guided diffusion inpainting in Table~\ref{tab:ablation}. We show the effect of using no guidance (B), only normal maps (C), only silhouette maps (D), and both normal and silhouette maps (E). We also show visual comparison in~\figref{shapeguidance}. The use of both normal maps and silhouette maps leads to better preserving the synthesized person’s shape and details and thus enhancing the quality of resulting 3D human models.

\subsubsection{Back-view synthesis.}
We validate the initial back-view synthesis using a human reposing technique~\cite{albahar2021pose} in Table~\ref{tab:ablation} (A vs. E). We also show visual comparison in~\figref{nobackview}. Having an initial back view encourages all other views to preserve the appearance of the input person, especially when clothing has nontrivial textures.

\label{sec:results}

\subsection{Limitations and Future Work}
\label{sec:limitations}
Our main limitation is the dependence on off-the-shelf methods~\cite{saito2020pifuhd,albahar2021pose} for the base geometry reconstruction and back-view synthesis. \figref{limitation} shows that our approach inherits the limitations of these methods. 
Another limitation is the lack of view-dependency. While clothing is mostly diffuse, human skin may exhibit view-dependent specular highlights. Extending our approach to view-dependent radiance would be an exciting direction, which can be addressed by future work.  
Furthermore, our work does not support human reposing and it requires per-subject UV texture optimization.
For the generality of our approach, we use off-the-shelf 3D shape reconstruction methods for clothed humans~\cite{saito2020pifuhd,xiu2022icon}, which are trained on 3D ground-truth data. We also use off-the-shelf human reposing method~\cite{albahar2021pose} for the back-view synthesis. Future work should also enable the high-fidelity 3D shape reconstruction of clothed humans and back-view synthesis with general-purpose 2D diffusion models.

\section{Conclusions}
\label{sec:conclusions}

We introduced a simple yet highly effective approach to generate a fully textured 3D human mesh from a \emph{single} image. 
Our experiments show that synthesizing a high-resolution and photorealistic texture for occluded views is now possible with shape-guided inpainting based on high-capacity latent diffusion models and a robust multi-view fusion method. 
While 3D human digitization relies on curated human-centric datasets either in 3D or 2D, our approach, for the first time, achieves superior synthesis results by leveraging a general-purpose large-scale diffusion model.
We believe our work will shed light on unifying data collection efforts for 3D human digitization and other general 2D/3D synthesis methods.

\bibliographystyle{ACM-Reference-Format}
\bibliography{main}

\end{document}